\documentclass{article} 

\usepackage{iclr2023_conference,times}

\usepackage[utf8]{inputenc} \usepackage[T1]{fontenc}    \usepackage[pagebackref=true]{hyperref}       \usepackage{url}            \usepackage{booktabs}       \usepackage{amsfonts}       \usepackage{nicefrac}       \usepackage{xcolor}         \usepackage{epsfig}
\usepackage{algorithm}
\usepackage{algorithmic}
\usepackage{amsthm}
\usepackage{amsmath}
\usepackage{float}
\usepackage{graphicx}
\usepackage{wrapfig}
\usepackage{makecell}
\usepackage{caption}
\usepackage{subcaption}
\usepackage{bm}
\usepackage{multirow}
\usepackage{microtype}
\usepackage{mathtools}
\usepackage{enumitem}
\usepackage{tikz}
\usepackage{listings}
\usepackage{pifont}\usepackage{xspace}

 \renewcommand*\backref[1]{\ifx#1\relax \else (Cited on #1) \fi}

\def\uint{\textsc{uint8}\xspace}
\def\graycode{\textsc{gray code}\xspace}
\def\uints{\textsc{uint8 (rand)}\xspace}
\def\onehot{\textsc{one hot}\xspace}
\def\cifar{\textsc{Cifar-10}\xspace}
\def\imagenet{\textsc{ImageNet 64$\times$64}\xspace}

\DeclareMathOperator*{\argmax}{arg\,max}

\title{Analog Bits: Generating Discrete Data using\\ Diffusion Models with Self-Conditioning}

\author{
   Ting Chen, Ruixiang Zhang$^\dagger$\thanks{$^\dagger$Work done as a student researcher at Google.}, Geoffrey Hinton\thanks{Code at https://github.com/google-research/pix2seq.} \\
   ~Google Research, Brain Team \\
   \texttt{\{iamtingchen,ruixiangz,geoffhinton\}@google.com} \\
}
\renewcommand\footnotemark{}

\iclrfinalcopy \begin{document}

\maketitle

\begin{abstract}
We present \textit{Bit Diffusion}: a simple and generic approach for generating discrete data with continuous state and continuous time diffusion models.
The main idea behind our approach is to first represent the discrete data as binary bits, and then train a continuous diffusion model to model these bits as real numbers which we call \textit{analog bits}.
    To generate samples, the model first generates the analog bits, which are then thresholded to obtain the bits that represent the discrete variables.
We further propose two simple techniques, namely Self-Conditioning and Asymmetric Time Intervals, which lead to a significant improvement in sample quality.
Despite its simplicity, the proposed approach can achieve strong performance in both discrete image generation and image captioning tasks.
For discrete/categorical image generation, we significantly improve previous state-of-the-art on both \cifar (which has $3K$ discrete 8-bit tokens) and \imagenet (which has $12K$ discrete 8-bit tokens), outperforming the best autoregressive model in both sample quality (measured by FID) and efficiency. For image captioning on MS-COCO dataset, our approach achieves competitive results compared to autoregressive models.
\end{abstract}

 \vspace{-.5em}
\section{Introduction}
\vspace{-.5em}

State-of-the-art generative models for discrete data, such as discrete images and text, are based on autoregressive modeling~\citep{van2016conditional,salimans2017pixelcnn++,parmar2018image,child2019generating,roy2021efficient,jun2020distribution,sutskever2014sequence,brown2020language,chowdhery2022palm}, where the networks, often Transformers~\citep{vaswani2017attention}, are trained to predict each token given its preceding ones in a sequential manner or with causal attention masks. One major drawback of such approaches is that they typically require computation and memory that is quadratic to the dimension of data (e.g., sequence length or image size), leading to difficulties in modeling large images or sequences. 
Another drawback is that, during generation, autoregressive models generate one token at a time so the total number of sequential sampling steps is often the same as the dimension of data, making it slow in generating large images or long sequences.

In contrast, diffusion models~\citep{sohl2015deep,ho2020denoising,song2020denoising}, or score-based generative models~\citep{song2019generative,song2020improved,song2021scorebased}, can model much higher dimensional data without running into computation and memory issues. 
During generation, diffusion models iteratively refine samples with a high degree of parallelism, so the total number of sequential sampling steps can be much less than the dimension of data. However, state-of-the-art diffusion models~\citep{dhariwal2021diffusion,ho2022cascaded,nichol2021glide,ramesh2022hierarchical,saharia2022photorealistic} can only generate continuous data (mainly real valued pixels), and have not yet achieved results competitive with autoregressive models in generating discrete/categorical data, such as generating discrete/categorical images~\citep{hoogeboom2021argmax,austin2021structured}.

\begin{figure}[t]
    \centering
    \includegraphics[width=0.85\linewidth]{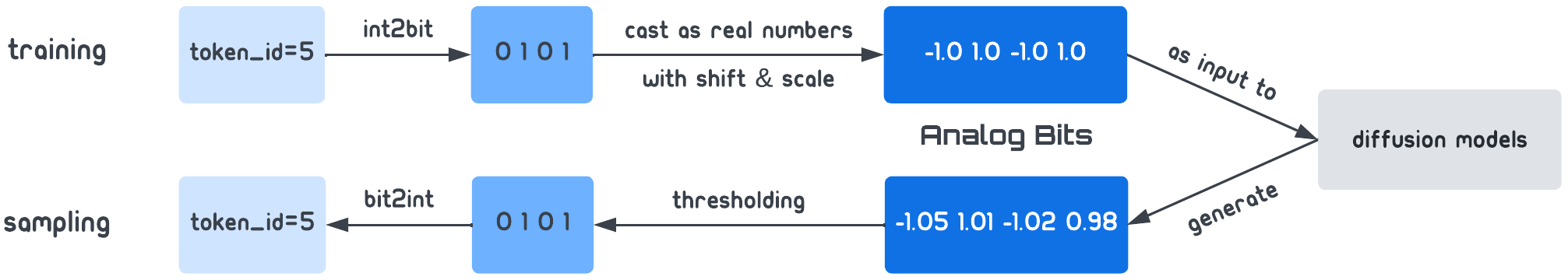}
    \vspace{-.5em}
    \caption{Bit Diffusion: modeling discrete data using continuous diffusion models with analog bits.}
    \label{fig:analog_bit_illustration}
    \vspace{-1em}
\end{figure}

In this work, we propose a simple and generic approach for enabling continuous state diffusion models to generate discrete data. The key ingredient in our approach is \textit{analog bits}: real numbers used to model the bits that represent the discrete data. Analog bits can be directly modeled by continuous state diffusion models, without requiring a discrete state space or re-formulation of the continuous diffusion process.
At sampling time, the generated analog bits can be decoded into discrete variables by a simple thresholding operation. Our approach, as illustrated in Figure~\ref{fig:analog_bit_illustration}, is based on the following high-level conjecture. With strong continuous generative models (diffusion models in particular), it should not be too difficult to generate highly concentrated bimodal data where each real-valued analog bit is close to a binary bit. To reduce the prediction loss (such as negative log likelihood), the network has to model structures among analog bits that can actually lead to meaningful discrete variables after thresholding.

Besides analog bits, we further propose two simple techniques, namely Self-Conditioning and Asymmetric Time Intervals that greatly improve the sample quality.
We evaluate the proposed approach on both discrete image generation, and image-conditional text / caption generation. On discrete \cifar and \imagenet, the proposed Bit Diffusion model significantly improves both existing discrete diffusion models but also the best autoregressive model. For example, on categorical \cifar, the best autoregressive model~\citep{jun2020distribution} obtains a FID of $12.75$, while our model (with $\nicefrac{1}{3}$ of the model size of the autoregressive model, using 100 instead of 3072 sequential inference steps) achieves a much better $6.93$. 
For image captioning on MS-COCO dataset, our model achieves a result competitive with a strong autoregressive captioner based on a Transformer.

 \vspace{-.5em}
\section{Method}
\vspace{-.5em}

\textbf{Preliminaries }
We start with a short introduction to diffusion models~\citep{sohl2015deep,ho2020denoising,song2020denoising,song2021scorebased}. Diffusion models learn a series of state transitions to map noise $\bm \epsilon$ from a known prior distribution to $\bm x_0$ from the data distribution.
To learn this (reverse) transition from the noise distribution to the data distribution, 
a forward transition from $\bm x_0$ to $\bm x_t$ is first defined:
\begin{equation}
\label{eq:diffusion_forward}
\bm x_t = \sqrt{\gamma(t)} ~\bm x_0 + \sqrt{1-\gamma(t)} ~\bm \epsilon,
\end{equation}
where $\bm \epsilon \sim \mathcal{N}(\bm 0, \bm I)$, $t\sim \mathcal{U}(0, T)$ is a continuous variable, and $\gamma(t)$ is a monotonically decreasing function from 1 to 0.
Instead of directly learning a neural net to model the transition from $\bm x_t$ to $\bm x_{t-\Delta}$, one can learn a neural net $f(\bm x_t, t)$ to predict $\bm x_0$ (or $\bm \epsilon$) from $\bm x_t$, and estimate $\bm x_{t-\Delta}$ from $\bm x_{t}$ and estimated $\tilde{\bm x}_0$ (or $\tilde{\bm \epsilon}$). 
This training of $f(\bm x_t, t)$ is based on denoising with a $\ell_2$ regression loss:
\begin{equation}
\label{eq:diffusion_l2loss}
\mathcal{L}_{\bm x_0} =\mathbb{E}_{t\sim\mathcal{U}(0, T), \bm \epsilon\sim\mathcal{N}(\bm 0, \bm 1)}\|f(\sqrt{\gamma(t)} ~\bm x_0 + \sqrt{1-\gamma(t)} ~\bm \epsilon, t) - \bm x_0 \|^2.
\end{equation}
To generate samples from a learned model, it follows a series of (reverse) state transition $\bm x_T\rightarrow \bm x_{T-\Delta} \rightarrow  \cdots  \rightarrow \bm x_0$. This can be achieved by iteratively applying denoising function $f$ on each state $\bm x_t$ to estimate $\bm x_0$, and then make a transition to $\bm x_{t-\Delta}$ with the estimated $\tilde{\bm x}_0$ (using transition rules such as those specified in DDPM~\citep{ho2020denoising} or DDIM~\citep{song2020denoising}).
Note that state transitions in these diffusion models assume a continuous data space and state space. Therefore, one cannot directly apply it to model and generate discrete/categorical data. 

\textbf{Analog Bits }
A discrete data variable from an alphabet of size $K$ can be represented
using $n=\lceil\log_2 K\rceil$ bits, as $\{0, 1\}^{n}$.  Due to the discreteness, existing work has to re-formulate continuous diffusion models by adopting a discrete data space and state space~\citep{sohl2015deep,hoogeboom2021argmax,austin2021structured}. 
In contrast, we propose to simply cast the binary bits $\{0, 1\}^{n}$ into real numbers $\mathbb{R}^{n}$ for the continuous diffusion models~\footnote{After casting as real numbers, one may also transform them by shifting and scaling from $0, 1$ to $-b, b$.}. We term these real numbers \textit{analog bits} since they learn to share the same bimodal values as binary bits but are modeled as real numbers.  To draw samples, we follow the same procedure as sampling in a continuous diffusion model, except that we apply a quantization operation at the end by simply thresholding the generated analog bits. 
This yields binary bits which can be then converted into original discrete/categorical variables.
Notably, there is no hard constraint to force the model to generate exact binary bits, but we expect a strong continuous generative model to generate real numbers that exhibit very clear bimodal concentrations and this is what happens in our experiments.

For simplicity, we use the same regression loss function (Eq.~\ref{eq:diffusion_l2loss}) for modeling analog bits. However, it is possible to use other loss functions such as the cross entropy loss. We also note that the binary encoding mechanism for constructing analog bits is extensible as well (e.g., one-hot encoding). Extensions of loss functions and binary encoding are described in the appendix~\ref{sec:alt_loss}.

\textbf{Self-Conditioning }
Conditioning is a useful technique for improving diffusion models~\citep{nichol2021improved,ho2022cascaded}. However, a typical conditioning variable is either from some external sources, such as class labels~\citep{nichol2021improved} or low-resolution images from another network~\citep{nichol2021improved,saharia2021image,ho2022cascaded}. Here we propose a technique for the model to directly condition on previously generated samples of its own during the iterative sampling process, which can significantly improve the sample quality of diffusion models.

\begin{figure}[t]
    \centering
    \begin{subfigure}{0.48\textwidth}
    \centering
    \begin{tikzpicture}[scale=0.7]
\node at (-1,0.4) () {$\cdots$};
    \node at (0,1.5) (x_tp1){$\bm x_{t+\Delta}$};
    \node at (3,1.5) (x) {$\bm x_{t}$};
    \node at (6,1.5) (x_tm1) {$\bm x_{t-\Delta}$};
\node at (7.,0.4) () {$\cdots$};
    \node at (1.5,0) (x0_1) {$\tilde{\bm{x}}_0$};
    \node at (5,0) (x0_2) {$\tilde{\bm{x}}'_0$};
    \path[->] 
(x_tp1)  edge [>=latex] node[below,rotate=-25] {} (x)
        (x)  edge [>=latex] node[below,rotate=25] {} (x_tm1)
(x_tp1)  edge [>=latex] node[left,rotate=0] {} (x0_1)
        (x0_1)  edge [>=latex] node[right,rotate=0] {} (x)
        (x)  edge [>=latex] node[left,rotate=0] {} (x0_2)
        (x0_2)  edge [>=latex] node[right,rotate=0] {} (x_tm1)
        ;
    \end{tikzpicture}
    \caption{\label{fig:self_cond_chain1}Standard reverse diffusion steps.}
    \end{subfigure}
    \hfill
    \begin{subfigure}{0.48\textwidth}
    \centering
    \begin{tikzpicture}[scale=0.7]
\node at (-1,0.4) () {$\cdots$};
    \node at (0,1.5) (x_tp1){$\bm x_{t+\Delta}$};
    \node at (3,1.5) (x) {$\bm x_{t}$};
    \node at (6,1.5) (x_tm1) {$\bm x_{t-\Delta}$};
\node at (7.,0.4) () {$\cdots$};
    \node at (1.5,0) (x0_1) {$\tilde{\bm{x}}_0$};
    \node at (5,0) (x0_2) {$\tilde{\bm{x}}'_0$};
    \path[->] 
(x_tp1)  edge [>=latex] node[below,rotate=-25] {} (x)
        (x)  edge [>=latex] node[below,rotate=25] {} (x_tm1)
(x_tp1)  edge [>=latex] node[left,rotate=0] {} (x0_1)
(x0_1)  edge [>=latex] node[right,rotate=0] {} (x)
        (x)  edge [>=latex] node[left,rotate=0] {} (x0_2)
        (x0_2)  edge [>=latex] node[right,rotate=0] {} (x_tm1)
        (x0_1)  edge [>=latex] node[right,rotate=0] {} (x0_2)
;
    \end{tikzpicture}
    \caption{\label{fig:self_cond_chain2}Self-Conditioning on the previous $\bm x_0$ estimate.}
    \end{subfigure}
    \vspace{-.5em}
    \caption{An illustration of reverse diffusion sampling steps (a) without or (b) with Self-Conditioning. $\tilde{\bm x}_0$ denotes the estimation of data sample by the denoising network $f$ at a sampling step. We propose to condition the network \textit{directly} on its previously generated/estimated samples.}
    \label{fig:self_cond_chain}
    \vspace{-1.5em}
\end{figure}
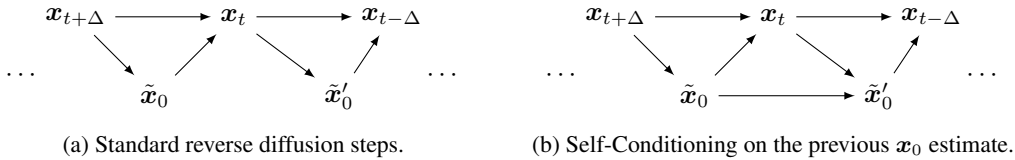

In a typical diffusion sampling process, the model iteratively predicts $\bm x_0$ (or $\bm \epsilon$) in order to progress the chain of mapping noise into data. However, as shown in Figure~\ref{fig:self_cond_chain1}, the previously estimated $\tilde{\bm x}_0$ is simply discarded when estimating $\bm x_0$ from a new time step, i.e. the denoising function $f(\bm x_t, t)$ does not directly depend on a previously estimated $\tilde{\bm x}_0$. Here we consider a slightly different denoising function of $f(\bm x_t, \tilde{\bm x}_0, t)$ that also takes previous generated samples as its input, illustrated in Figure~\ref{fig:self_cond_chain2}. A simple implementation of Self-Conditioning is to concatenate $\bm x_t$ with previously estimated $\tilde{\bm x}_0$. Given that $\tilde{\bm x}_0$ is from the earlier prediction of the model in the sampling chain, this comes at a negligible extra cost during sampling.
In order to train the denoising function $f(\bm x_t, \tilde{\bm x}_0, t)$, we make some small changes to the training. With some probability (e.g., $50\%$), we set $\tilde{\bm x}_0=\bm 0$ which falls back to modeling without Self-Conditioning. At other times, we first estimate $\tilde{\bm x}_0 = f(\bm x_t, \bm 0, t)$ and then use it for Self-Conditioning. Note that we do not backpropagate through the estimated $\tilde{\bm x}_0$ so the overall increase of training time is small (e.g., less than 25\%).

\textbf{Asymmetric Time Intervals }
Besides Self-Conditioning, we identify another factor, time step $t$, that can also impact Bit Diffusion models. Time step $t$ is an integral part of both denoising network $f(\bm x_t, t)$ as well as the state transitions. 
During a typical reverse diffusion process, the model takes symmetric time intervals (i.e., $\Delta$ as in $t\rightarrow t-\Delta$) for both the state transition and time reduction itself, resulting in the same/shared $t$ for both arguments of $f(\bm x_t, t)$.
However, we find that, when taking large reverse steps, using asymmetric time intervals, implemented via a simple manipulation of time scheduling at generation, can lead to improved sampling quality for Bit Diffusion models.

\begin{figure}[b]
    \vspace{-1.5em}
    \captionsetup[subfigure]{labelformat=empty}
    \centering
    \begin{subfigure}{0.1\textwidth}
      \centering
      \includegraphics[width=\linewidth]{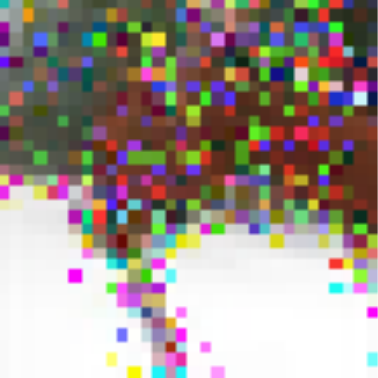}
      \caption{{\tiny$\tilde{\bm x}_{t=0.1}|\bm x_{t=0.6}$}}
    \end{subfigure}
    ~~~~~~~~
    \begin{subfigure}{0.1\textwidth}
      \centering
      \includegraphics[width=\linewidth]{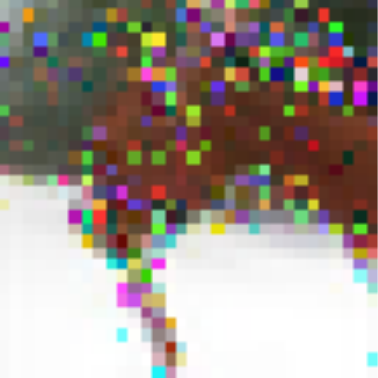}
      \caption{{\tiny$f(\tilde{\bm x}_{t=0.1}, t'\text{=}0.1)$}}
    \end{subfigure}
    ~~~~~~~~
    \begin{subfigure}{0.1\textwidth}
      \centering
      \includegraphics[width=\linewidth]{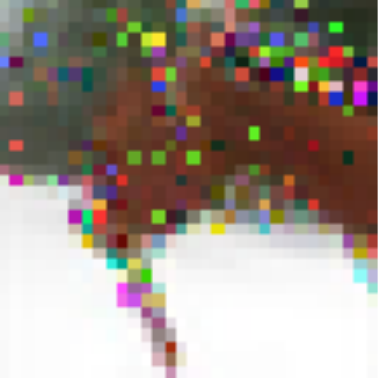}
      \caption{{\tiny$f(\tilde{\bm x}_{t=0.1}, t'\text{=}0.3)$}}
    \end{subfigure}
    ~~~~~~~~
    \begin{subfigure}{0.1\textwidth}
      \centering
      \includegraphics[width=\linewidth]{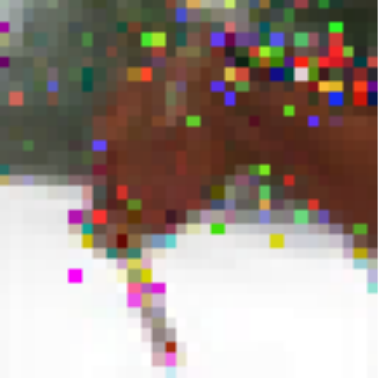}
      \caption{{\tiny $f(\tilde{\bm x}_{t=0.1}, t'\text{=}0.5)$}}
    \end{subfigure}
    ~~~~~~~~
    \begin{subfigure}{0.1\textwidth}
      \centering
      \includegraphics[width=\linewidth]{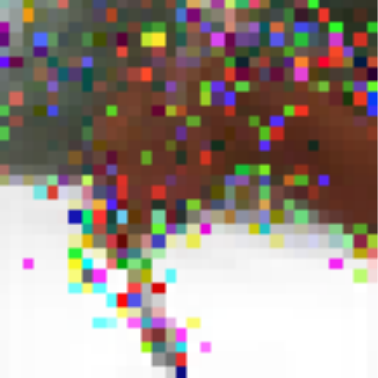}
      \caption{{\tiny $f(\tilde{\bm x}_{t=0.1}, t'\text{=}0.7)$}}
    \end{subfigure}
    \vspace{-.5em}
    \caption{When taking a large reverse step from $\bm x_{t=0.6}$ to $\bm x_{t=0.1}$ in Bit Diffusion with maximum time $T=1.0$, we see that asymmetric time intervals with a positive time difference $\xi$ improve the denoising quality of $\bm x_{t=0.1}$ (by reducing the number of noisy pixels).}
\label{fig:td_toy_denoising}
\end{figure}
More specially, with asymmetric time intervals during the sampling process, we have $f(\bm x_t, t')$, where $t'=t+\xi$ and $\xi$ is a small non-negative time difference parameter.
Note that training remains unchanged, and the same/shared $t$ is used for both arguments of the $f(\bm x_t, t)$.
Figure~\ref{fig:td_toy_denoising} illustrates the effect with a trained Bit Diffusion model, where it is asked to take two reversing steps from a state $x_t$ constructed using the forward diffusion, and it shows that asymmetric time intervals reduce the number of noisy pixels (after thresholding and converting back to discrete variables).

\textbf{Putting it together }
Algorithm~\ref{alg:train} and~\ref{alg:sample} summarize the training and sampling algorithms for the proposed Bit Diffusion model with Analog Bits, Self-Conditioning, and Asymmetric Time Intervals (via the \texttt{td} parameter). The proposed changes to the existing diffusion models are highlighted in blue. Note that unlike standard diffusion models~\citep{sohl2015deep,ho2020denoising,nichol2021improved}, we use a continuous time parameterization between 0 and 1 instead of a fixed discrete time for maximal flexibility but they perform similarly. More details of the algorithm (including some important functions) can be found in Appendix~\ref{sec:alg_details}.

\vspace{-1.5em}
\begin{minipage}[t]{0.499\textwidth}
\begin{algorithm}[H]
\small
\caption{\small Bit Diffusion training algorithm.
}
\label{alg:train}
\definecolor{codeblue}{rgb}{0.25,0.5,0.5}
\definecolor{codekw}{rgb}{0.85, 0.18, 0.50}
\lstset{
  backgroundcolor=\color{white},
  basicstyle=\fontsize{7.2pt}{7.2pt}\ttfamily\selectfont,
  columns=fullflexible,
  breaklines=true,
  captionpos=b,
  commentstyle=\fontsize{7.2pt}{7.2pt}\color{codeblue},
  keywordstyle=\fontsize{7.2pt}{7.2pt}\color{codekw},
  escapechar={|}, 
}
\begin{lstlisting}[language=python]
def train_loss(x):
  # Binary encoding: discrete to analog bits.
  |\color{blue}x\_bits = int2bit(x).astype(float)|
  |\color{blue}x\_bits = (x\_bits * 2 - 1) * scale|
    
  # Corrupt data.
  t = uniform(0, 1)
  eps = normal(mean=0, std=1)
  x_crpt = sqrt(gamma(t)) * x_bits + 
              sqrt(1 - gamma(t)) * eps

  # Compute self-cond estimate.
  |\color{blue}x\_pred = zeros\_like(x\_crpt)|
  |\color{blue}if self\_cond and uniform(0, 1) > 0.5:|
    |\color{blue}x\_pred = net(cat([x\_crpt, x\_pred], -1), t)|
    |\color{blue}x\_pred = stop\_gradient(x\_pred)|
    
  # Predict and compute loss.
  x_pred = net(cat([x_crpt, |\color{blue}x\_pred|], -1), t)
  loss = (x_pred - x_bits)**2
  return loss.mean()
\end{lstlisting}
\end{algorithm}
\end{minipage}
\hfill
\begin{minipage}[t]{0.499\textwidth}
\begin{algorithm}[H]
\small
\caption{\small Bit Diffusion sampling algorithm.
}
\label{alg:sample}
\definecolor{codeblue}{rgb}{0.25,0.5,0.5}
\definecolor{codekw}{rgb}{0.85, 0.18, 0.50}
\lstset{
  backgroundcolor=\color{white},
  basicstyle=\fontsize{7.5pt}{7.5pt}\ttfamily\selectfont,
  columns=fullflexible,
  breaklines=true,
  captionpos=b,
  commentstyle=\fontsize{7.5pt}{7.5pt}\color{codeblue},
  keywordstyle=\fontsize{7.5pt}{7.5pt}\color{codekw},
  escapechar={|}, 
}
\begin{lstlisting}[language=python]
def generate(steps, |\color{blue}td|=0):
  x_t = normal(mean=0, std=1)
  |\color{blue}x\_pred = zeros\_like(x\_t)|
  
  for step in range(steps):
    # Get time for current and next states.
    t_now = 1 - step / steps
    t_next = max(1 - (step+1+|\color{blue}td|) / steps, 0)
        
    # Predict x_0.
    |\color{blue}if not self\_cond:|
      |\color{blue}x\_pred = zeros\_like(x\_t)|
    x_pred = net(cat([x_t,|\color{blue}x\_pred|],-1), t_now)
        
    # Estimate x at t_next.
    x_t = ddim_or_ddpm_step(
        x_t, x_pred, t_now, t_next)
    
  # Binary decoding to discrete data.
  return |\color{blue} bit2int(x\_pred > 0)|
\end{lstlisting}
\end{algorithm}
\end{minipage}

\vspace{-0.5em} \vspace{-.5em}
\section{Experiments}
\vspace{-.5em}

We experiment with two different discrete data generation tasks, namely discrete/categorical image generation, and image captioning (image-conditional text generation).

\vspace{-.5em}
\subsection{Experimental Setup and Implementation Details}
\vspace{-.5em}

\textbf{Datasets } We use \cifar~\citep{krizhevsky2009learning} and \imagenet~\citep{deng2009imagenet}~\footnote{Following~\citep{brock2018large,nichol2021improved}, we center crop and area downsample
images to 64$\times$64.} for image generation experiments. We adopt widely used FID~\citep{heusel2017gans} as the main evaluation metric, and it is computed between 50K generated samples and the whole training set. For image captioning, following ~\citep{chen2022unified}, we use MS-COCO 2017 captioning dataset~\citep{lin2014microsoft}. 

\textbf{Binary encoding } Each pixel consists of 3 sub-pixels (RGB channels), and each sub-pixel is an integer in $[0, 256)$ representing the intensity. Standard continuous generative models cast RGB channels as real numbers and normalize them in $[-1,1]$. For discrete image generation, we consider three discrete encoding for sub-pixels, namely \uint, \graycode, and \uints. In \uint, we use 8-bit binary codes converted from the corresponding sub-pixel integer in $[0, 256)$. In \graycode, we assign 8-bit binary codes uniquely to each sub-pixel integer such that two adjacent integers only differ by 1 bit. And in \uints, we assign 8-bit binary codes to every sub-pixel integer by randomly shuffling the integer-to-bits mapping in \uint. The binary codes in \uint and \graycode are loosely correlated with its original sub-pixel intensities, while \uints has no correlation so each sub-pixel is a categorical variable. The details of the binary codes and their correlations with sub-pixel intensity can be found in the appendix~\ref{sec:pixel_binary_code}. We shift and scale the binary bits from $0, 1$ to $-1, 1$ for the analog bits.

For image captioning, we follow~\citep{chen2022unified}, and use sentencepiece~\citep{kudo2018sentencepiece} with a vocabulary of size 32K to tokenize the captions. After tokenization, we encode each token into 15 analog bits using the binary codes converted from the corresponding integer. We set the maximum number of tokens to 64 so the total sequence length is 960 bits. Since we directly model bits, it is also possible to directly work with their byte representations without a tokenizer, but we leave this for future work.

\textbf{Architecture } We use the U-Net architecture~\citep{ho2020denoising,nichol2021improved,ronneberger2015u} for image generation. For \cifar, we use a single channel dimension of 256, 3 stages and 3 residual blocks~\citep{he2016deep} per stage, with a total of 51M parameters. We only use dropout~\citep{srivastava2014dropout} of 0.3 for continuous diffusion models on \cifar. For \imagenet, following~\citep{nichol2021improved}, we use a base channel dimension of 192, multiplied by 1,2,3,4 in 4 stages and 3 residual blocks per stage, which account for a total of 240M parameters~\footnote{Our model is about 30M parameters smaller than that used in~\citep{nichol2021improved} as we drop the middle blocks for convenience, which may have a minor effect on performance.}. For \uints encoding, we find the following ``\textit{softmax factorization}'' architectural tweak on the final output layer can lead to a better performance. Instead of using a linear output layer to predict analog bits directly, we first predict a probability distribution over 256 classes per sub-pixel (with each class corresponds to one of the 256 different 8-bit codes), and then map class distribution into analog bits by taking weighted average over all 256 different 8-bit codes.

For image captioning, we follow the architecture used in~\citep{chen2021pix2seq,chen2022unified}, with a pre-trained image encoder using the object detection task, for both autoregressive baseline as well as the proposed method. Both decoders are randomly initialized 6-layer Transformer~\citep{vaswani2017attention} decoder with 512 dimension per layer. For the autoregressive decoder, the token attention matrix is offset by the causal masks, but it is non-masked all-to-all attention for our Bit Diffusion.

\textbf{Other settings } We train our models with the Adam optimizer~\citep{kingma2014adam}. For \cifar, we train the model for $1.5$M steps with a constant learning rate of $0.0001$ and batch size of $128$. For \imagenet, we train the model for 500K steps with a constant learning rate of $0.0002$~\footnote{For \uints encoding, we use learning rate of $0.0001$ instead.} and batch size of $1024$. For Bit Diffusion, we use Self-Conditioning by default, unless otherwise specified. We use an exponential moving average of the weights during training with a decay factor of $0.9999$. For our best image generation results, we sweep over a few sampling hyper-parameters, such as sampler (DDIM vs DDPM), sampling steps in $\{100, 250, 400, 1000\}$, and time difference in $\{0., 0.01, 0.1, 0.2, 0.5\}$. 

\vspace{-.5em}
\subsection{Discrete Image Generation}
\vspace{-.5em}
\label{sec:discrete_img_gen}
\begin{table}[t]
    \small
    \centering
    \caption{Comparison of FIDs on unconditional and class-conditional \cifar. Note that both \uint and \graycode are only partial/weakly ordinal (see Appendix~\ref{sec:pixel_binary_code}). Our Bit Diffusion achieves state-of-the-art FIDs in generating discrete images, beating the best autoregressive model.}
    \vspace{-.5em}
    \begin{tabular}{lccc}
        \toprule
        \multirow{2}{*}{Method} & \multirow{2}{*}{State space} & FID & FID \\ 
        && (Unconditional)&(Conditional) \\
        \midrule
        \textit{On continuous pixels (as reference):\hfill} & &\\
        ~~DDPM~\citep{ho2020denoising} & Continuous & $3.17$ & - \\
        ~~DDPM (our reproduction) & Continuous & $\mathbf{3.14}$ & $\mathbf{2.95}$ \\
        \midrule
        \textit{On discrete (partial) ordinal pixels:\hfill} & &\\
        ~~D3PM Gauss+Logistic~\citep{austin2021structured}  & Discrete & $7.34$ & - \\
        ~~$\tau$LDR-10~\citep{campbell2022continuous} & Discrete & $3.74$ & - \\
        ~~Bit Diffusion on \uint & Continuous & $\mathbf{3.48}$ & $\mathbf{2.72}$ \\
        ~~Bit Diffusion on \graycode & Continuous & $3.86$ & $2.94$ \\
        \midrule
        \textit{On categorical pixels:\hfill} & &\\
        ~~D3PM uniform~\citep{austin2021structured} & Discrete & $51.27$ & - \\
        ~~D3PM absorbing~\citep{austin2021structured} & Discrete & $30.97$ & - \\
        ~~Autoregressive Transformer~\citep{jun2020distribution} & Discrete & $12.75$ & - \\
        ~~Bit Diffusion on \uints & Continuous & $\mathbf{6.93}$ & $\mathbf{6.43}$ \\
        \bottomrule
    \end{tabular}
    \label{tab:cifar10_sota}
    \vspace{-1em}
\end{table}

\begin{table}[t]
    \small
    \centering
    \caption{Comparison of FIDs on class-conditional \imagenet. The corresponding samples can be found in Figure~\ref{fig:imagenet_cond_samples} and~\ref{fig:rand_samples_imagenet}.}
    \vspace{-.5em}
    \begin{tabular}{cccc}
        \toprule
        DDPM (our repo.) & Bit Diffusion & Bit Diffusion & Bit Diffusion\\
        on continuous pixels & on \uint & on \graycode & on \uints\\ 
        \midrule
         $3.43$ & $4.84$ & $5.14$ & $8.76$ \\
        \bottomrule
    \end{tabular}
    \label{tab:imagenet_sota}
    \vspace{-1.5em}
\end{table}

\begin{figure*}[t]
    \centering
    \begin{subfigure}{0.46\textwidth}
      \centering
      \includegraphics[width=\linewidth]{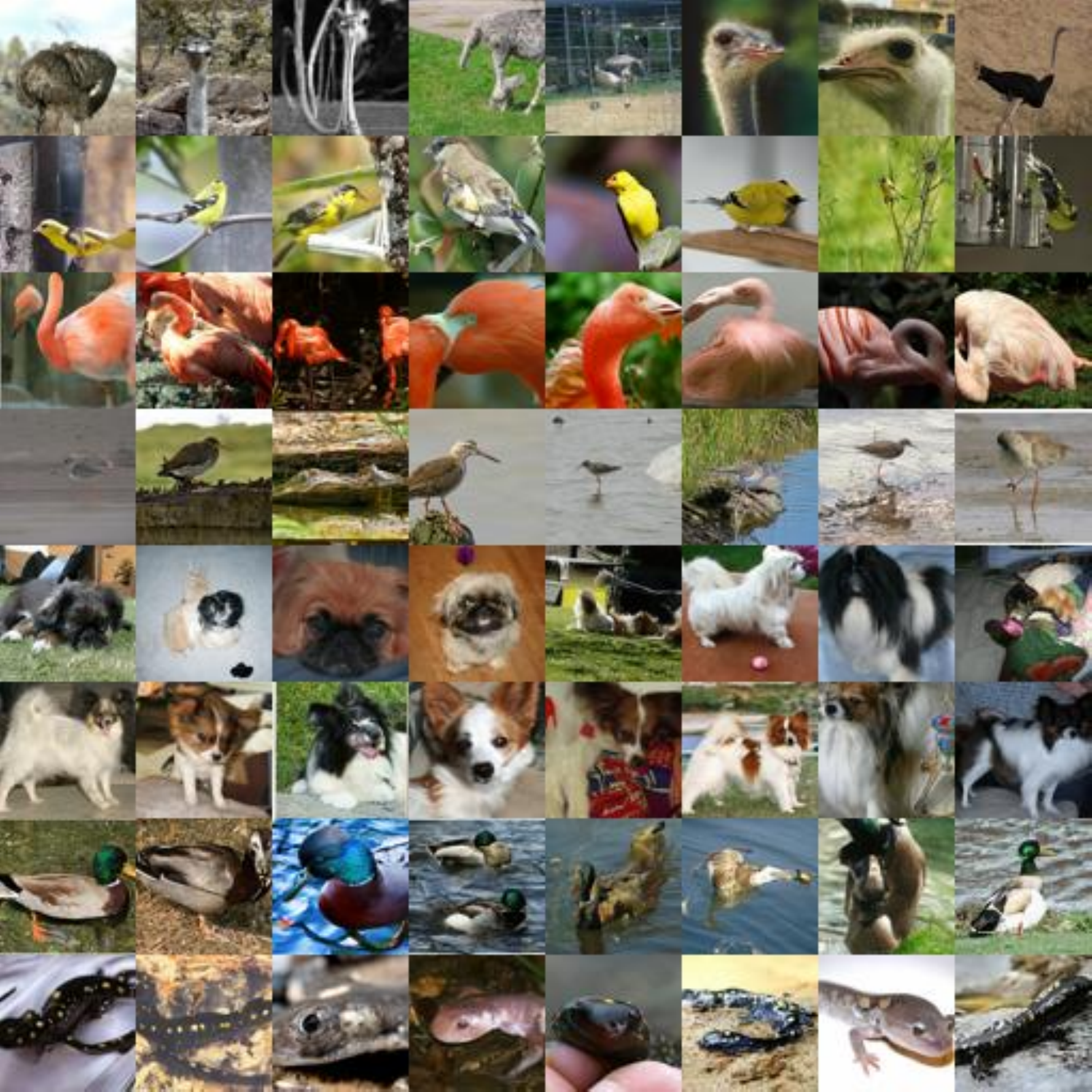}
      \caption{DDPM on continuous pixels (FID=$3.43$)}
    \end{subfigure}
    ~~~~
    \begin{subfigure}{0.46\textwidth}
      \centering
      \includegraphics[width=\linewidth]{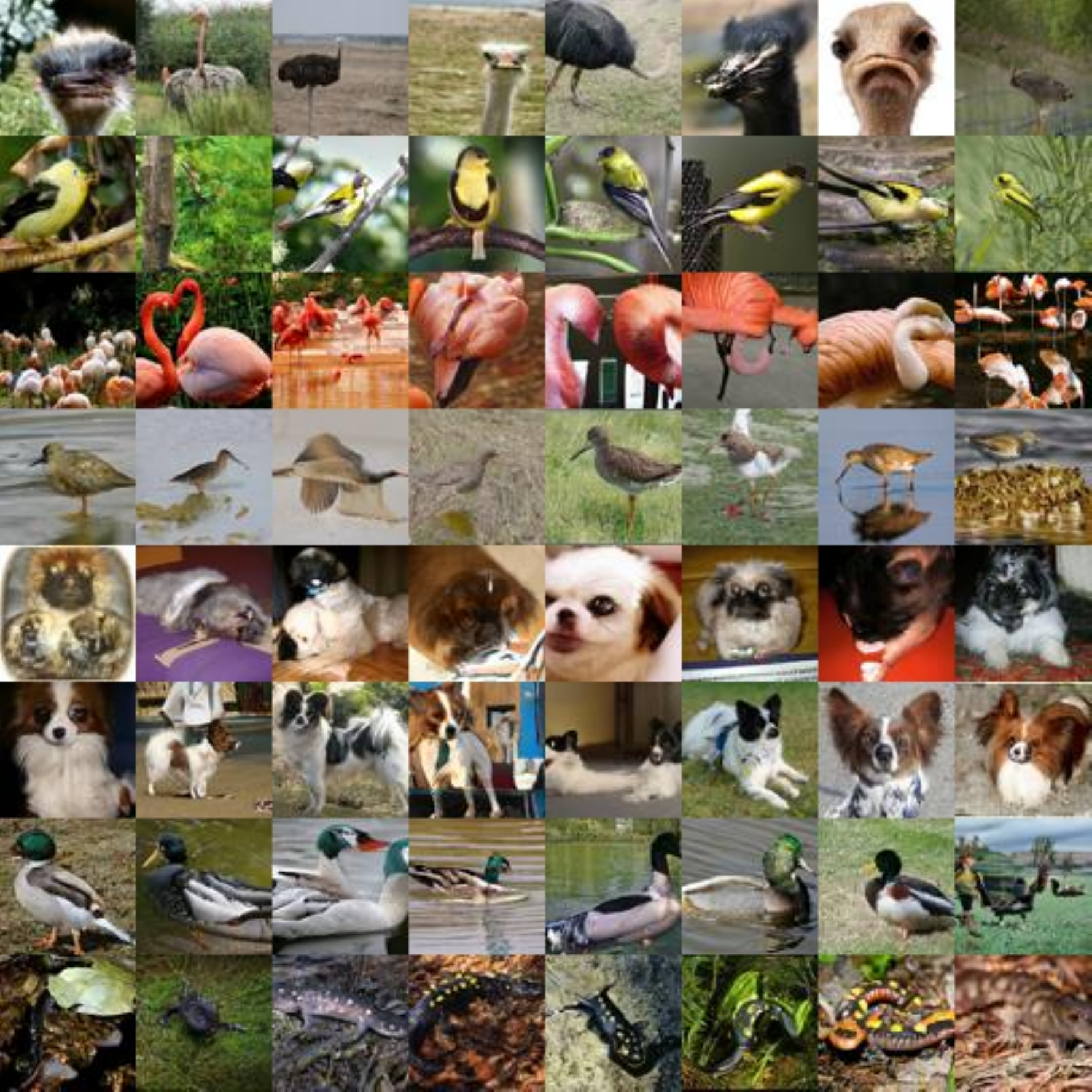}
      \caption{Bit Diffusion on \uint. (FID=$4.84$)}
    \end{subfigure}
    \begin{subfigure}{0.46\textwidth}
      \centering
      \includegraphics[width=\linewidth]{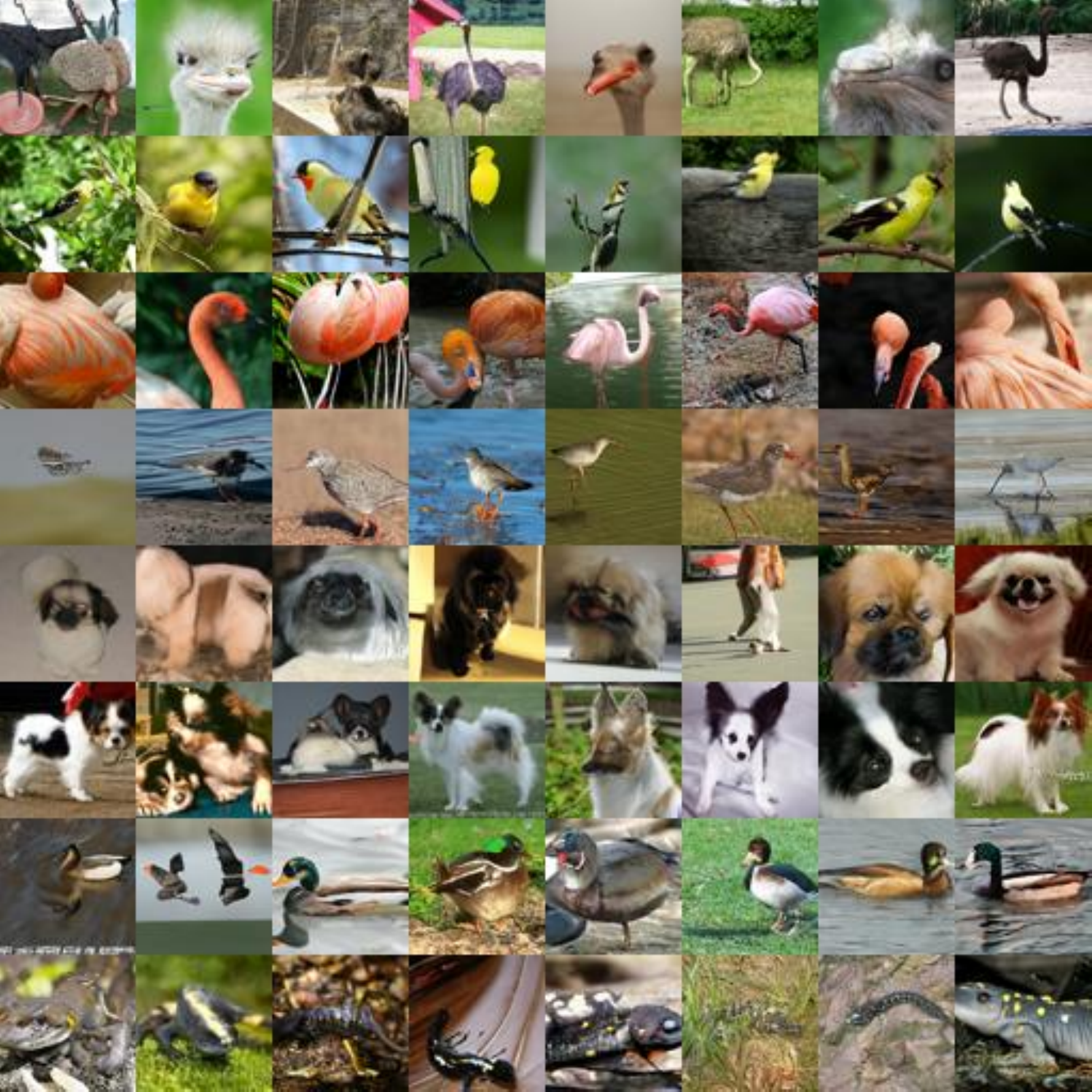}
      \caption{Bit Diffusion on \graycode (FID=$5.14$)}
    \end{subfigure}
    ~~~~
    \begin{subfigure}{0.46\textwidth}
      \centering
      \includegraphics[width=\linewidth]{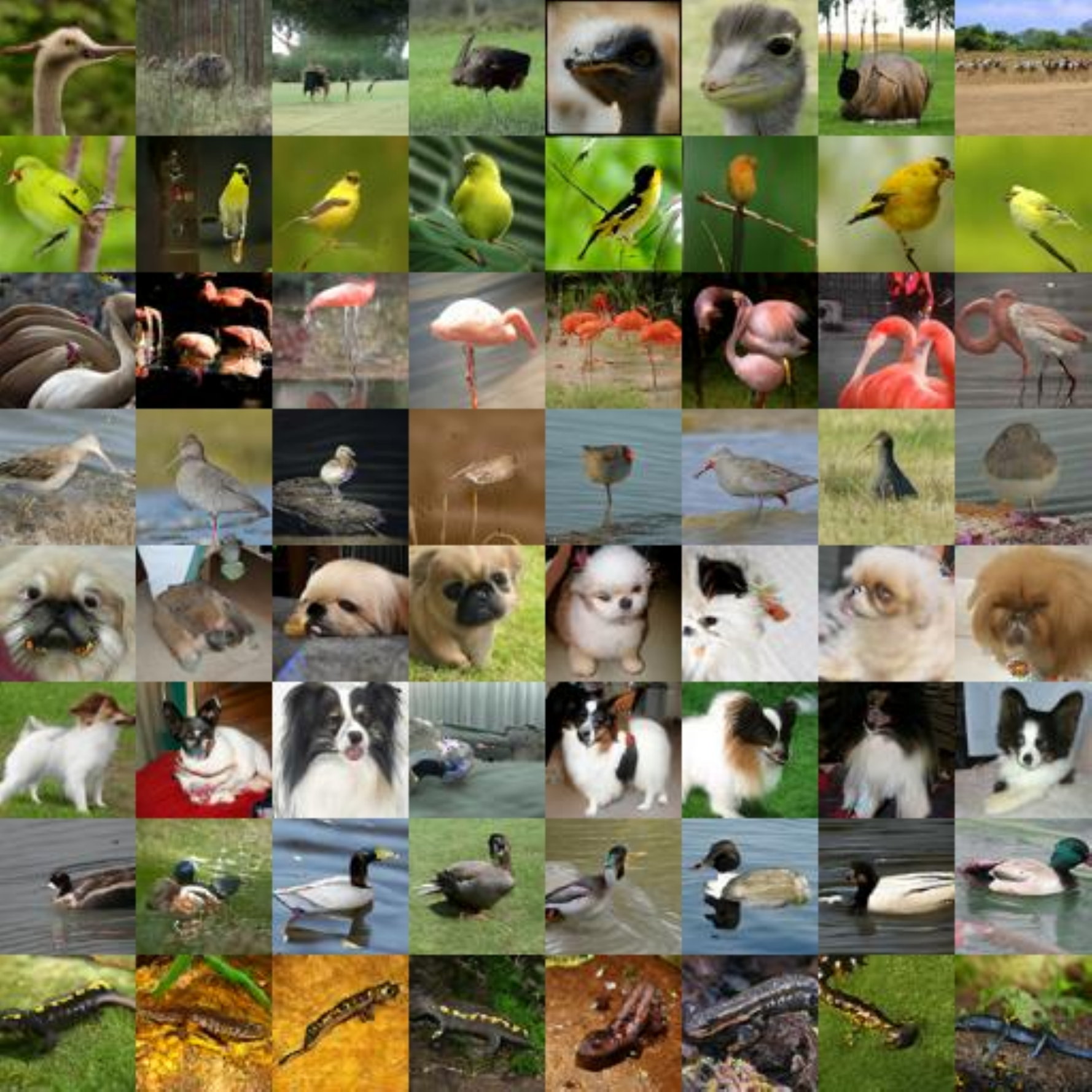}
      \caption{Bit Diffusion on \uints (FID=$8.76$)}
    \end{subfigure}
    \caption{\label{fig:imagenet_cond_samples} Class-conditional generations on continuous v.s. discrete ImageNet 64$\times$64. Each row represents random samples conditioned on a class, and the classes are adopted from~\citep{nichol2021improved}, namely, $9$: ostrich, $11$: goldfinch, $130$: flamingo, $141$: redshank, $154$: pekinese, $157$: papillon, $97$: drake and $28$: spotted salamander. 
    More samples from random classes are shown in Figure~\ref{fig:rand_samples_imagenet}.
    \vspace{-2em}
    }
\end{figure*}
We compare our model against state-of-the-art generative models~\citep{ho2020denoising,austin2021structured,campbell2022continuous,jun2020distribution} on generating discrete \cifar images in Table~\ref{tab:cifar10_sota}. Our model achieves better results compared to both existing discrete diffusion models and the best autoregressive model. When compared to continuous diffusion models (i.e., DDPM), our Bit Diffusion models on \uint and \graycode can achieve similar performance.

Discrete generation of \imagenet is significantly harder than \cifar, and we have not found other competing methods that report FIDs,
so we only compare the proposed method against DDPM on continuous pixels. Results are shown in Table~\ref{tab:imagenet_sota}. We find that the diffusion model on continuous pixels has the best FID while the diffusion model on \uints, i.e., categorical data, has the worst FID, indicating the increase of hardness when removing intensity/order information in sub-pixels.
Note that, in these experiments, there is no extra model capacity to compensate for the loss of intensity/order information since the model sizes are the same.
Figure~\ref{fig:imagenet_cond_samples} shows generated images of different diffusion models on continuous and discrete \imagenet. Despite the differences in FIDs, visually these samples look similar.

\begin{figure*}[t]
    \centering
    \begin{subfigure}{0.328\textwidth}
      \centering
      \includegraphics[width=\linewidth]{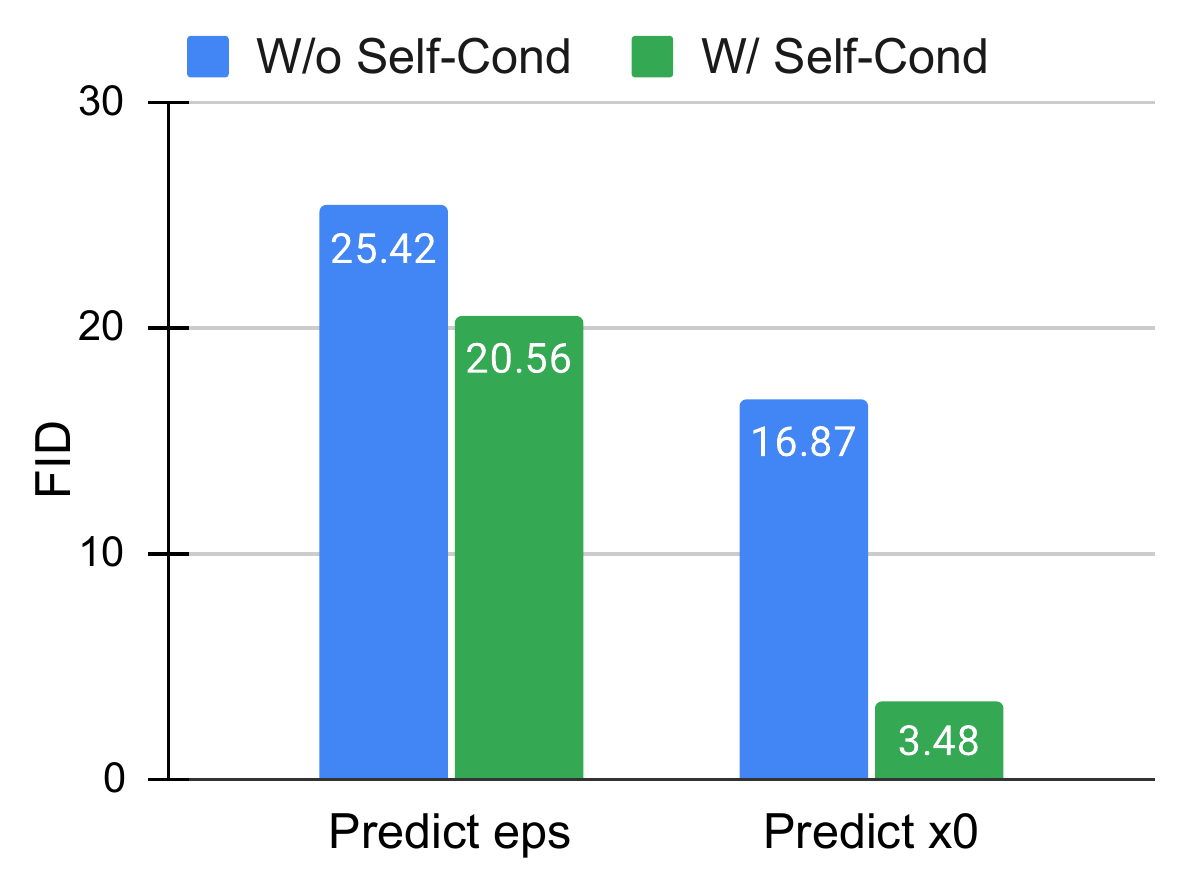}
      \caption{\cifar, \uint.}
    \end{subfigure}
    \begin{subfigure}{0.328\textwidth}
      \centering
      \includegraphics[width=\linewidth]{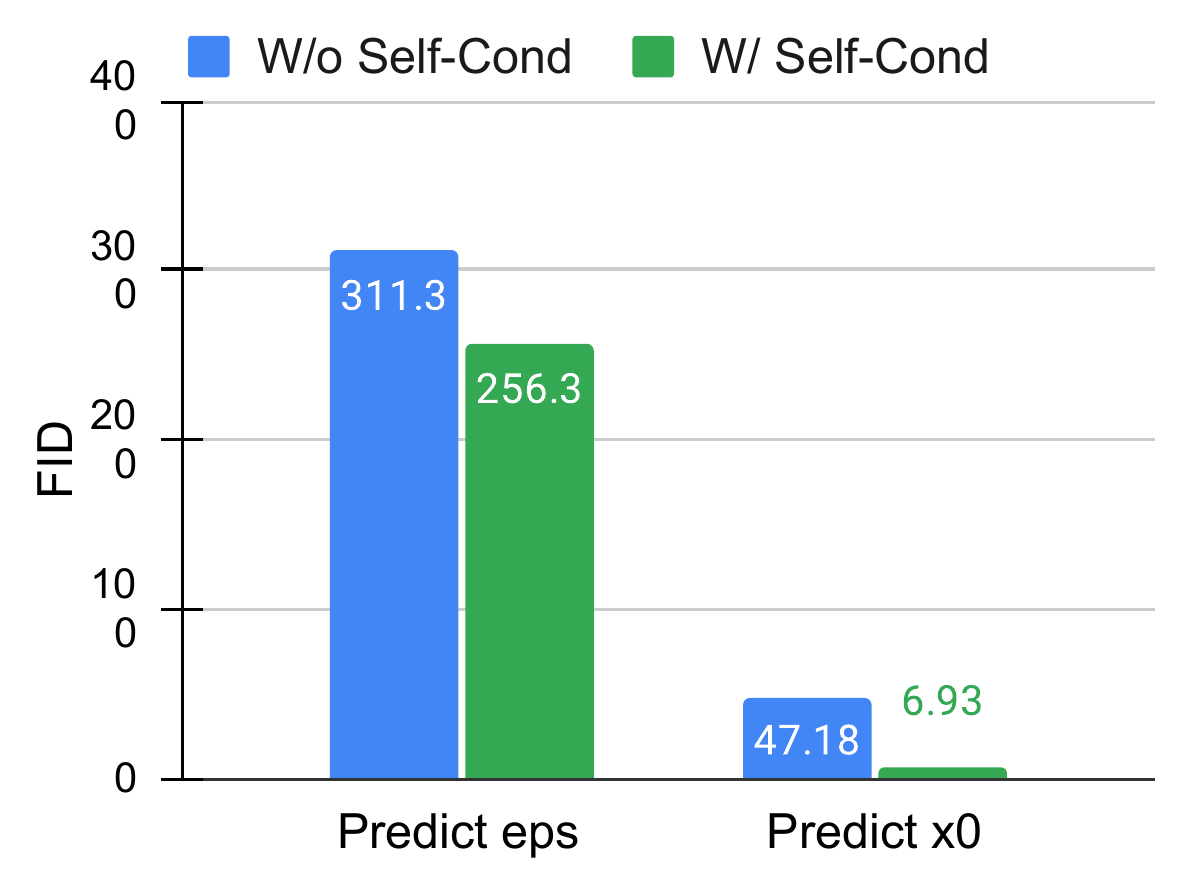}
      \caption{\cifar, \uints.}
    \end{subfigure}
    \begin{subfigure}{0.328\textwidth}
      \centering
      \includegraphics[width=\linewidth]{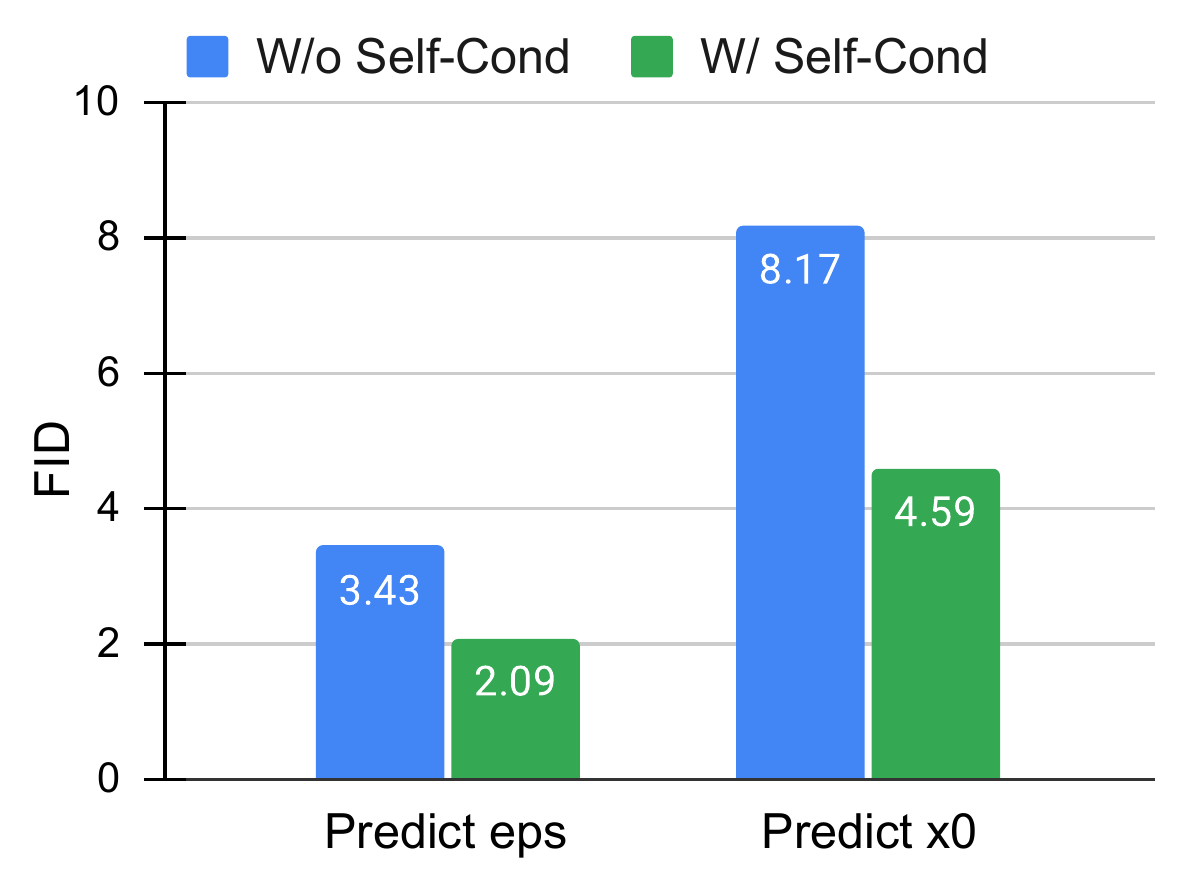}
      \caption{\imagenet.}
    \end{subfigure}
    \vspace{-.5em}
    \caption{\label{fig:ablation_self_cond} Self-conditioning is a generic technique that not only greatly improves Bit Diffusion but also leads to improved results for continuous diffusion models.}
    \vspace{-1em}
\end{figure*}

\textbf{Ablation of Self-Conditioning } Figure~\ref{fig:ablation_self_cond} shows the effectiveness of the Self-Conditioning technique in both Bit Diffusion and continuous diffusion models. Note that the experiments are performed in three settings, namely \cifar with \uint, \cifar with \uints, and \imagenet with continuous pixels, where the only difference for pairs in each setting is whether the Self-Conditioning is used. For \cifar, we find that Self-Conditioning greatly improves the performance across different binary encodings. We also notice that for Bit Diffusion, predicting $\bm x_0$ is much more effective than predicting $\bm \epsilon$. For \imagenet, we find that the proposed Self-Conditioning also leads to improved FIDs for continuous diffusion (i.e., DDPM). Therefore, we conclude that Self-Conditioning by itself is a generic technique that can benefit diffusion models on both continuous and discrete data.

\begin{figure*}[t]
    \centering
    \begin{subfigure}{0.47\textwidth}
      \centering
      \includegraphics[width=\linewidth]{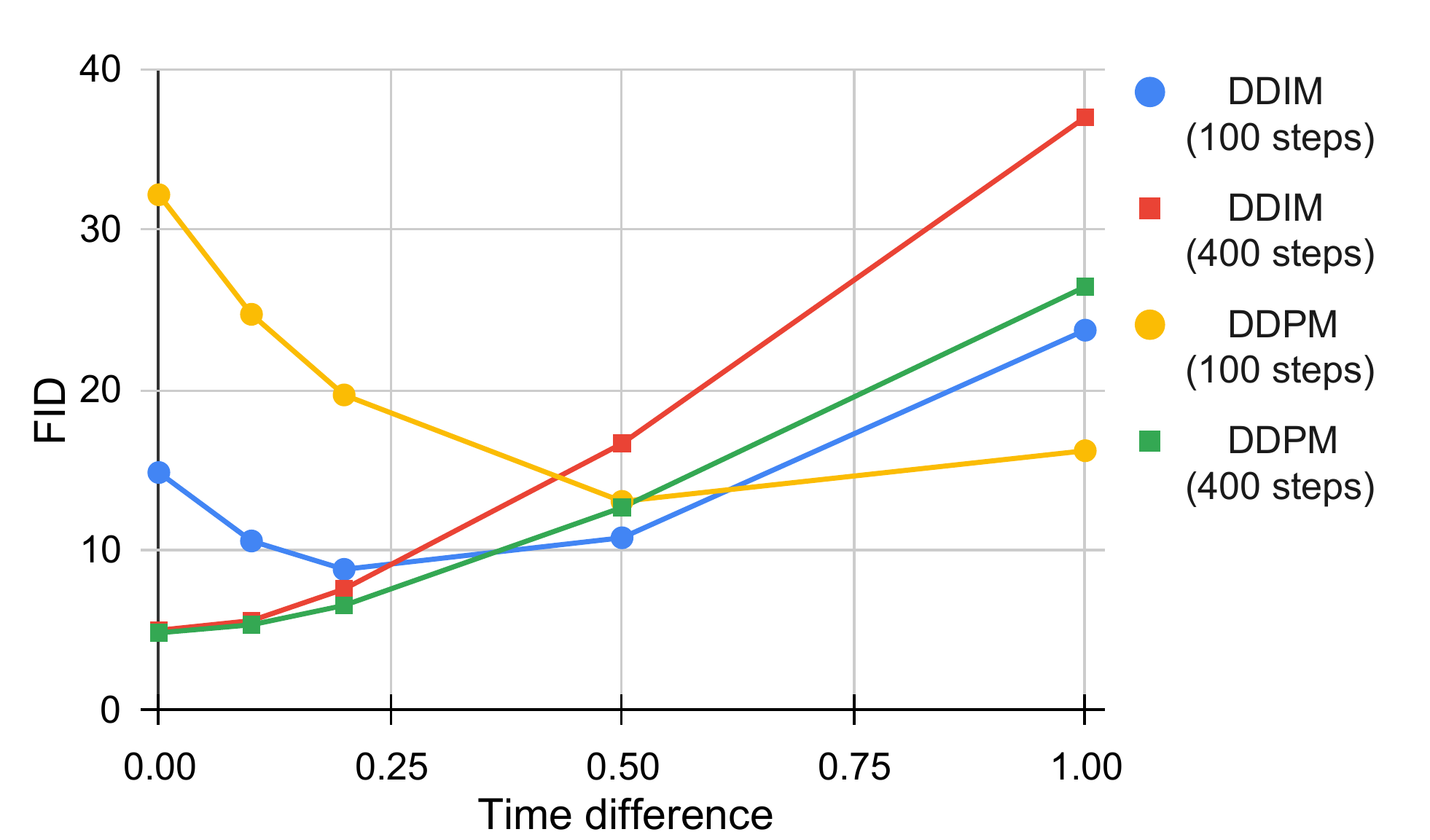}
      \caption{\uint.}
    \end{subfigure}
    ~~~
    \begin{subfigure}{0.47\textwidth}
      \centering
      \includegraphics[width=\linewidth]{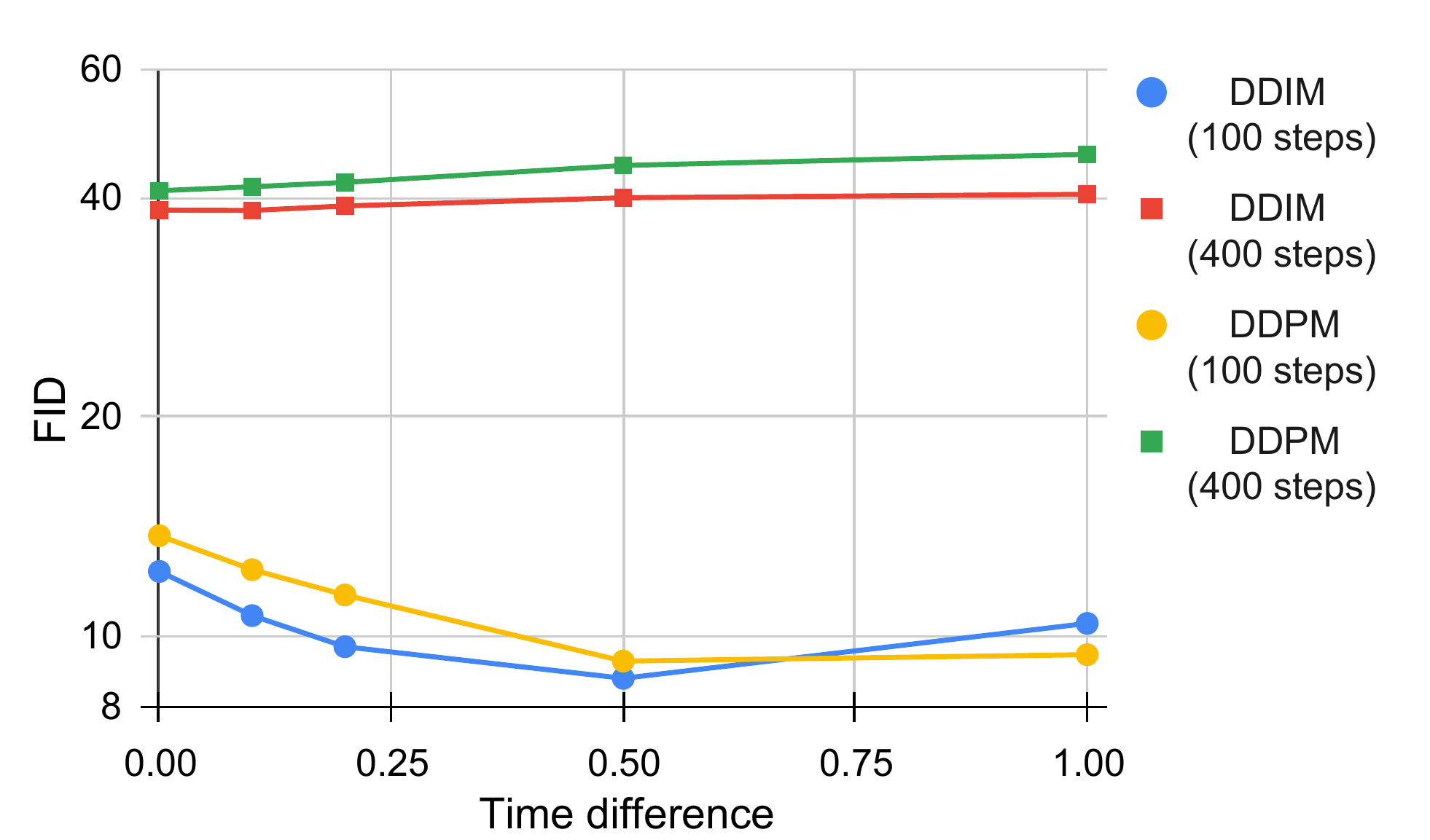}
      \caption{\uints.
}
    \end{subfigure}
    \vspace{-.5em}
    \caption{\label{fig:td_imagenet} Effect of time difference in class-conditional \imagenet. Optimal time difference shrinks to zero as the number of sampling steps increases. For 100 sampling steps, non-zero time difference leads to improved FIDs.}
    \vspace{-2em}
\end{figure*}

\textbf{Ablation of asymmetric time intervals } Figure~\ref{fig:td_imagenet} shows the FID on generated \imagenet samples as we vary the time difference parameter during the sampling process. We find that as the number of steps increases (from $100$ to $400$), the optimal time difference shrinks to $0$. For $100$ steps, a non-zero time difference leads to a significant improvement of FID. We also note that for Bit Diffusion on \uints, using $400$ sampling steps actually leads to a drastically worse sample quality than using $100$ steps. This is related to how the Self-Conditioning is applied and we present alternative Self-Conditioning sampling strategies in the Appendix~\ref{sec:self_cond_sampling}, some of which lead to improved FIDs at a cost of longer sampling time.

\begin{figure*}[ht]
    \centering
    \begin{subfigure}{0.22\textwidth}
    \includegraphics[width=\linewidth]{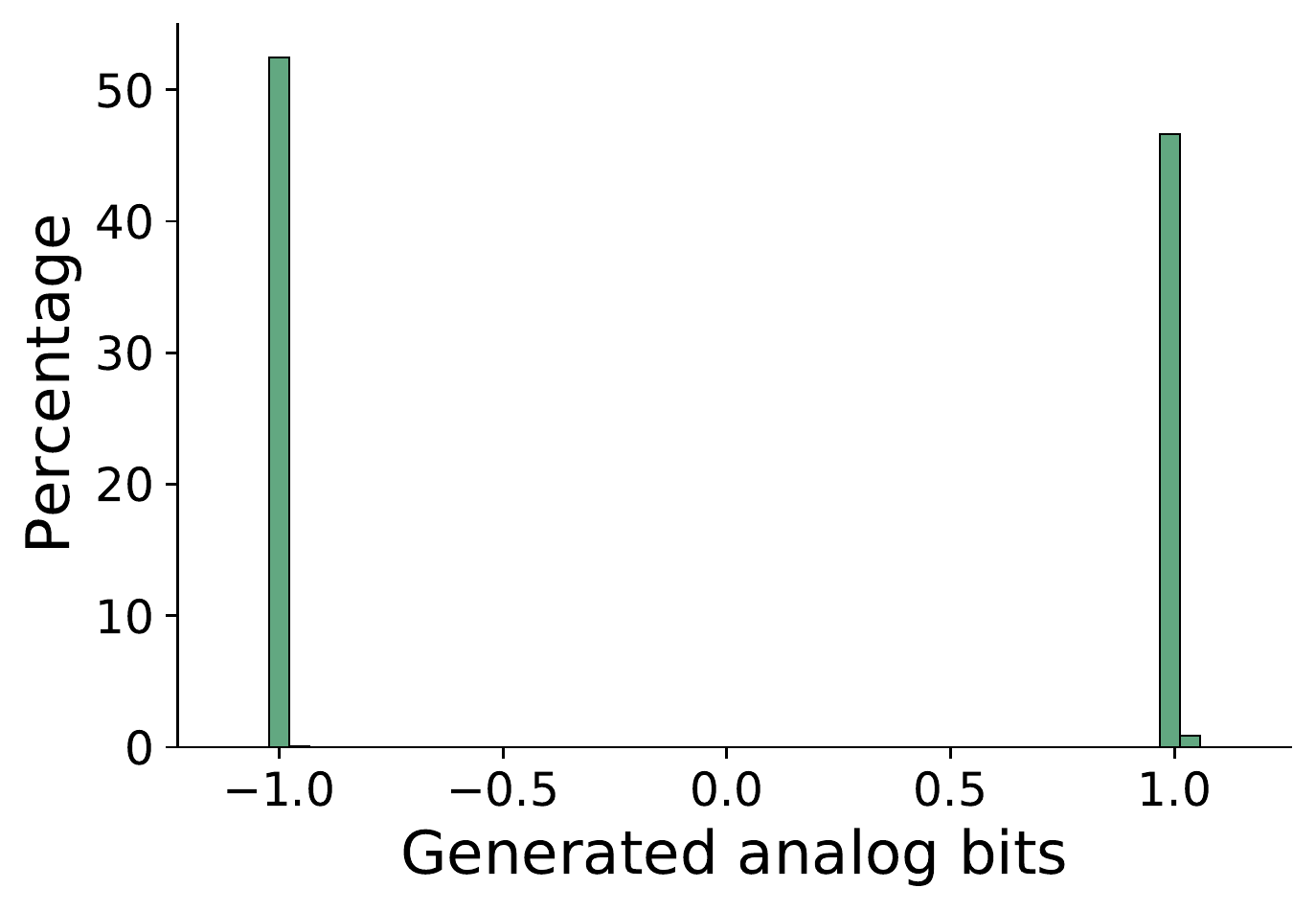}
    \caption{\uint.}
    \end{subfigure}
    ~~~~~~
    \begin{subfigure}{0.22\textwidth}
    \includegraphics[width=\linewidth]{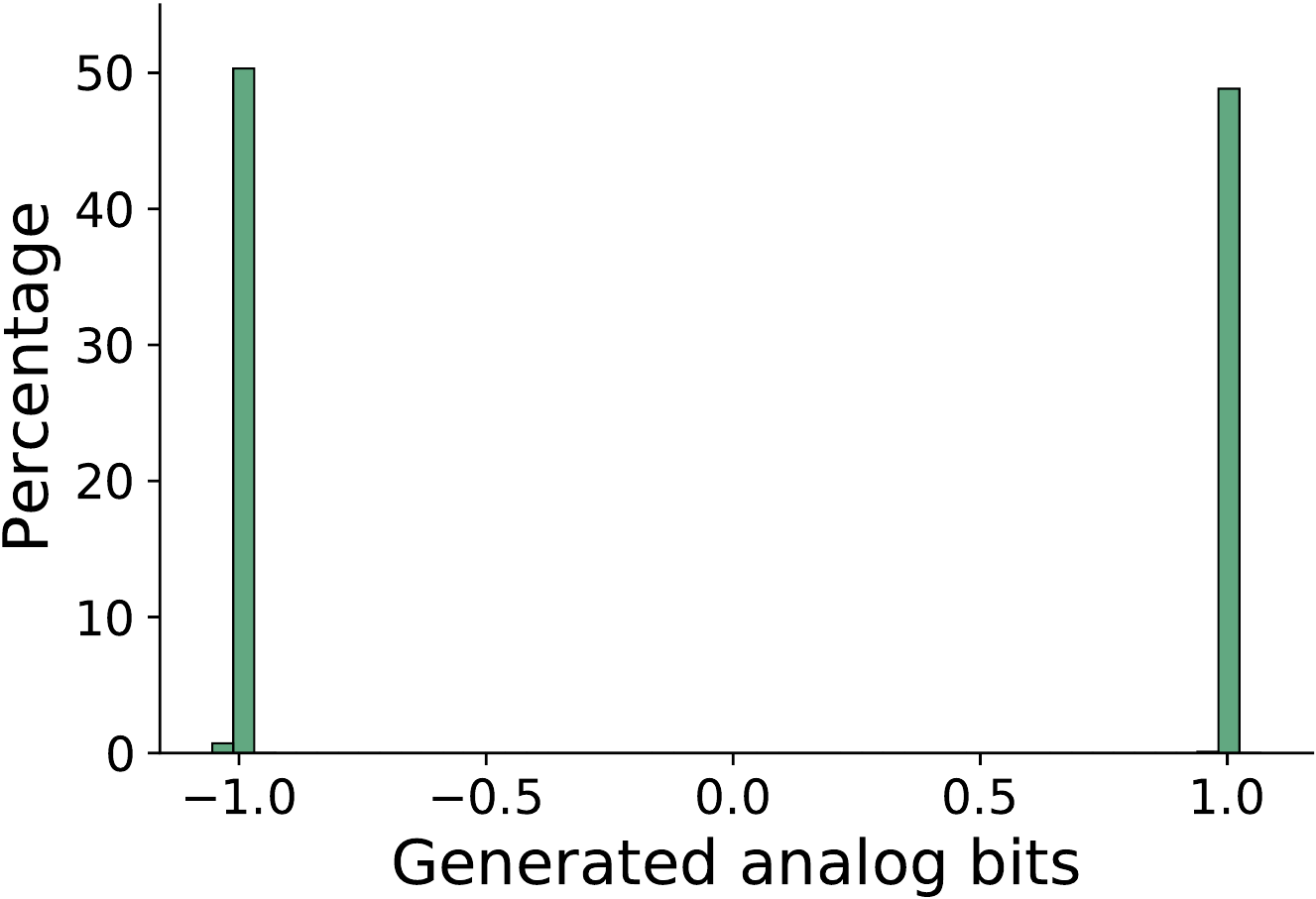}
    \caption{\graycode.}
    \end{subfigure}
    ~~~~~~
    \begin{subfigure}{0.22\textwidth}
    \includegraphics[width=\linewidth]{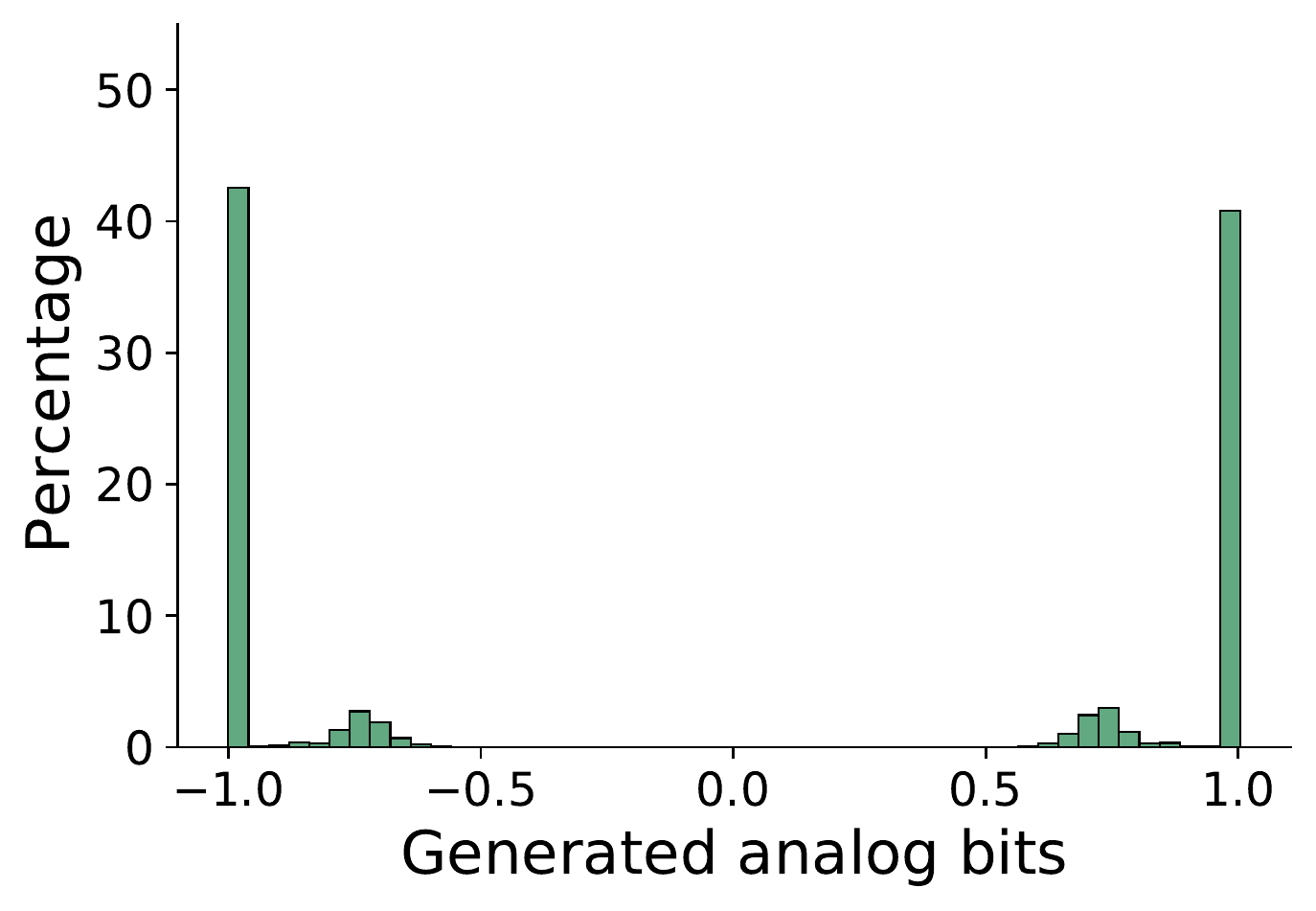}
    \caption{\uints.}
    \end{subfigure}    
    \vspace{-.5em}
    \caption{Histogram distribution (50 bins) of the analog bits from 64 randomly generated \imagenet samples at $\bm{\tilde{x}}_0$, with 100 DDIM steps. Most of the generated analog bits are very concentrated.}
\label{fig:analog_bit_hist}
    \vspace{-1em}
\end{figure*}

\textbf{Concentration of generated analog bits } Figure~\ref{fig:analog_bit_hist} visualizes the distribution of generated analog bits from $64$ generated images on \imagenet. Although there is no hard constraint on the analog bits being binary / bimodal, the generated ones are highly concentrated on two modes, which makes the thresholding / quantization easy and robust. 

\vspace{-.5em}
\subsection{Image Captioning}
\vspace{-.5em}

We compare our Bit Diffusion model with an autoregressive Transformer baseline~\citep{chen2022unified}. As mentioned, both models have similar architectures, with an object detection pretrained~\citep{chen2021pix2seq} image encoder, and a randomly initialized Transformer~\citep{vaswani2017attention} decoder. Table~\ref{tab:caption_result} presents the main comparison. Overall, our model achieves similar performance as the autoregressive model. We find that generally it only needs about 10 steps for the model to achieve good results, despite that there are a total of maximum 960 bits for caption that the model has to model. We find that the asymmetric time intervals play an important role in the final performance of our model, as demonstrated in Table~\ref{tab:caption_td}, especially when sampling steps are fewer.

\begin{table}[h]
    \small
    \centering
    \caption{Image captioning results on MS-COCO dataset with a randomly initialized text decoder.}
    \vspace{-.5em}
    \begin{tabular}{c|ccc}
    \toprule
    Method & BLEU-4 & CIDEr & ROUGE-L \\
    \midrule
    Autoregressive Transformer & $33.9$ & $\mathbf{1.18}$ & $0.57$ \\ 
    \midrule
    Bit Diffusion (5 steps) & $31.5$ & $1.00$ & $0.55$ \\
    Bit Diffusion (10 steps) & $34.5$ & $1.13$ & $0.57$ \\
    Bit Diffusion (20 steps) & $\mathbf{34.7}$ & $1.15$ & $\mathbf{0.58}$ \\
    Bit Diffusion (40 steps) & $34.4$ & $1.15$ & $0.57$ \\
    \bottomrule
    \end{tabular}
    \label{tab:caption_result}
    \vspace{-1em}
\end{table}

\begin{table}[h]
    \small
    \centering
    \caption{Asymmetric time intervals significantly improves the performance of Bit Diffusion.}
    \vspace{-.5em}
    \label{tab:caption_td}
    \begin{tabular}{c|ccccccccc}
    \toprule
     & \multicolumn{9}{c}{Time difference} \\
     & $0.0$ & $1.0$ & $2.0$ & $3.0$ & $4.0$ & $5.0$ & $6.0$ & $7.0$ & $8.0$ \\
    \midrule
    $5$ steps & $17.1$ & $27.8$ & $30.8$ & $31.5$ & $\mathbf{31.6}$ & $31.5$ & $31.5$ & $31.5$ & $\mathbf{31.6}$ \\
    $10$ steps & $17.6$ & $26.3$ & $30.7$ & $32.6$ & $33.4$ & $34.0$ & $34.3$ & $34.5$ & $\mathbf{34.6}$ \\
    $20$ steps & $20.0$ & $27.9$ & $30.6$ & $32.0$ & $32.3$ & $33.9$ & $34.4$ & $\mathbf{34.7}$ & $34.5$ \\
    $40$ steps & $20.7$ & $27.5$ & $30.7$ & $32.2$ & $32.9$ & $33.2$ & $33.8$ & $\mathbf{34.4}$ & $\mathbf{34.4}$ \\
    \bottomrule
    \end{tabular}
\end{table}
Table~\ref{tab:caption_cases} provides some generated samples of our model when different inference steps are used. The model makes mistakes when the sampling steps are too few, and the mistakes may not always be interpretable due to that the model directly predicts the bits behind the tokenized word pieces and a small difference in bits can lead to total different words.
{\def\arraystretch{1.7}
\begin{table}[!th]
    \centering
    \small
    \caption{Generated image captions under different number of sampling steps.}
    \vspace{-.5em}
    \begin{tabular}{cl}
    \toprule
\multirow{4}{*}{\includegraphics[height=0.15\linewidth]{figures/162543_orig.pdf}}
      & Steps=$1$: A group of diets in for foraa \\
      & Steps=$2$: A group of elephants in in a\textbackslash ufffd. \\
      & Steps=$5$: A group of elephants standing in a Irish. \\
      & Steps=$10$: A group of elephants standing by a fence.\\
      \hline
      \multirow{4}{*}{\includegraphics[height=0.155\linewidth]{figures/230983_orig.pdf}}
      & Steps=$1$: A says man  fora You  a\textbackslash u0000. \\
      & Steps=$2$: A says man a skateboard' aa. \\
      & Steps=$5$: A man on a skateboard on a skateboard. \\
      & Steps=$10$: A man is doing a trick on a skateboard. \\
\bottomrule
    \end{tabular}
    \label{tab:caption_cases}
    \vspace{-1em}
\end{table}
} \vspace{-.5em}
\section{Related Work}
\vspace{-.5em}

\textbf{Autoregressive models for discrete data } Autoregressive models have demonstrated state-of-the-art results when it comes to generating discrete data. In particular, text generation, or language modeling, is dominated by autoregressive approaches~\citep{sutskever2014sequence,brown2020language,chowdhery2022palm}. Autoregressive models are also applied to discrete/categorical image generation~\citep{van2016conditional,salimans2017pixelcnn++,parmar2018image,child2019generating,roy2021efficient,jun2020distribution,chen2020generative}, where they work well on small image resolutions. However, the computation cost and memory requirement increase drastically (typically in a quadratic relation) as the size of sequence or the image resolution increase, so it becomes very challenging to scale these approaches to data with large dimensions.

\textbf{Diffusion models for discrete data } State-of-the-art diffusion models~\citep{dhariwal2021diffusion,ho2022cascaded,nichol2021glide,ramesh2022hierarchical,saharia2022photorealistic} cannot generate discrete or categorical data. Existing extensions of these continuous diffusion models to discrete data are based on both discrete data space and state space~\citep{sohl2015deep,hoogeboom2021argmax,austin2021structured,campbell2022continuous}. Compared to discrete state space, continuous state space is more flexible and potentially more efficient. Our approach is also compatible with both discrete and continuous time, and does not require re-formulation of existing continuous models, thus it is simpler and can potentially be plugged into a broader family of generative models.

Another line of discrete diffusion models is based on the embedding of discrete data~\citep{li2022diffusionlm}. One can also consider our binary encoding with analog bits as a simple fixed encoder, and the decoding / quantization of bimodal analog bits is easy and robust via a simple thresholding operation. In contrast, the quantization of real numbers in generated continuous embedding vectors may contain multiple modes per dimension, leading to potential difficulty in thresholding/quantization.

\textbf{Normalizing Flows for discrete data } Normalizing Flows~\citep{rezende2015variational,real_nvp,glow} are a powerful family of generative models for high-dimensional continuous distributions based on some invertible mapping. However, straightforward application of flow-based models on categorical data is limited due to the inherent challenges on discrete support.
Discrete flows~\citep{discrete_flow,idf,discrete_denoising_flow} introduce invertible transformations of random variables in discrete space without the need of computing the log-determinant of Jacobian.
Other works~\citep{lippe2021categorical,hoogeboom2021argmax,tan2021learning} introduce various embedding methods for transforming discrete data into continuous space with disjoint support, which can be interpreted as a variational inference problem~\citep{theis2015note} with different dequantization distribution families.
Several works~\citep{kingma2016improved,ziegler2019latent,zhang2020perceptual} also explore normalizing flows on discrete data under the Variational Autoencoders~\citep{kingma2013auto} framework by enriching the prior. Compared to our diffusion-based approach, these models suffer from strict invertible restrictions on network architecture, thus limiting their capacity.

\textbf{Other generative models for discrete data }
Other generative models, such as Varational Autoencoders~(VAE)~\citep{kingma2013auto}, Generateive Adversarial Networks~(GAN)~\citep{goodfellow2014generative,yu2017seqgan,maligan,hjelm2017boundary,maskgan} have also been applied to generate discrete data. These methods have not yet achieved the level of performance as autoregressive models on tasks such as discrete image generation or text generation, in terms of sample quality or data likelihood.
Potentially, the proposed analog bits can also be applied to these continuous generative models, by having the networks directly model and generate analog bits, but it is not explored in this work.

\textbf{Other related work } The proposed Self-Conditioning technique shares some similarities with self-modulation in GANs~\citep{chen2018self} (where the earlier latent state can directly modulate the later latent states) and SUNDAE~\citep{savinov2021step} (where an inference step is incorporated for denoising).
 \vspace{-.5em}
\section{Conclusion}
\vspace{-.5em}

We introduce a simple and generic technique that enables continuous state diffusion models to generate discrete data. The main idea is to encode discrete or categorical data into bits and then model these bits as real numbers that we call analog bits. 
We also propose two simple techniques, namely Self-Conditioning (i.e., condition the diffusion models directly on their previously generated samples) and Asymmetric Time Intervals, that lead to improved sample quality.
We demonstrate that our approach leads to state-of-the-art results in discrete / categorical image generation, beating the best autoregressive model. In an image-conditional text generation task on MS-COCO dataset, we also achieve competitive results compared to autoregressive models.
One limitation of our approach, similar to other existing diffusion models, is that they still require a significant number of inference steps for generating good (image) samples. However, we expect that future improvements from diffusion models for continuous data can also transfer to discrete data using analog bits.

 \newpage

\section*{Acknowledgements} We would like to thank Priyank Jaini, Kevin Swersky for providing helpful feedback to our draft. Our implementation is partially based on the Pix2Seq codebase, and we thank Lala Li, Saurabh Saxena, for their contributions to the Pix2Seq codebase.

{\small
\bibliography{content/ref}
\bibliographystyle{iclr2023_conference}
}

\newpage
\appendix
\section{More Details of Algorithm~\ref{alg:train} and~\ref{alg:sample}}
\label{sec:alg_details}
Algorithm~\ref{alg:encode_decode} and~\ref{alg:ddim_ddpm} provide more detailed implementations of functions in Algorithm~\ref{alg:train} and~\ref{alg:sample}.
\begin{minipage}[t]{0.8\textwidth}
\centering
\begin{algorithm}[H]
\small
\caption{\small Binary encoding and decoding algorithms (in Tensorflow).
}
\label{alg:encode_decode}
\definecolor{codeblue}{rgb}{0.25,0.5,0.5}
\definecolor{codekw}{rgb}{0.85, 0.18, 0.50}
\lstset{
  backgroundcolor=\color{white},
  basicstyle=\fontsize{7.5pt}{7.5pt}\ttfamily\selectfont,
  columns=fullflexible,
  breaklines=true,
  captionpos=b,
  commentstyle=\fontsize{7.5pt}{7.5pt}\color{codeblue},
  keywordstyle=\fontsize{7.5pt}{7.5pt}\color{codekw},
  escapechar={|}, 
}
\begin{lstlisting}[language=python]
import tensorflow as tf

def int2bit(x, n=8):
  # Convert integers into the corresponding binary bits.
  x = tf.bitwise.right_shift(tf.expand_dims(x, -1), tf.range(n))
  x = tf.math.mod(x, 2)
  return x
    
def bit2int(x):
  # Convert binary bits into the corresponding integers.
  x = tf.cast(x, tf.int32)
  n = x.shape[-1]
  x = tf.math.reduce_sum(x * (2 ** tf.range(n)), -1)
  return x
\end{lstlisting}
\end{algorithm}
\end{minipage}

\begin{minipage}[t]{0.8\textwidth}
\centering
\begin{algorithm}[H]
\small
\caption{\small $x_t$ estimation with DDIM / DDPM updating rules.
}
\label{alg:ddim_ddpm}
\definecolor{codeblue}{rgb}{0.25,0.5,0.5}
\definecolor{codekw}{rgb}{0.85, 0.18, 0.50}
\lstset{
  backgroundcolor=\color{white},
  basicstyle=\fontsize{7.5pt}{7.5pt}\ttfamily\selectfont,
  columns=fullflexible,
  breaklines=true,
  captionpos=b,
  commentstyle=\fontsize{7.5pt}{7.5pt}\color{codeblue},
  keywordstyle=\fontsize{7.5pt}{7.5pt}\color{codekw},
  escapechar={|}, 
}
\begin{lstlisting}[language=python]
def gamma(t, ns=0.0002, ds=0.00025):
  # A scheduling function based on cosine function.
  return numpy.cos(((t + ns) / (1 + ds)) * numpy.pi / 2)**2

def ddim_step(x_t, x_pred, t_now, t_next):
  # Estimate x at t_next with DDIM updating rule.
  |$\gamma_{\text{now}}$| = gamma(t_now)
  |$\gamma_{\text{next}}$| = gamma(t_next)
  x_pred = clip(x_pred, -scale, scale)
  eps =|$\frac{1}{\sqrt{1-\gamma_\text{now}}}$| * (x_t - |$\sqrt{\gamma_\text{now}}$| * x_pred)
  x_next = |$\sqrt{\gamma_\text{next}}$| * x_pred + |$\sqrt{1-\gamma_\text{next}}$| * eps
  return x_next
  
def ddpm_step(x_t, x_pred, t_now, t_next):
  # Estimate x at t_next with DDPM updating rule.
  |$\gamma_{\text{now}}$| = gamma(t_now)
  |$\alpha_{\text{now}}$| = gamma(t_now) / gamma(t_next)
  |$\sigma_{\text{now}}$| = sqrt(1 - |$\alpha_{\text{now}}$|)
  z = normal(mean=0, std=1)
  x_pred = clip(x_pred, -scale, scale)
  eps =|$\frac{1}{\sqrt{1-\gamma_\text{now}}}$| * (x_t - |$\sqrt{\gamma_\text{now}}$| * x_pred)
  x_next = |$\frac{1}{\sqrt{\alpha_{\text{now}}}}$| * (x_t - |$\frac{1 - \alpha_{\text{now}}}{\sqrt{1 - \gamma_{\text{now}}}}$| * eps) + |$\sigma_{\text{now}}$| * z
  return x_next
\end{lstlisting}
\end{algorithm}
\end{minipage}

\section{Alternative Binary Encoding and Loss Functions}
\label{sec:alt_loss}

\subsection{Analog Bits based on One-Hot Encoding}

An alternative binary encoding to the base-2 encoding of the discrete data used in the main paper, is the one-hot encoding, where a discrete variable is represented as a vector whose length is the same as the vocabulary size $K$, with a single slot being $1$ and the rest being $0$. The resulting one-hot vector can be similarly treated as analog bits and modeled by continuous state diffusion models. To obtain discrete variables corresponding to the generated analog bits, we use an $\argmax$ operation over all candidate categories, instead of the thresholding operation in base-2 analog bits.
Note that the one-hot encoding requires $K$ bits, which is less efficient compared to base-2 encoding that only requires  $\lceil\log_2 K\rceil$ bits, especially for large $K$. \footnote{Although, one can reduce the vocabulary size $K$ by using sub-tokenization (e.g., subword~\citep{sennrich2015neural}), or learned discrete codes~\citep{chen2018learning,chen2020differentiable}.}

\subsection{Sigmoid Cross Entropy Loss}

As we use $\ell_2$ loss by default for its simplicity and compatibility with continuous diffusion models. The proposed Bit Diffusion models can work with other loss functions too. Since the analog bits are bimodal,
we can use the following sigmoid cross entropy loss:
$$
\mathcal{L}_{\bm x_0, \bm x_t, t} = \log\sigma(\bm x_0 f(\bm x_t, t)),
$$
where we assume $\bm x_0 \in \{-1, 1\}^n$, and $\sigma$ is a sigmoid function. During the sampling process, we use $2\sigma(f(\bm x_t, t))-1$ as the output of denoising network.

\subsection{Softmax Cross Entropy Loss}

For one-hot analog bits, one could also add a softmax activation function for the output of denosing network $f$, and use the following softmax cross entropy loss:
$$
\mathcal{L}_{\bm x_0, \bm x_t, t} = \bm x_0 \log \text{softmax}(f(\bm x_t, t)),
$$
where we assume $\bm x_0 \in \{0, 1\}^n$ which is the one-hot representation.

\subsection{Preliminary Experiments}

Table~\ref{tab:loss_vars} presents FIDs of Bit Diffusion models with different types of analog bits and loss functions on unconditional \cifar. Note that it is possible some of these results can be improved by more tuning of hyper-parameters or tweaks of the network, but we do not focus on them in this work.
\begin{table}[h]
    \centering
    \small
    \caption{FIDs of Bit Diffusion models with different types of analog bits and loss functions on unconditional \cifar.}
    \begin{tabular}{c|ccc}
    \toprule
     & $\ell_2$ loss & Logistic loss & Softmax loss \\
    \midrule
    \onehot & $46.32$ & $26.82$ & $29.49$ \\
    \uint & $3.48$ & $3.53$ & - \\
    \graycode & $3.86$ & $3.71$ & - \\
    \uints & $6.93$ & $49.29$ & - \\
    \bottomrule
    \end{tabular}
    \label{tab:loss_vars}
\end{table}

\section{On Binary Encoding of Pixels: \uint, \graycode, \uints}
\label{sec:pixel_binary_code}

In the main paper, we describe three different types of binary encodings of pixels.
Here we provide additional detail on how we generate \uints: we first apply a random permutation to 256 sub-pixel values, and then assign the binary binary bits of permuted integers to the non-permuted integers. For example, assume 0 is mapped to 228 after the permutation, the analog bits of 0 would be the binary bits of 228. The random permutation is generated by \texttt{numpy.random.seed(42); numpy.random.shuffle(numpy.arange(256))}.

\begin{figure*}[h]
    \centering
    \begin{subfigure}{0.31\textwidth}
      \centering
      \includegraphics[width=\linewidth]{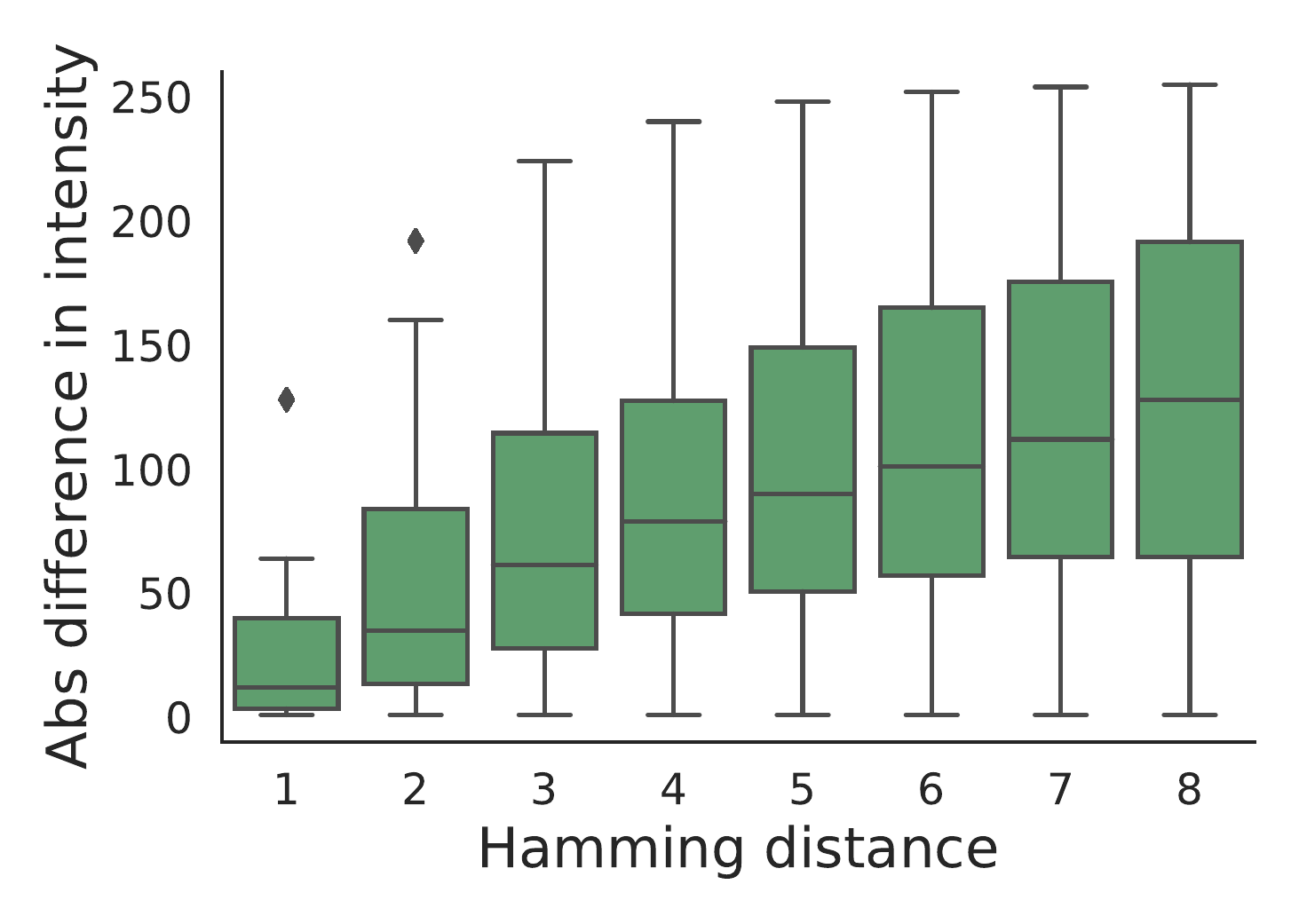}
      \caption{\uint.}
    \end{subfigure}
    \begin{subfigure}{0.31\textwidth}
      \centering
      \includegraphics[width=\linewidth]{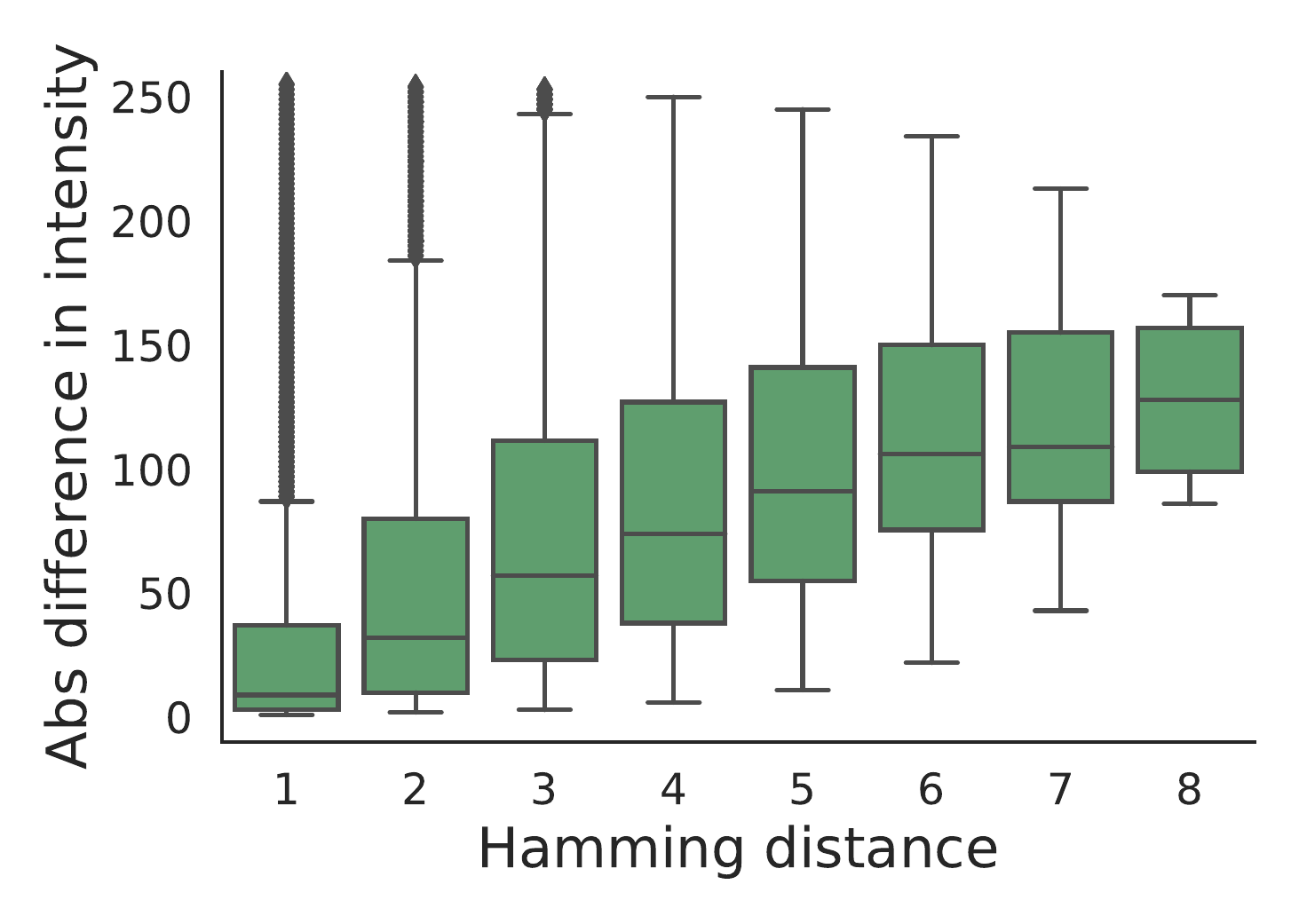}
      \caption{\graycode.}
    \end{subfigure}
    \begin{subfigure}{0.31\textwidth}
      \centering
      \includegraphics[width=\linewidth]{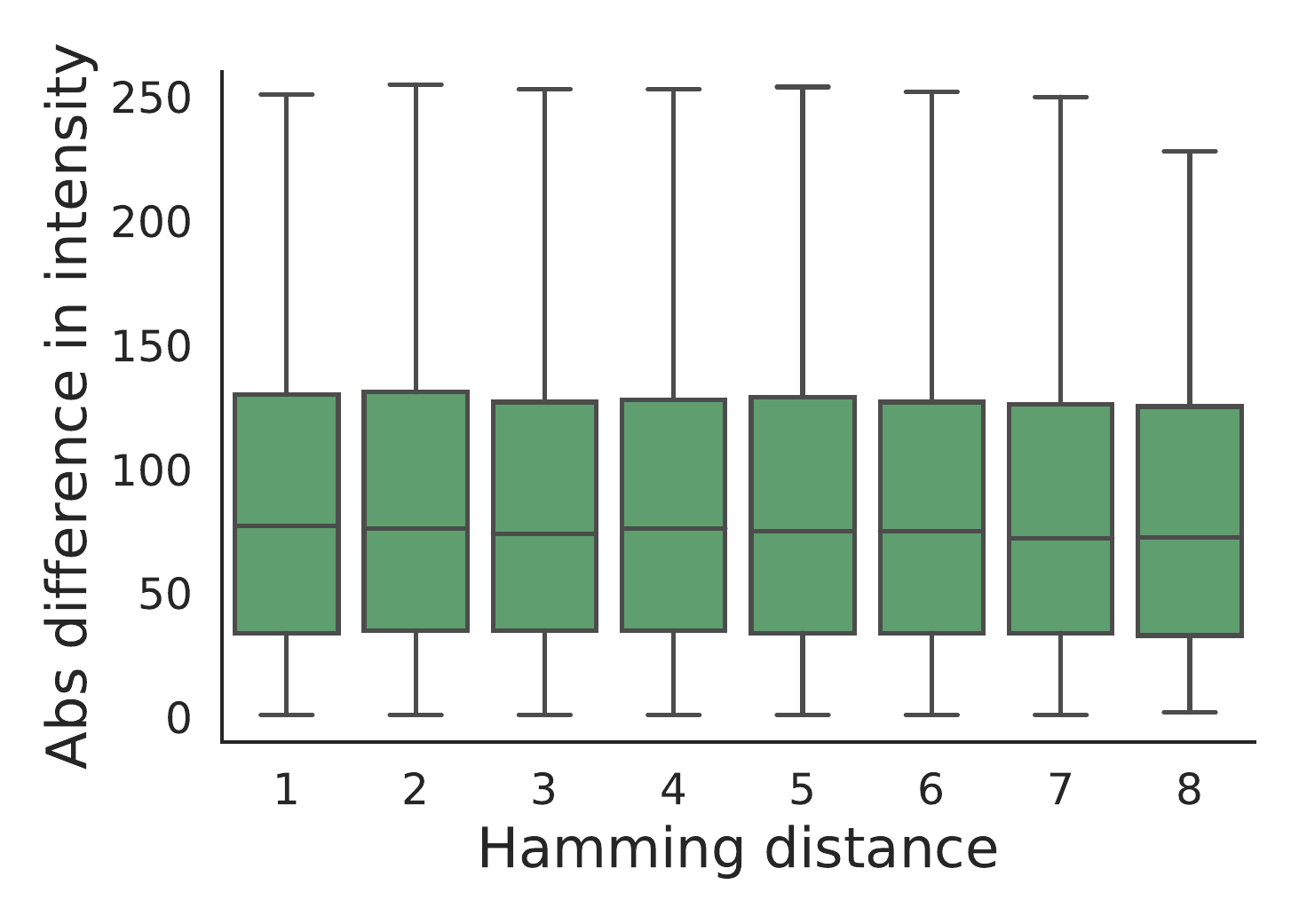}
      \caption{\uints.}
    \end{subfigure}
    \caption{\label{fig:code_dist} 
    Correlation between (absolute) difference in subpixel intensity and the Hamming distance of the corresponding binary bits.
}
\end{figure*}

Figure~\ref{fig:code_dist} show the correlation between Hamming distance of three different binary encodings we use and the (absolute) difference of sub-pixel intensity. This is done by taking every pair of subpixel integers (in $[0, 256)$), compute their absolute difference, as well as the Hamming distance between the corresponding binary bits. We find that both \uint and \graycode exhibit partial correlation between the two quantities (with different correlation patterns), meaning that these codes partially contain the order information about the original sub-pixel intensity. However, \uints exhibits no correlation between hamming distance and sub-pixel intensity, indicating the order information is fully removed, thus can be considered as categorical data.

\section{A Toy Example on Continuous Modeling of Discrete Variables}
\label{sec:toy_illustration}

\begin{figure*}[h]
    \centering
    \begin{subfigure}{0.35\textwidth}
      \centering
      \includegraphics[width=\linewidth]{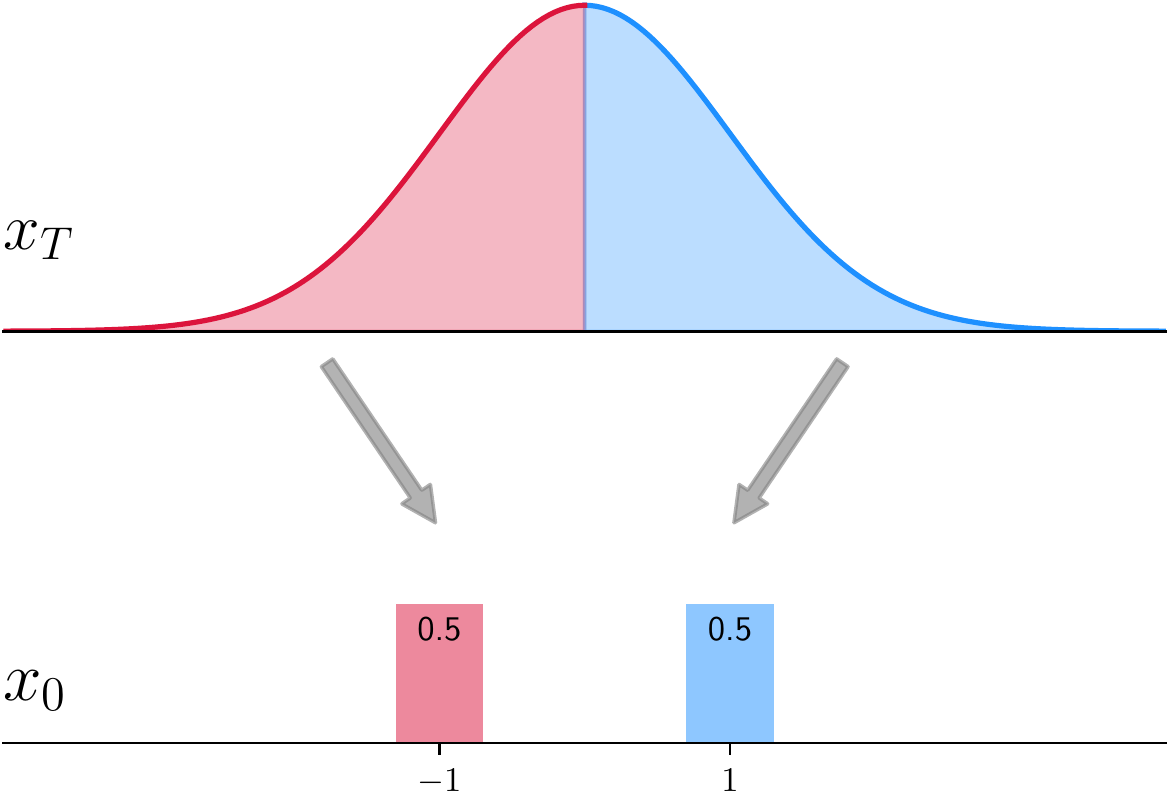}
      \caption{Mapping from $\mathcal{N}(0, 1)$ to Bernoulli distribution with $P(x_0 = 1) = 0.5$.}
    \end{subfigure}
    ~~~~~~~~~~~~~~~
    \begin{subfigure}{0.35\textwidth}
      \centering
      \includegraphics[width=\linewidth]{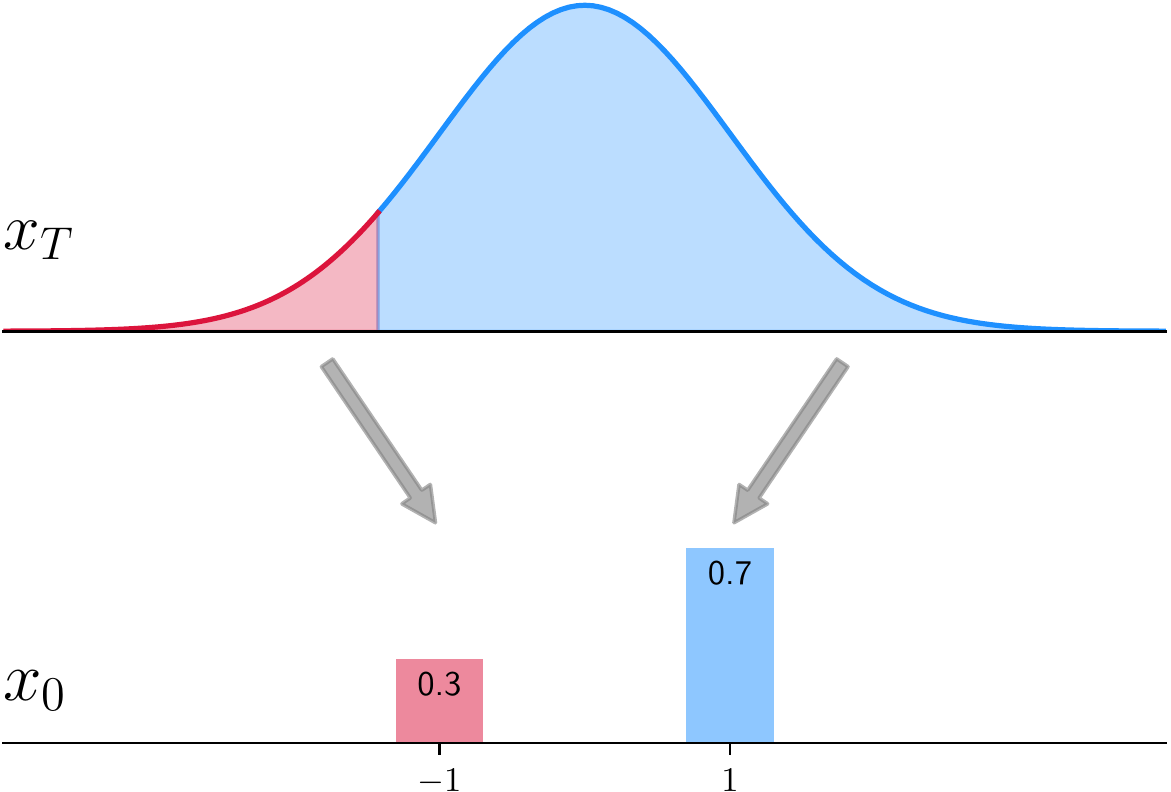}
      \caption{Mapping from $\mathcal{N}(0, 1)$ to Bernoulli distribution with $P(x_0 = 1) = 0.7$.}
    \end{subfigure}
    \caption{\label{fig:1d_toy} A toy example on continuous modeling of discrete variables.}
\end{figure*}

An intuitive toy example of how a continuous generative model can generate binary data is given in Figure~\ref{fig:1d_toy}, where a mapping from prior distribution at $\bm x_T$ to data distribution at $\bm x_0$ is shown. With a deterministic sampler (such as DDIM), it is straight-forward how they can represent any Bernoulli distribution by dividing the prior into two regions of probability densities corresponding to the Bernoulli distribution. For stochastic samplers, they can achieve a similar effect but the mapping from noise to data is stochastic. For an arbitrary discrete variable, represented as m-dimensional Bernoulli distribution, the mapping from continuous noise distribution to the target Bernoulli distribution also exists but it is more complicated (and difficult to visualize).

\section{On Other Samplers for Continuous Diffusion Models}

As our models are based on continuous diffusion models, in theory our models are able to incorporate faster samplers. To this end, we conduct preliminary exploration of using DPM-Solver~\citep{lu2022dpm} for sampling some of our models. 

We find that DPM-Solver provides a boost to diffusion models based on analog bits, similar to what it is able to do for continuous data. This shows a potential of our model enjoying faster continuous sampler while other baselines (e.g., D3PM) may not be able to do due to their use of discrete states. Table~\ref{tab:cont_samplers} below shows the FID scores of bit diffusion models on ImageNet-64x64 under different binary encoding schemes. We find that the DPM-Solver  is able to provide a significant reduction in function evaluations for bit diffusion on discrete/categorical data (with 30 NFEs it gets comparable FIDs as 100 NFEs of DDIM), similar to that in continuous diffusion models.

\begin{table}[h!]
\small
\centering
\caption{\label{tab:cont_samplers}Comparison of Continuous Diffusion Samplers. FIDs on ImageNet 64$\times$64 shown below.}
\begin{tabular}{c|c|c|c}
\toprule
{Samplers} & {UINT8} & {Gray Code} & {UINT8 (RAND)} \\
\midrule
DDIM @ 1000 NFE & 5.51 & 5.14 & 58.45 \\
DDPM @ 1000 NFE & 7.71 & 6.91 & 64.08 \\
DDIM @ 400 NFE & 5.00 & 5.52 & 38.44 \\
DDPM @ 400 NFE & 4.84 & 5.37 & 40.91 \\
DDIM @ 100 NFE & 8.80 & 11.31 & 8.76 \\
DDPM @ 100 NFE & 13.04 & 12.77 & 9.25 \\
\midrule
DPM-Solver @ 30 NFE & 7.85 & 9.64 & 10.39 \\
DPM-Solver @ 50 NFE & 6.46 & 7.61 & 10.96 \\
\bottomrule
\end{tabular}
\end{table}

Furthermore, we also find that self-conditioning continues to provide a boost with DPM-solver. For example, the table~\ref{tab:sc_dpm} shows FID scores of diffusion models on ImageNet 64x64 (continuous rgb values). And we find that the self-conditioning consistently improves the performance of DPM-Solver with fixed number of function evaluations.

\begin{table}[h!]
\small
\centering
\caption{The effect of Self-Conditioning for sampling with DPM-Solver. FIDs on ImageNet 64$\times$64 shown below.}
\begin{tabular}{c|c|c}
\toprule
{Model} & {DPM-Solver @ 20 NFE} & {DPM-Solver @ 30 NFE} \\
\midrule
$\epsilon$ prediction, w/o self-conditioning & 6.10 & 5.58 \\
$\epsilon$ prediction, w/ self-conditioning & 4.24 & 4.15\\
\midrule
$x_0$ prediction, w/o self-conditioning & 12.13 & 11.05 \\
$x_0$ prediction, w/ self-conditioning & 6.94 & 6.43 \\
\bottomrule
\end{tabular}
\label{tab:sc_dpm}
\end{table}

 \section{Extra Random Samples on \cifar and \imagenet}
\label{sec:extra_samples}

\begin{figure*}[h]
    \centering
    \begin{subfigure}{0.4\textwidth}
      \centering
      \includegraphics[width=\linewidth]{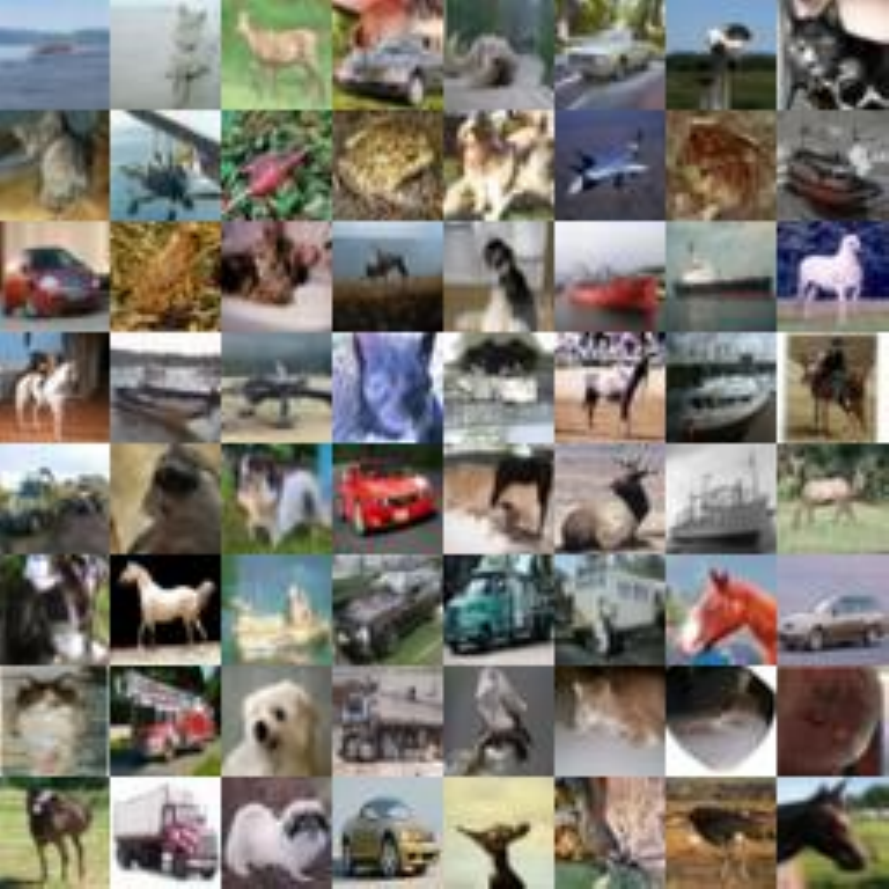}
      \caption{DDPM on continuous pixels (FID=$3.14$)}
    \end{subfigure}
    ~~~~
    \begin{subfigure}{0.4\textwidth}
      \centering
      \includegraphics[width=\linewidth]{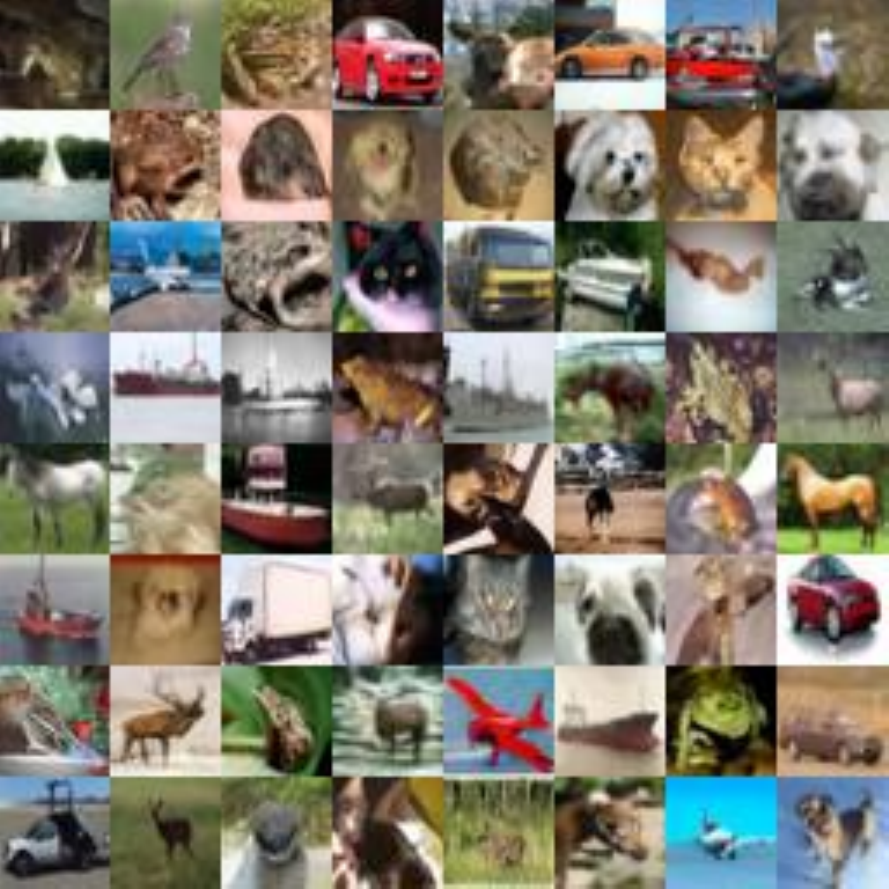}
      \caption{Bit Diffusion on \uint (FID=$3.71$)}
    \end{subfigure}
    \begin{subfigure}{0.4\textwidth}
      \centering
      \includegraphics[width=\linewidth]{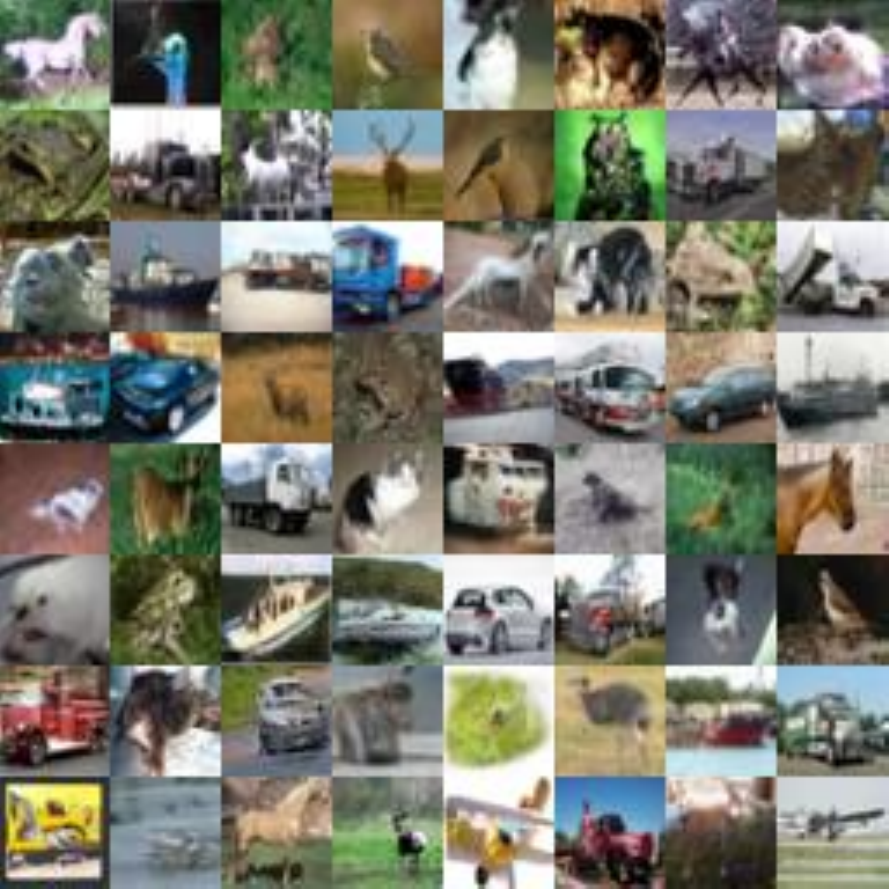}
      \caption{Bit Diffusion on \graycode (FID=$3.88$)}
    \end{subfigure}
    ~~~~
    \begin{subfigure}{0.4\textwidth}
      \centering
      \includegraphics[width=\linewidth]{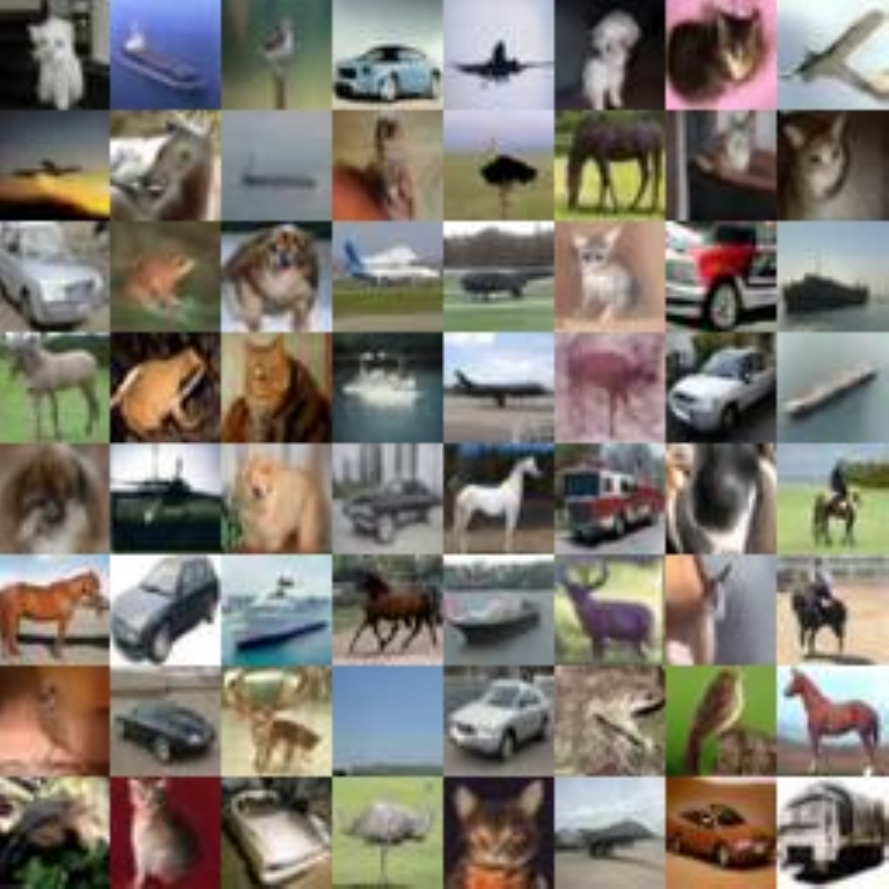}
      \caption{Bit Diff. on \uints (FID=$6.93$)}
    \end{subfigure}
    \caption{\label{fig:rand_samples_cifar10} Random samples from unconditional models trained on \cifar.
    (a) is for continuous image generation, (b), (c), and (d) are for discrete/categorial image generation.
    }
\end{figure*}

\begin{figure*}[h]
    \centering
    \begin{subfigure}{0.49\textwidth}
      \centering
      \includegraphics[width=\linewidth]{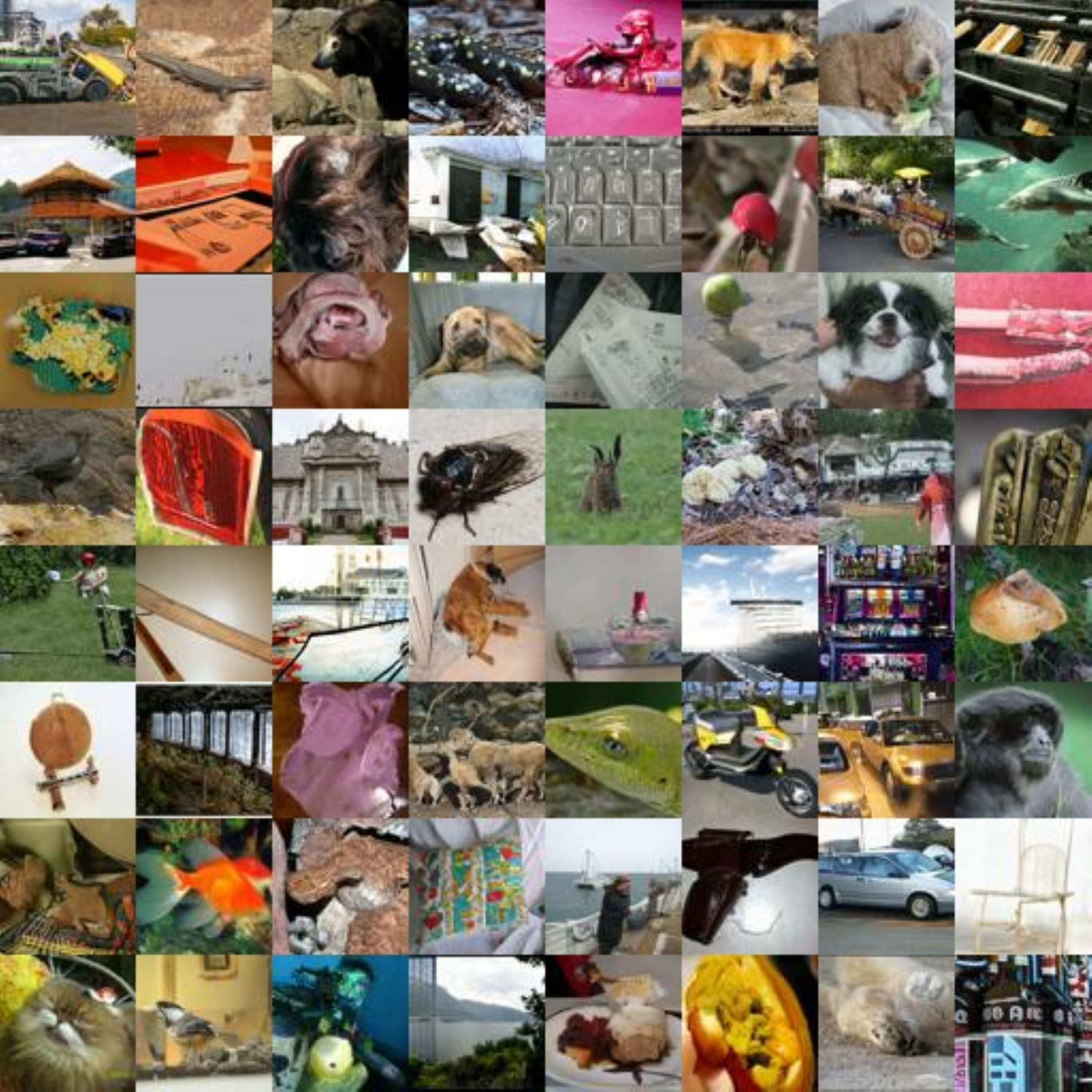}
      \caption{DDPM on continuous pixels (FID=$3.43$)}
    \end{subfigure}
    \begin{subfigure}{0.49\textwidth}
      \centering
      \includegraphics[width=\linewidth]{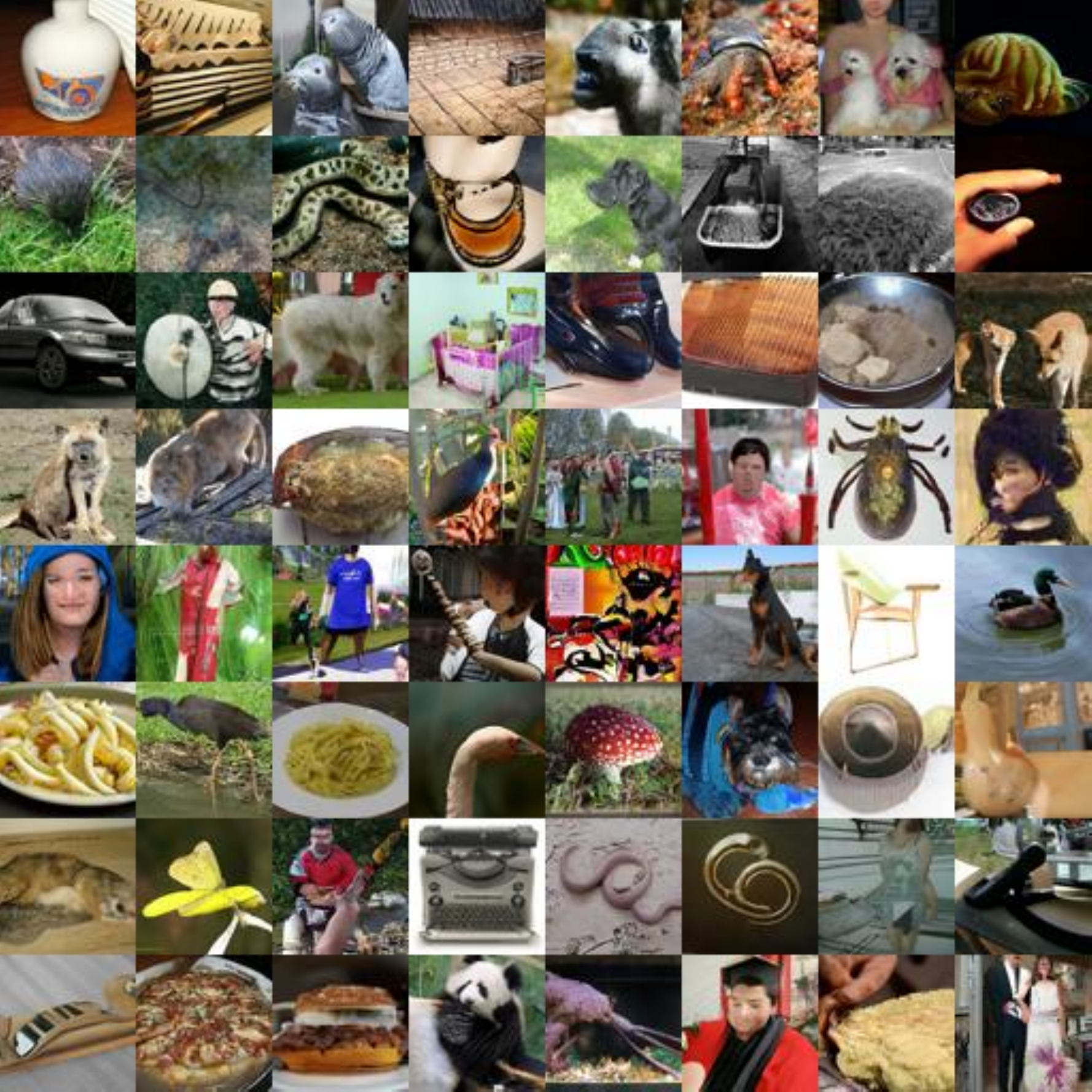}
      \caption{Bit Diffusion on \uint. (FID=$4.84$)}
    \end{subfigure}
    \begin{subfigure}{0.49\textwidth}
      \centering
      \includegraphics[width=\linewidth]{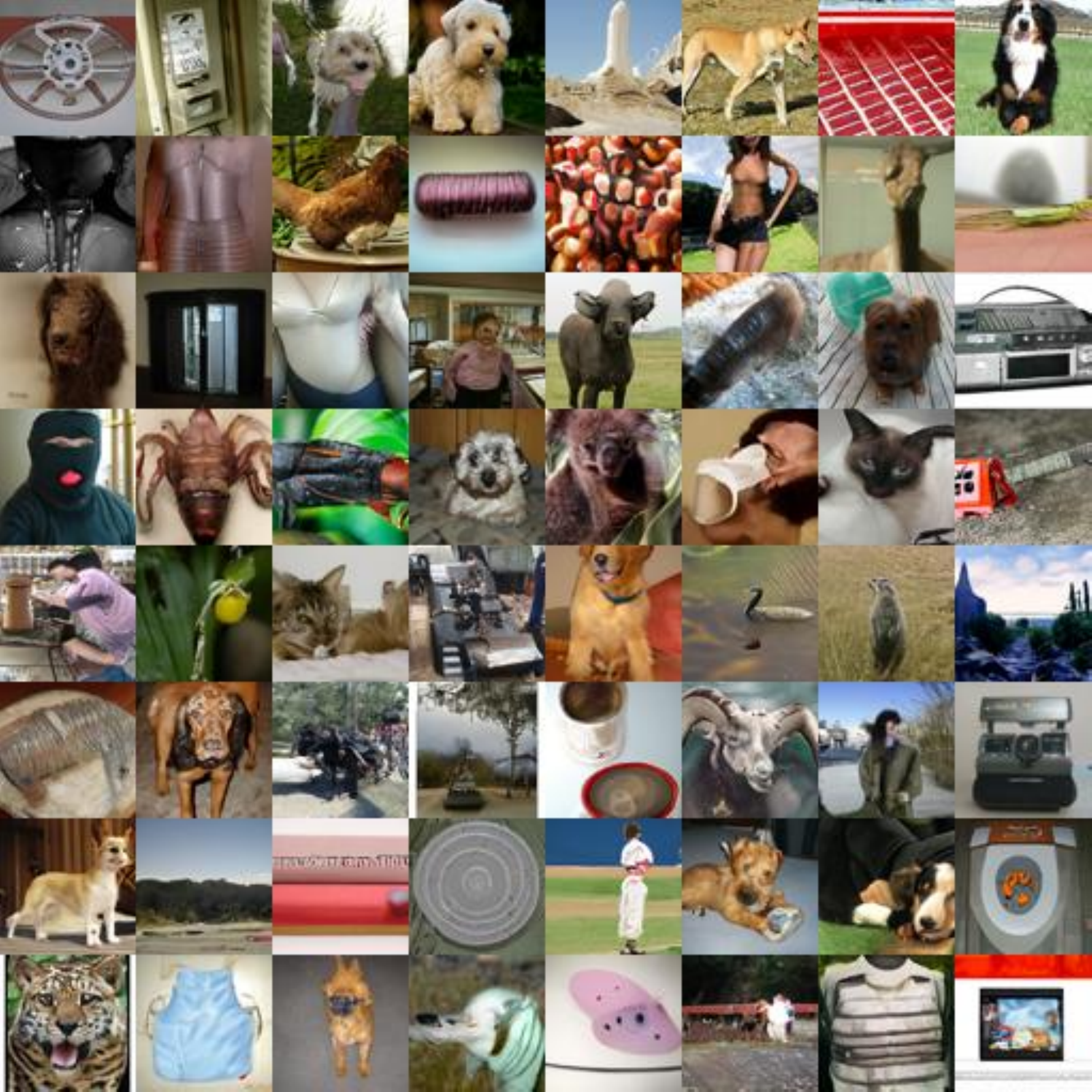}
      \caption{Bit Diffusion on \graycode (FID=$5.14$)}
    \end{subfigure}
    \begin{subfigure}{0.49\textwidth}
      \centering
      \includegraphics[width=\linewidth]{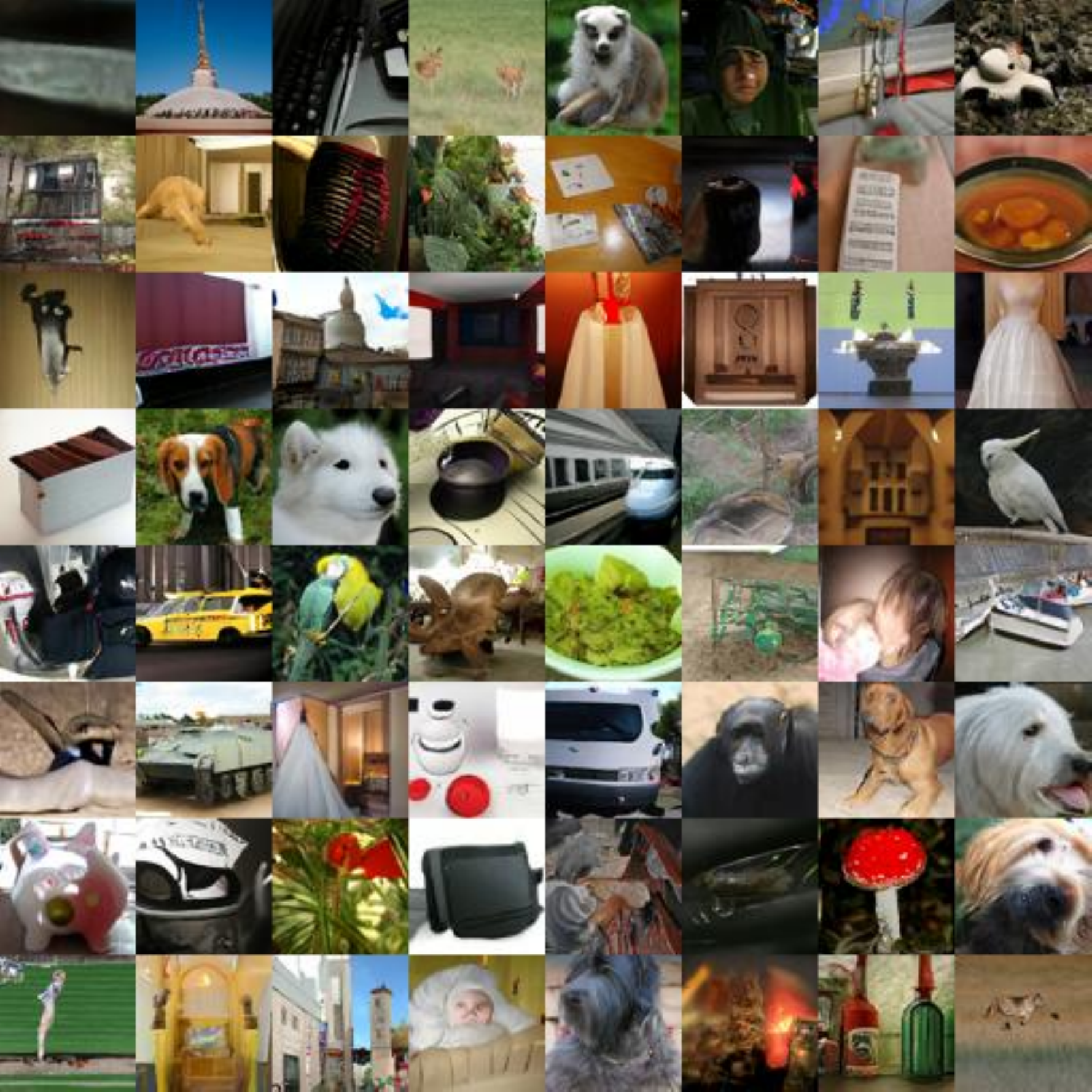}
      \caption{Bit Diffusion on \uints (FID=$8.76$)}
    \end{subfigure}
    \caption{\label{fig:rand_samples_imagenet} Random samples from class-conditional models trained on \imagenet. 
    (a) is for continuous image generation, (b), (c), and (d) are for discrete/categorial image generation.
    }
\end{figure*}

Figure~\ref{fig:rand_samples_cifar10} shows random samples (non cherry-picked) from unconditional diffusion models on \cifar with continuous pixels and analog bits. 

Figure~\ref{fig:rand_samples_imagenet} shows random samples (non cherry-picked) from class-conditional diffusion models on \imagenet with continuous pixels and analog bits.

\clearpage\newpage

\section{On Sampling Strategies with Self-Conditioning}
\label{sec:self_cond_sampling}

\subsection{Method}
In this section, we present extensions to the default sampling strategy with Self-Conditioning in Algorithm~\ref{alg:sample}. The default sampling strategy utilizes data estimate from the previous step as the conditional input to the denoising network for producing data estimate at the current step. While this is both simple and effective, we observe that, for \uints encoding of pixels, as the number of sampling steps increases (with both DDIM or DDPM samplers), the generated samples tend to be over-smoothed. We propose the following two extensions of the default sampling strategy to mitigate the issues and provide improvements when using larger sampling steps.

\textbf{Self-Conditioning based on Momentum Estimate }
The first extension to the default sampling strategy is to adopt an exponential moving average over the previous data estimate to provide a more reliable conditioning input, similar to a momentum optimizer.
The detailed procedure is shown in algorithm~\ref{alg:sample_strategy_ema}, where the differences from the default sampling strategy are highlighted in blue. Note that the default sampling strategy can also be considered as a special case of this generalized form in that the momentum is set to zero.

\textbf{Self-Conditioning based on Self-Guidance }
One potential issue with the default sampling strategy is the slight discrepancy of the Self-Conditioning signal during training and inference/sampling. Specifically, during training, the Self-Conditioning signal is the data estimate from the same time step, while, during sampling, it is from the past time step(s). Therefore, here we propose an approach that also use the same step data estimate for self-conditioning, which comes at the cost of extra forward pass over the denoising network at sampling time. Specifically, we conduct two forward passes of denoising network per sampling step, one with zero data estimate and the other with current data step estimate, and then we use a weighted combination, similar to~\citep{ho2021classifier}, of both prediction to form the final prediction at the current step. The detailed procedure is given in algorithm~\ref{alg:sample_strategy_selfguide} with differences to the default sampling strategy highlighted.

\vfill
\begin{minipage}[]{0.89\textwidth}
\begin{algorithm}[H]
\small
\caption{\small Sampling with Self-Conditioning based on Momentum Estimate.
}
\label{alg:sample_strategy_ema}
\definecolor{codeblue}{rgb}{0.25,0.5,0.5}
\definecolor{codekw}{rgb}{0.85, 0.18, 0.50}
\lstset{
  backgroundcolor=\color{white},
  basicstyle=\fontsize{7.5pt}{7.5pt}\ttfamily\selectfont,
  columns=fullflexible,
  breaklines=true,
  captionpos=b,
  commentstyle=\fontsize{7.5pt}{7.5pt}\color{codeblue},
  keywordstyle=\fontsize{7.5pt}{7.5pt}\color{codekw},
  escapechar={|}, 
}
\begin{lstlisting}[language=python]
def generate(steps, |td|=0, |\color{blue}momentum=0.|):
  x_t = normal(mean=0, std=1)
  |x\_pred = zeros\_like(x\_t)|
  |\color{blue}x\_accu = zeros\_like(x\_t)|
  
  for step in range(steps):
    # Get time for current and next states.
    t_now = 1 - step / steps
    t_next = max(1 - (step + 1 + |td|) / steps, 0)
        
    # Predict x_0 (with self-cond).
    |\color{blue}x\_accu = momentum * x\_accu + (1 - momentum) * x\_pred|
    x_pred = net(cat([x_t, |\color{blue}x\_accu|], -1), t_now)
    
    # Estimate x at t_next.
    x_t = ddim_or_ddpm_step(x_t, x_pred, t_now, t_next)
    
  # Binary decoding: analog bits to discrete data.
  |x\_int = bit2int(x\_pred > 0)|
  return x_int
\end{lstlisting}
\end{algorithm}
\end{minipage}

\begin{minipage}[]{0.89\textwidth}
\begin{algorithm}[H]
\small
\caption{\small Sampling with Self-Conditioning based on Self-Guidance.
}
\label{alg:sample_strategy_selfguide}
\definecolor{codeblue}{rgb}{0.25,0.5,0.5}
\definecolor{codekw}{rgb}{0.85, 0.18, 0.50}
\lstset{
  backgroundcolor=\color{white},
  basicstyle=\fontsize{7.5pt}{7.5pt}\ttfamily\selectfont,
  columns=fullflexible,
  breaklines=true,
  captionpos=b,
  commentstyle=\fontsize{7.5pt}{7.5pt}\color{codeblue},
  keywordstyle=\fontsize{7.5pt}{7.5pt}\color{codekw},
  escapechar={|}, 
}
\begin{lstlisting}[language=python]
def generate(steps, |td|=0, |\color{blue}guide\_w=3.0|):
  x_t = normal(mean=0, std=1)
  |x\_pred = zeros\_like(x\_t)|
  
  for step in range(steps):
    # Get time for current and next states.
    t_now = 1 - step / steps
    t_next = max(1 - (step + 1 + |td|) / steps, 0)
        
    # Predict x_0  wo/ self-cond.
    |\color{blue}x\_pred\_uncond = net(cat([x\_t, zeros\_like(x\_t)], -1), t\_now)|
    # Predict x_0 w/ self-cond.
    |\color{blue}x\_pred\_selfcond = net(cat([x\_t, x\_pred\_uncond], -1), t\_now)|
    # Apply self-guidance.
    |\color{blue}x\_pred = guide\_w * x\_pred\_selfcond + (1.0 - guide\_w) * x\_pred\_uncond|
    
    # Estimate x at t_next.
    x_t = ddim_or_ddpm_step(x_t, x_pred, t_now, t_next)
    
  # Binary decoding: analog bits to discrete data.
  |x\_int = bit2int(x\_pred > 0)|
  return x_int
\end{lstlisting}
\end{algorithm}
\end{minipage}

\subsection{Experiments}

Table~\ref{tab:best_FID_for_samplers} reports the best FID scores across various sampling strategies discussed here (as well as samplers, sampling steps, time difference in asymmetric time intervals).
\begin{table}[h]
    \centering
    \small
    \caption{Best FIDs of Bit Diffusion models with different Self-Conditioning sampling strategies on conditional \imagenet.}
    \begin{tabular}{c|ccc}
    \toprule
     & \uint & \graycode & \uints \\
    \midrule
    Default sampling (momentum$=0$) & $\mathbf{4.84}$ & $\mathbf{5.14}$ & $8.76$ \\ Momentum Estimate & $\mathbf{4.85}$ & $\mathbf{5.14}$ & $8.51$ \\ Self-Guidance & $5.15$ & $5.65$ & $\mathbf{7.87}$ \\ \bottomrule
    \end{tabular}
    \label{tab:best_FID_for_samplers}
\end{table}

\begin{figure*}[h]
    \centering
    \begin{subfigure}{0.4\textwidth}
    \includegraphics[width=\linewidth]{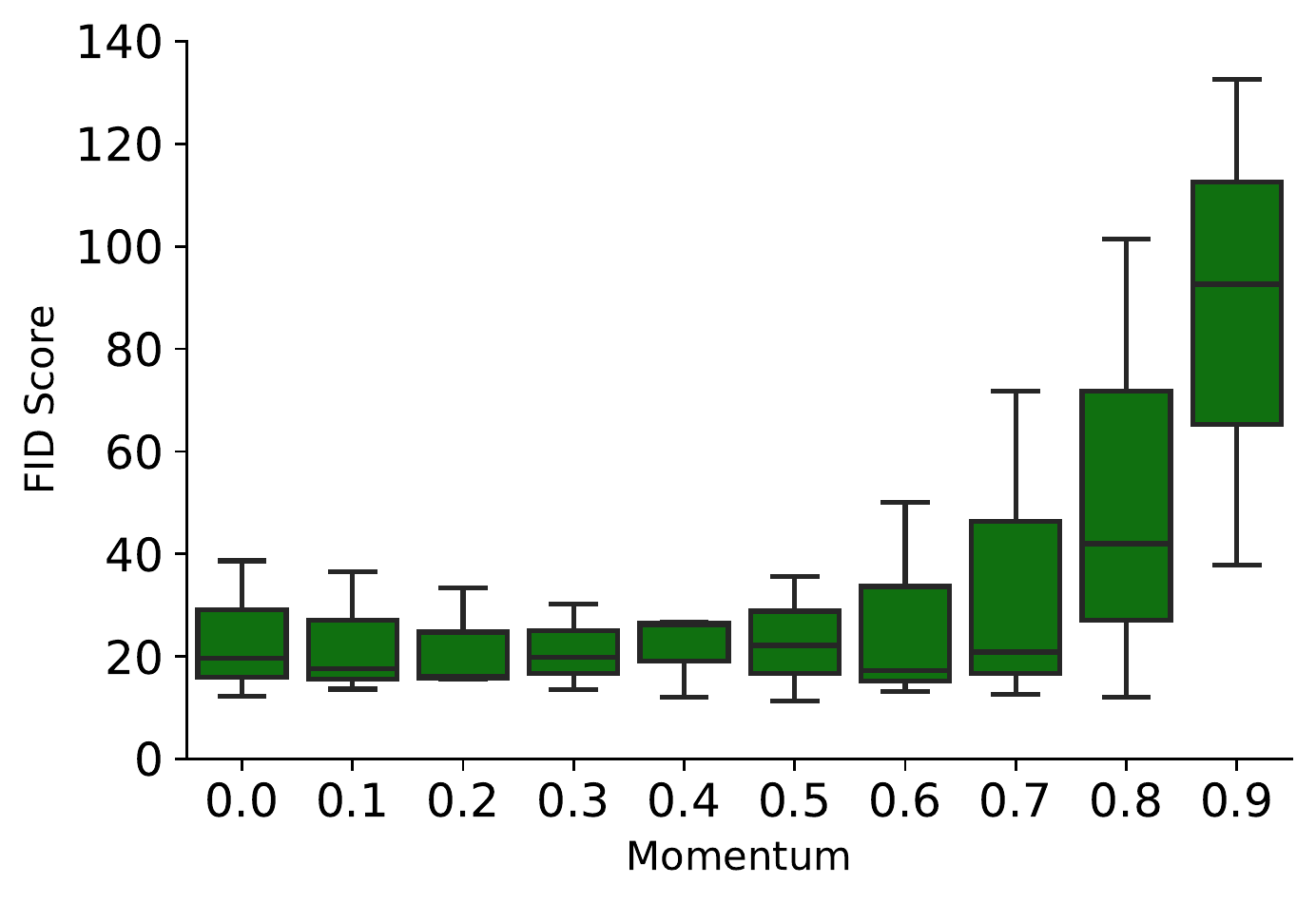}
    \caption{DDIM, number of sampling steps $< 500$.}
    \end{subfigure}
    ~~~~~
    \begin{subfigure}{0.4\textwidth}
    \includegraphics[width=\linewidth]{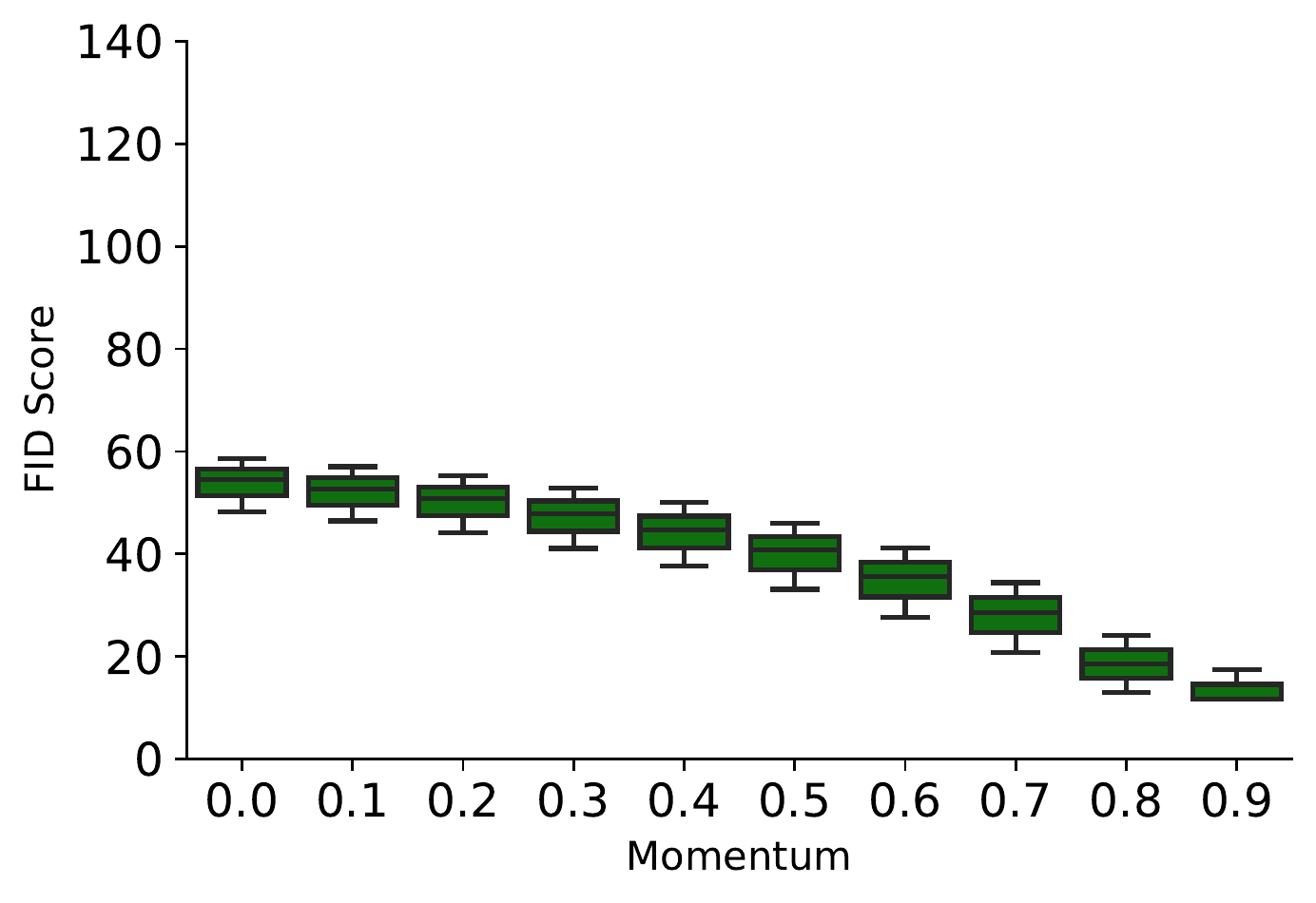}
    \caption{DDIM, number of sampling steps $> 500$.}
    \end{subfigure}
    \begin{subfigure}{0.4\textwidth}
    \includegraphics[width=\linewidth]{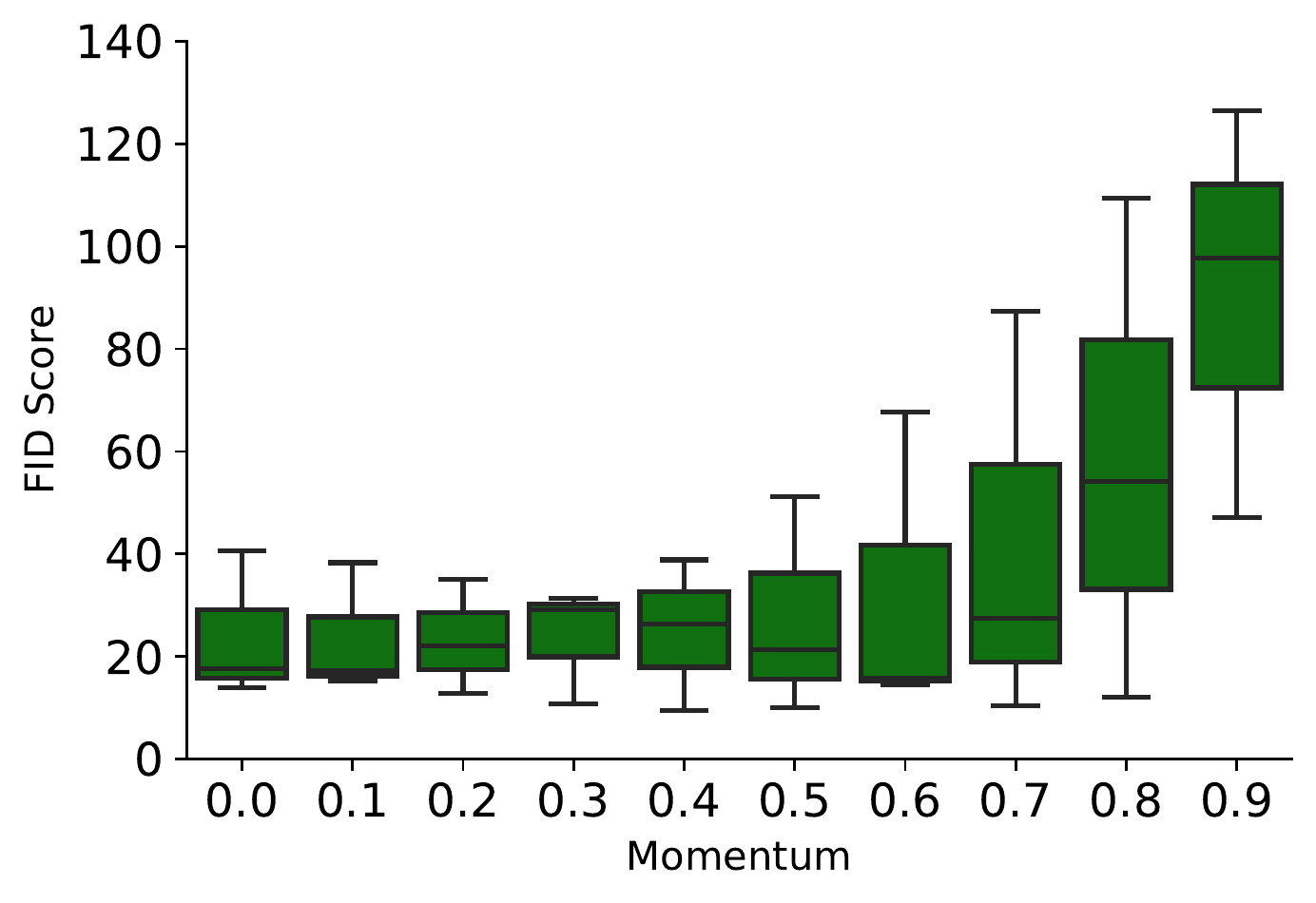}
    \caption{DDPM, number of sampling steps $< 500$.}
    \end{subfigure}    
    ~~~~~
    \begin{subfigure}{0.4\textwidth}
    \includegraphics[width=\linewidth]{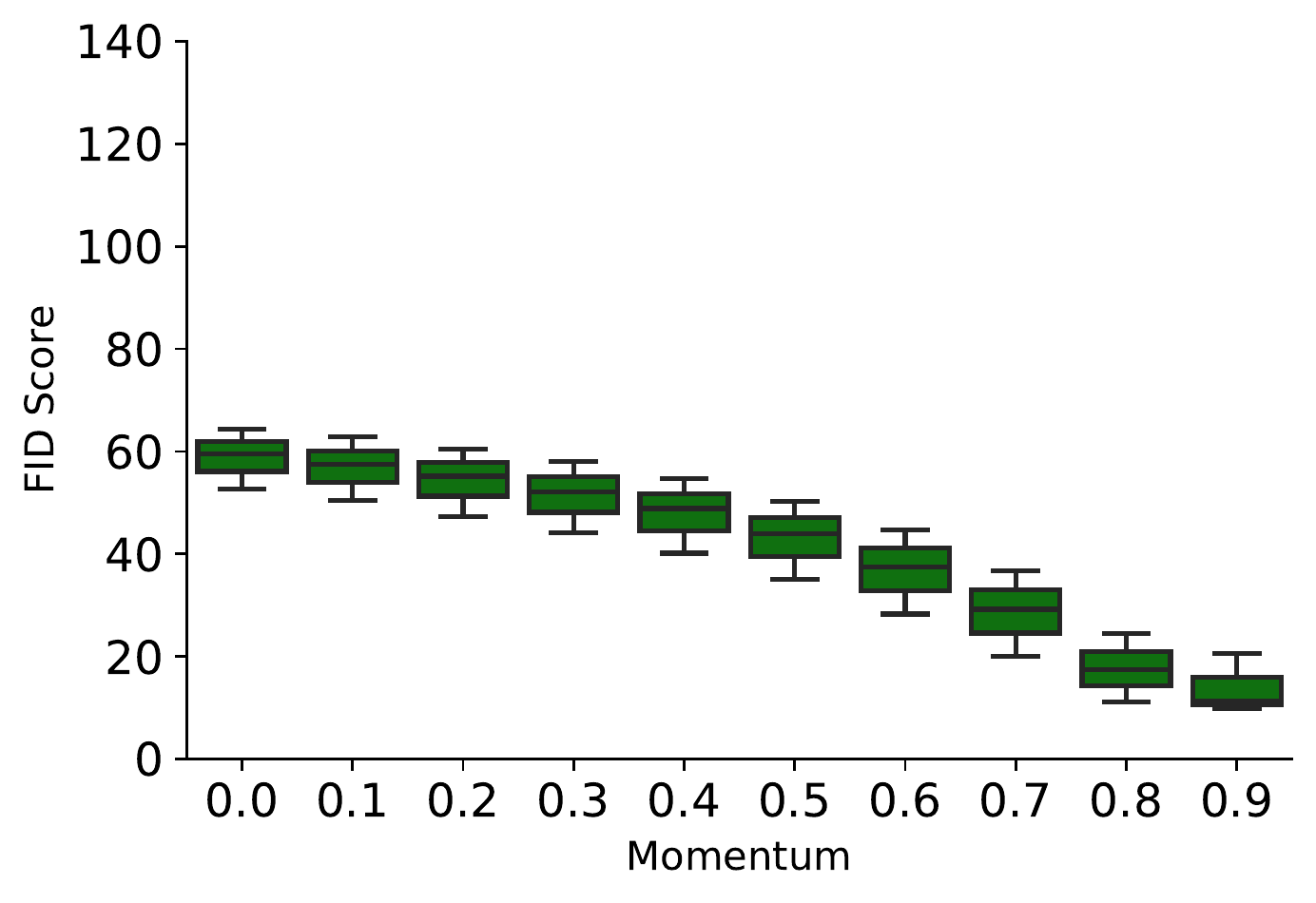}
    \caption{DDPM, number of sampling steps $> 500$.}
    \end{subfigure}
    \caption{\label{fig:ema_fid_distribution} FID on conditional \imagenet with \uints encoding using self-condition sampling based on momentum estimate. The statistics of FID scores in each group are aggregated over the number of sampling steps in $\{ 100, 200, 400, 600, 800, 1000\}$, time difference in $\{0.0, 0.2, 0.4, 0.6, 0.8\}$.}
\end{figure*}

Figure~\ref{fig:ema_fid_distribution} shows FIDs on conditional \imagenet with \uint encoding, using Momentum Estimate with different sampling steps. We find that the momentum on the data estimate is only helpful when sampling steps are larger.

\begin{figure*}[h]
    \centering
    \begin{subfigure}{0.4\textwidth}
    \includegraphics[width=\linewidth]{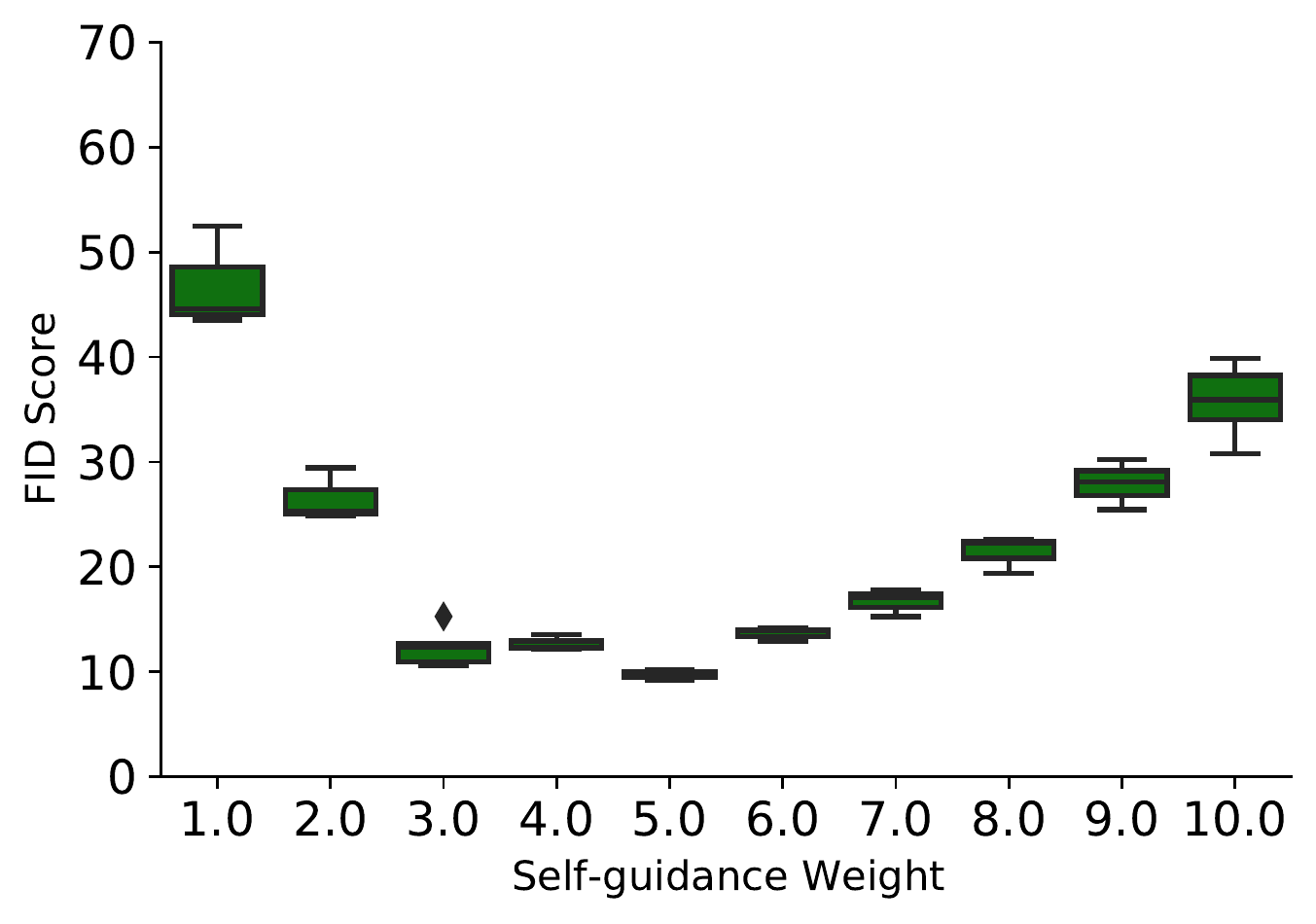}
    \caption{DDIM, number of sampling steps $< 500$.}
    \end{subfigure}
    ~~~~~
    \begin{subfigure}{0.4\textwidth}
    \includegraphics[width=\linewidth]{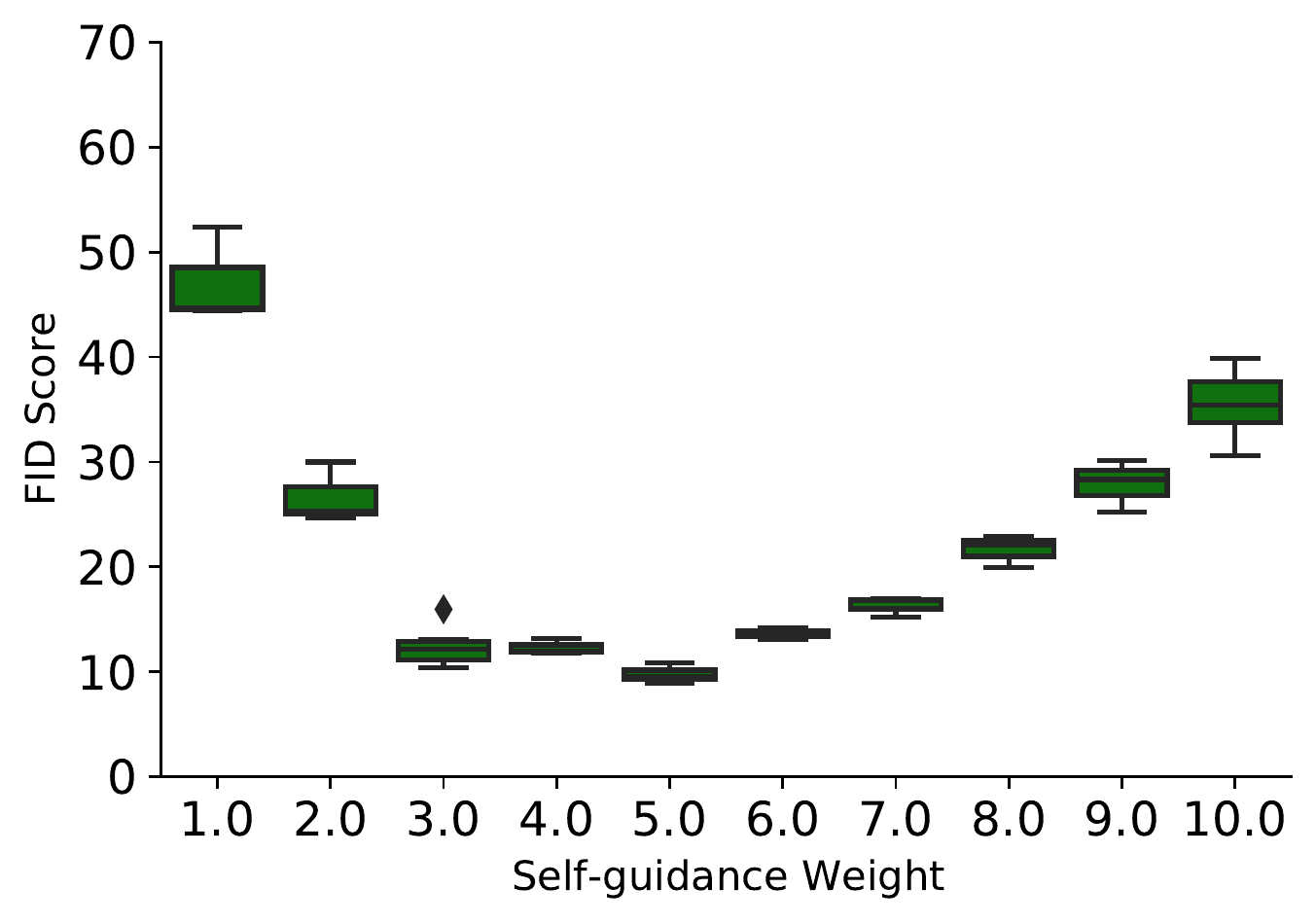}
    \caption{DDIM, number of sampling steps $> 500$.}
    \end{subfigure}
    \begin{subfigure}{0.4\textwidth}
    \includegraphics[width=\linewidth]{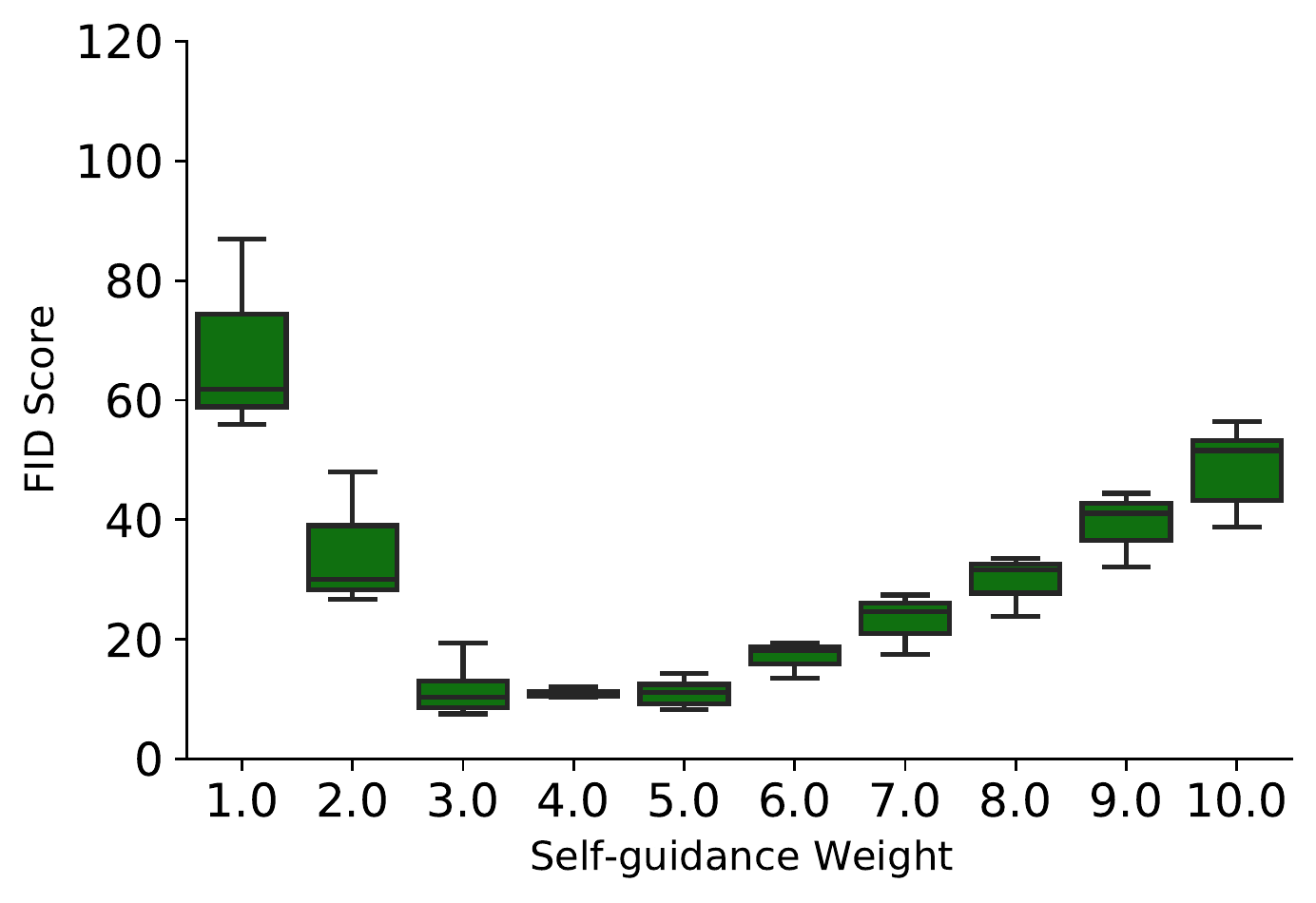}
    \caption{DDPM, number of sampling steps $< 500$.}
    \end{subfigure}
    ~~~~~
    \begin{subfigure}{0.4\textwidth}
    \includegraphics[width=\linewidth]{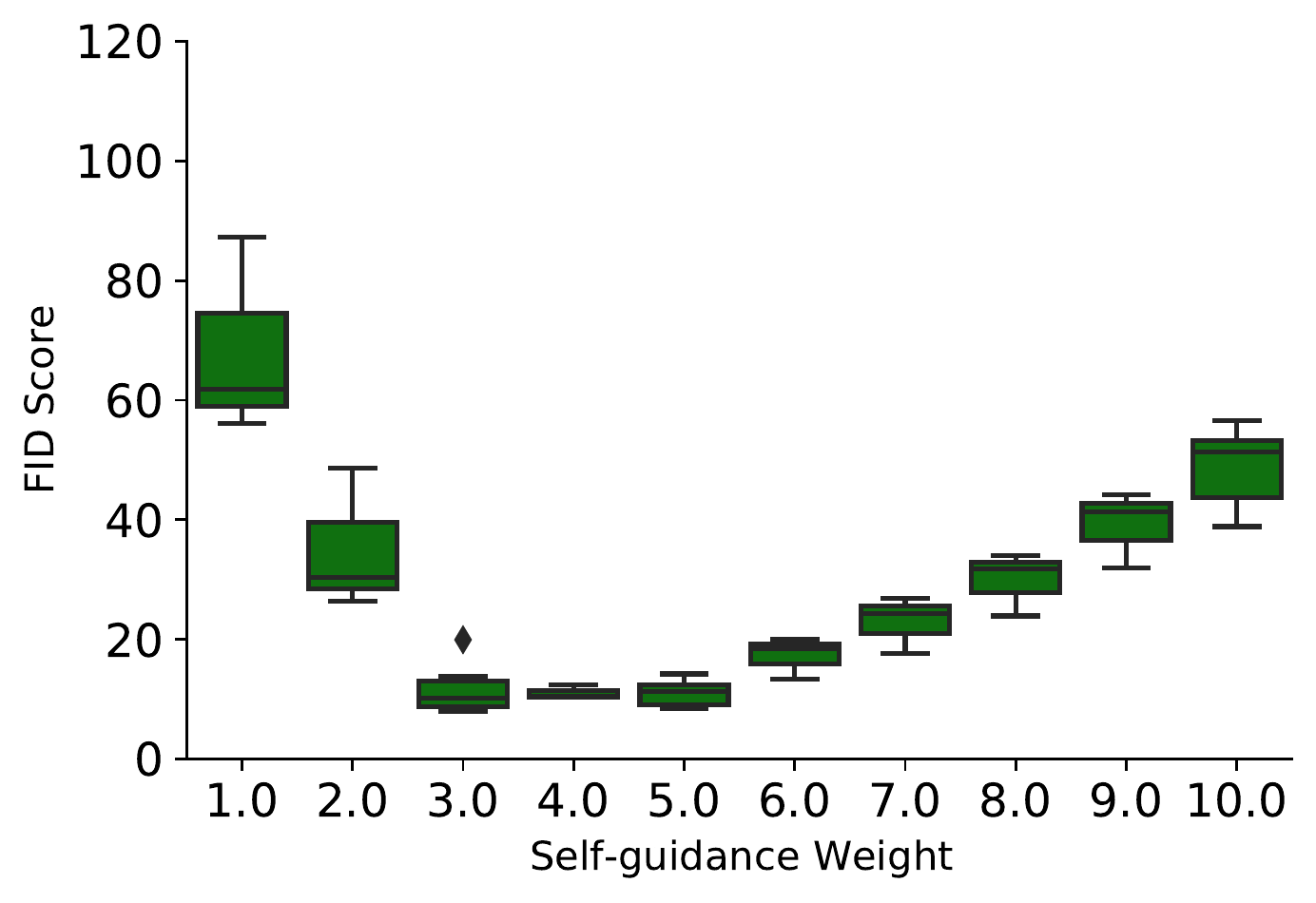}
    \caption{DDPM, number of sampling steps $> 500$.}
    \end{subfigure}
    \caption{\label{fig:sg_fid_distribution} FID on conditional \imagenet with \uints encoding using self-condition sampling based on self-guidance. The statistics of FID scores in each group are aggregated over the number of sampling steps in $\{ 100, 200, 400, 600, 800, 1000\}$, time difference in $\{0.0, 0.1\}$.}
\end{figure*}

Figure~\ref{fig:sg_fid_distribution} shows FIDs on conditional \imagenet with \uint encoding, using Self-Guidance with different sampling steps. We find that a guidance weight between 3.0 and 5.0 is generally preferable and robust to other hyper-parameters (such as sampler choice, sampling steps, and time difference).

\subsection{Samples}

\begin{figure*}\begin{subfigure}{0.49\textwidth}
      \includegraphics[width=\linewidth]{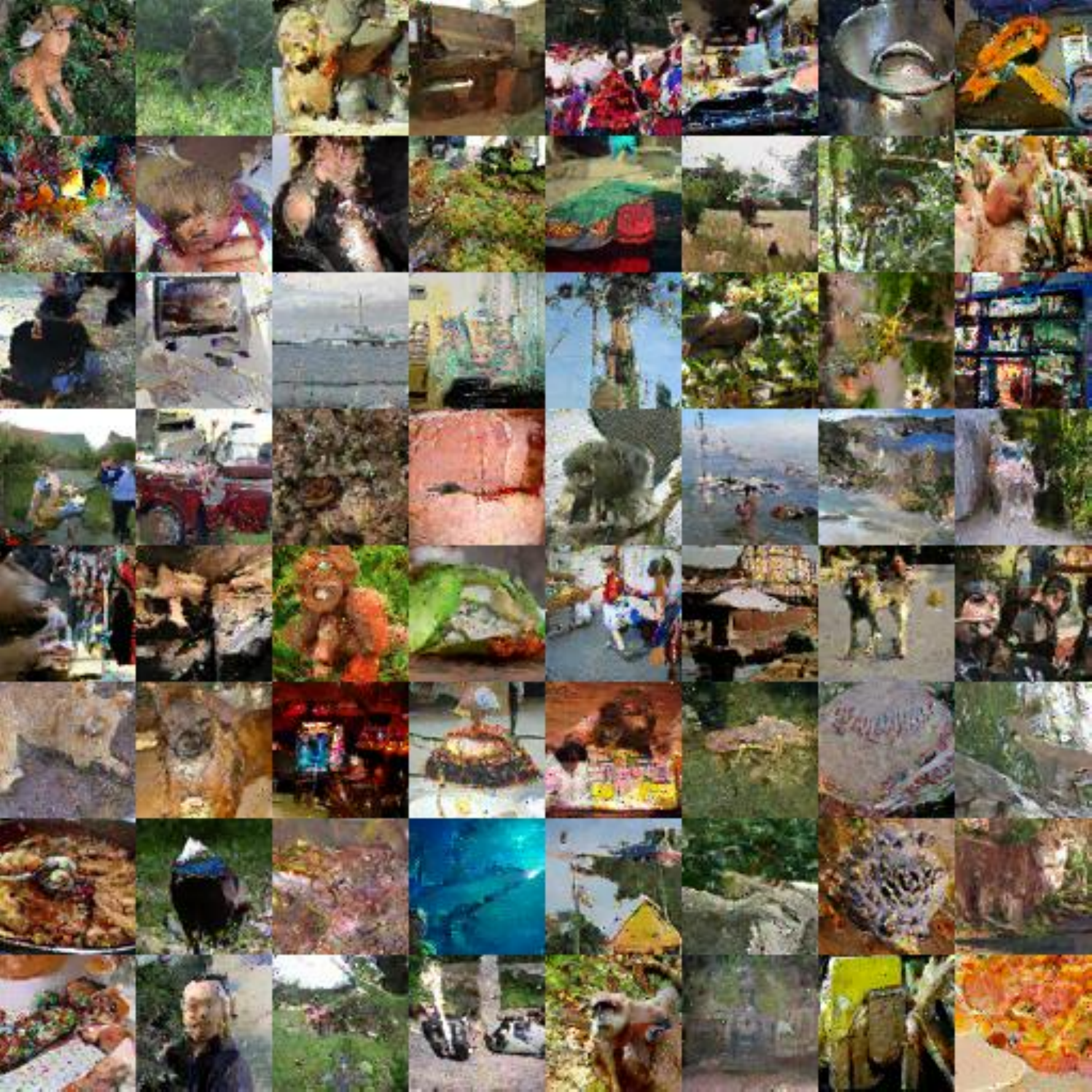}
      \caption{W/o Self-Conditioning, FID=$90.7$.}
    \end{subfigure}
    \begin{subfigure}{0.49\textwidth}
      \includegraphics[width=\linewidth]{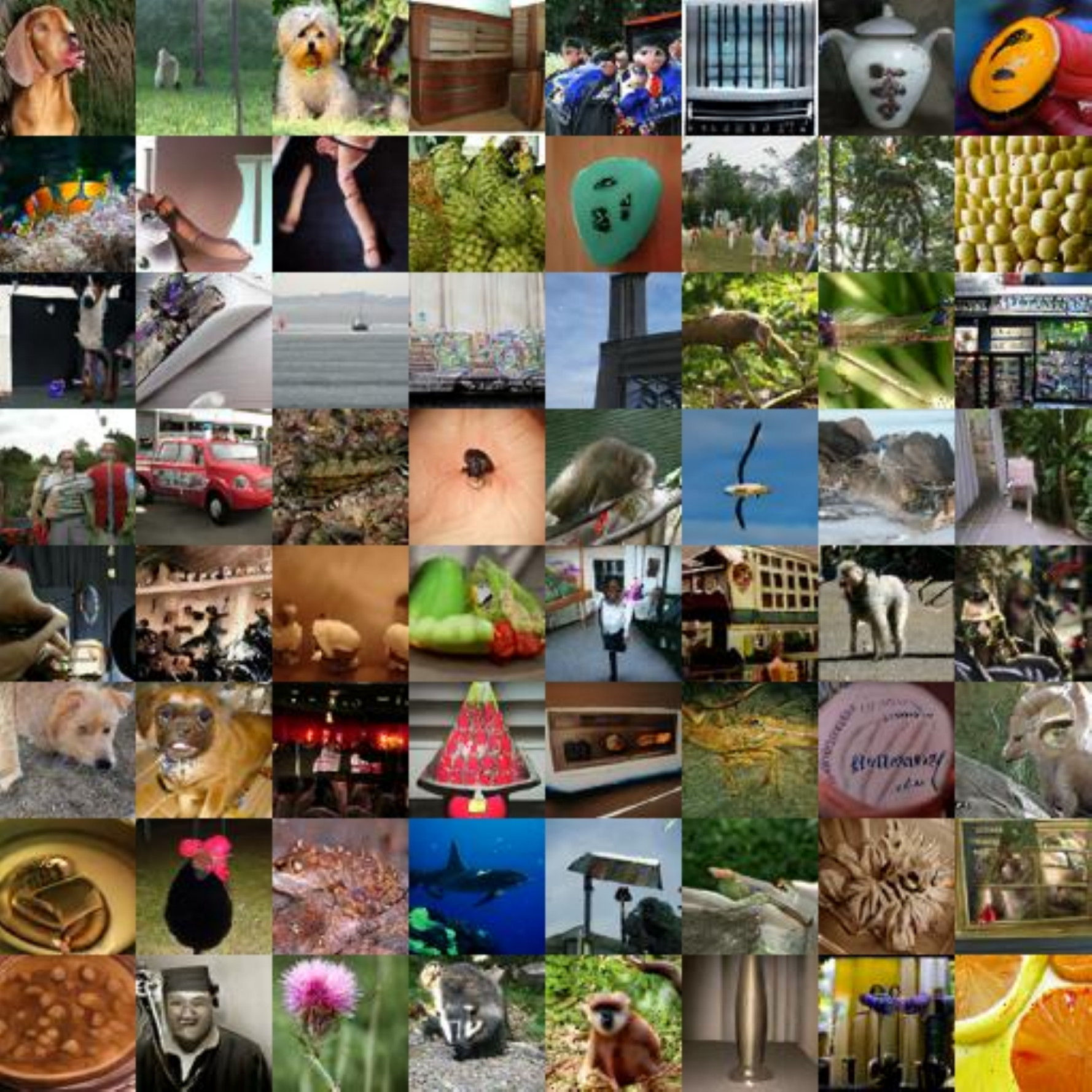}
      \caption{W/ previous estimate (momentum=0), FID=$12.3$.}
    \end{subfigure}
    \begin{subfigure}{0.49\textwidth}
      \includegraphics[width=\linewidth]{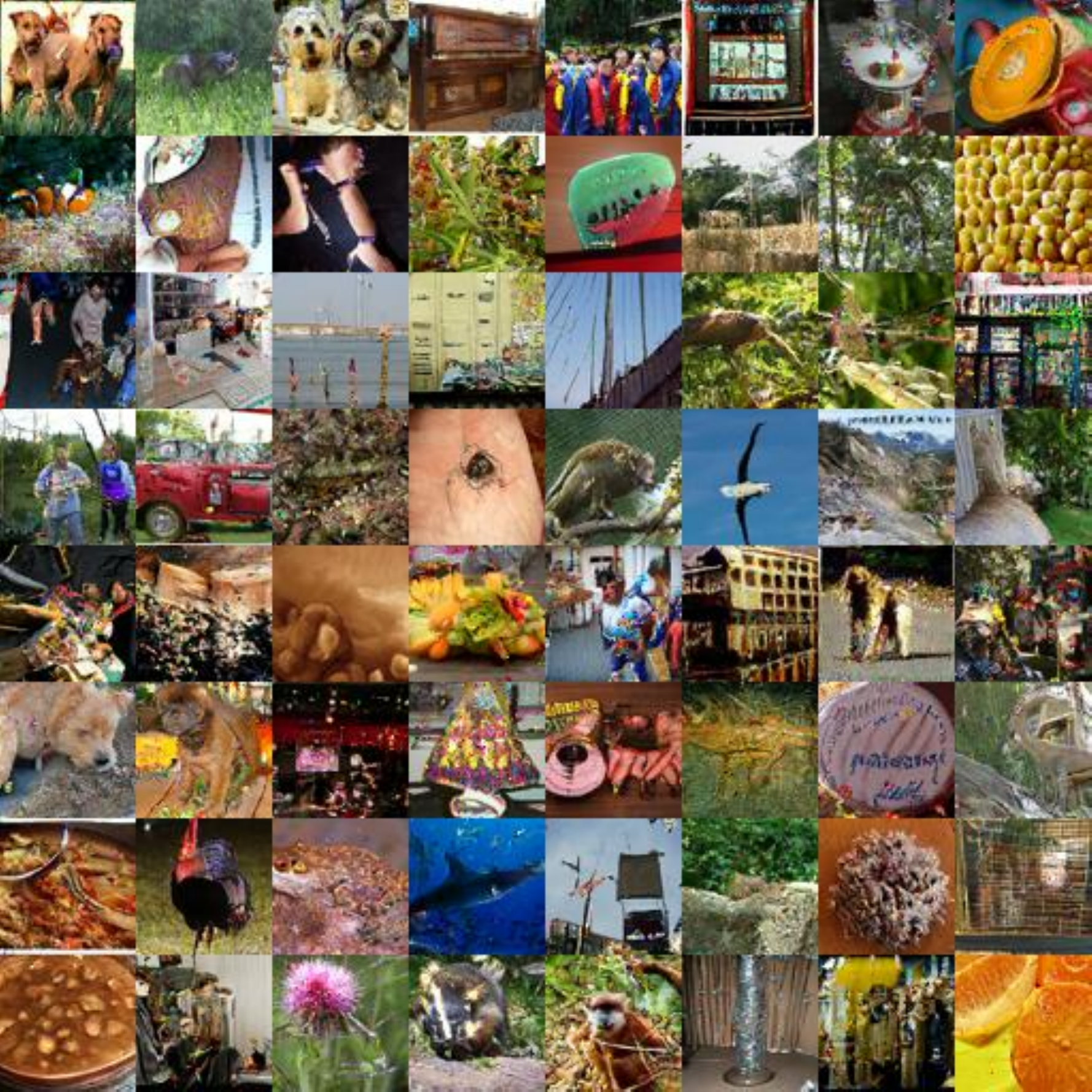}
      \caption{W/ previous estimate (momentum=0.5), FID=$35.5$.}
    \end{subfigure}
    \;
    \begin{subfigure}{0.49\textwidth}
      \includegraphics[width=\linewidth]{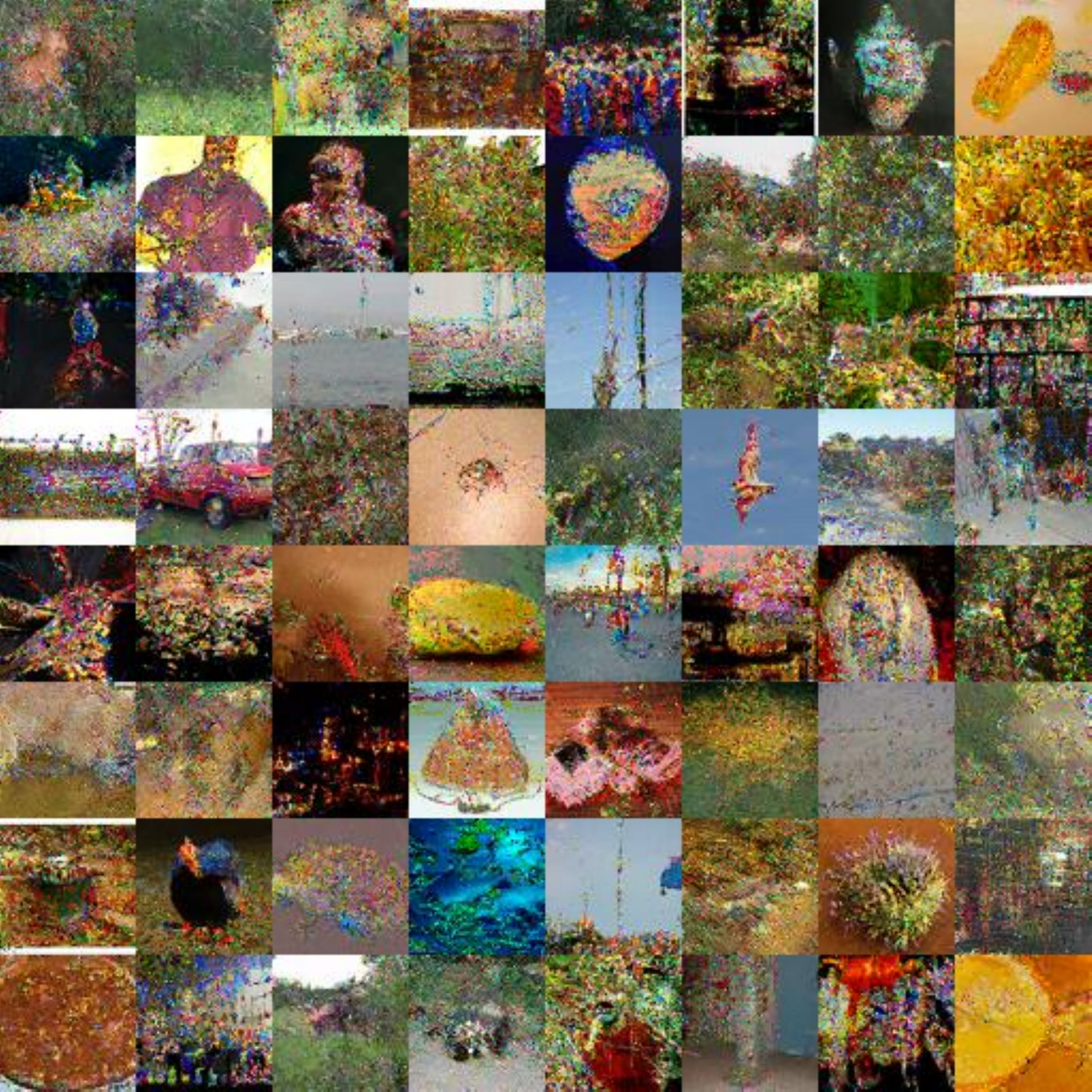}
      \caption{W/ prev. estimate (momentum=0.9), FID=$132.6$.}
    \end{subfigure}
    \begin{subfigure}{0.49\textwidth}
      \includegraphics[width=\linewidth]{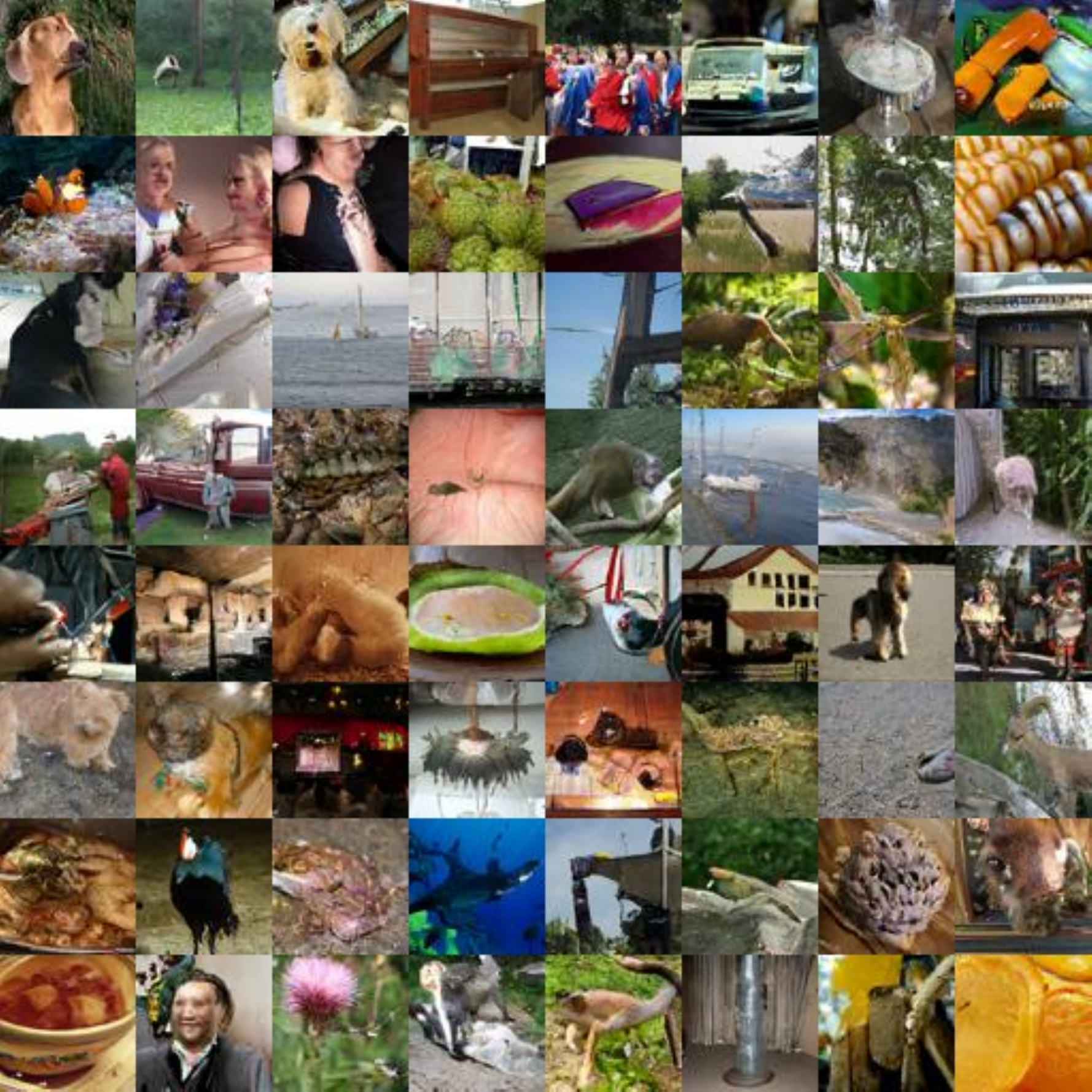}
      \caption{W/ same-step estimate ($w=3.0$), FID=$15.6$.}
    \end{subfigure}
    ~~
    \begin{subfigure}{0.49\textwidth}
      \includegraphics[width=\linewidth]{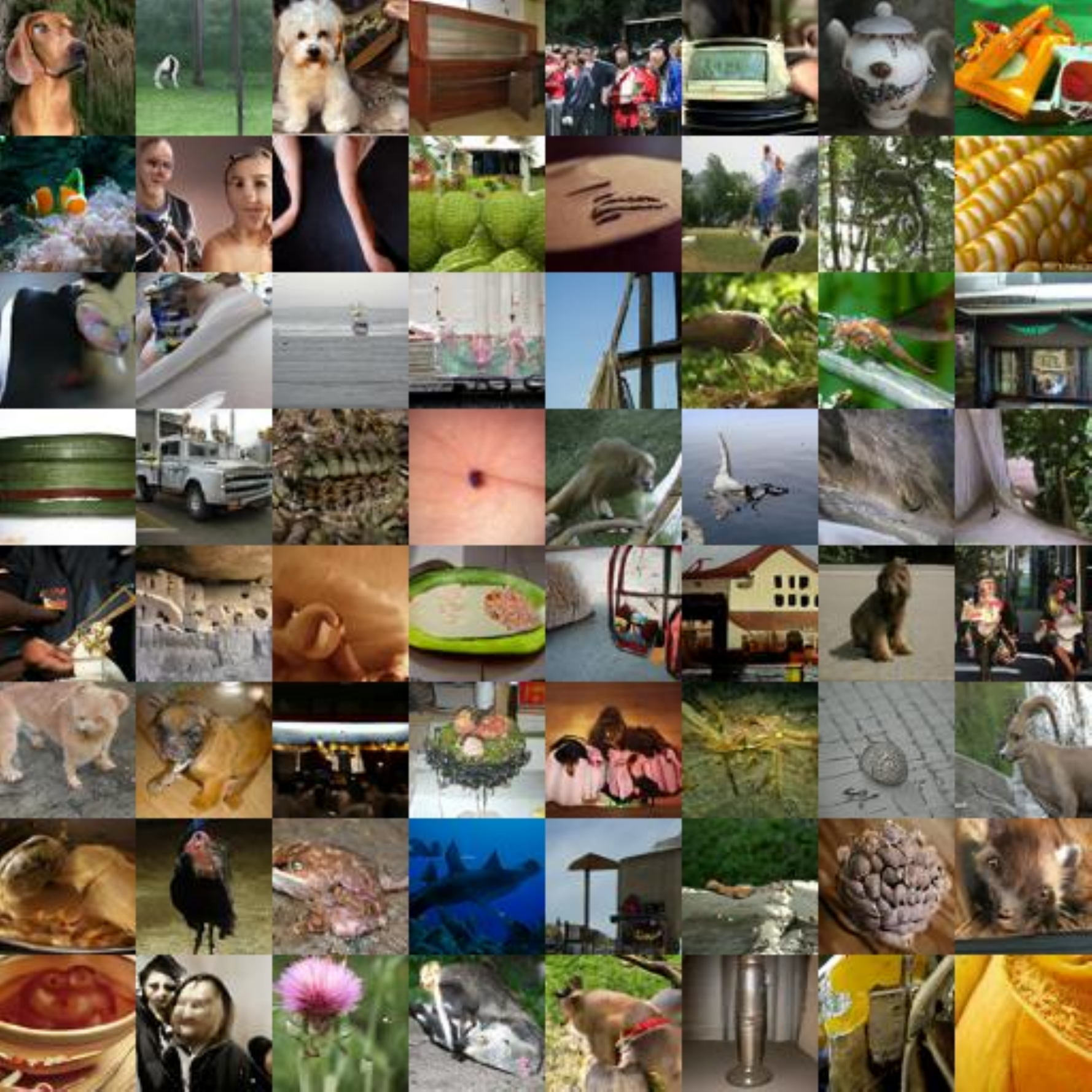}
      \caption{W/ same-step estimate ($w=5.0$), FID=$9.3$.}
    \end{subfigure}
    \caption{\label{fig:same_latent_samples_100} Random samples of Bit Diffusion with \uints on categorical \imagenet using various Self-Conditioning sampling strategies. Different plots share the same set of $\bm{x}_T$. Sampling with {100} steps of DDIM without asymmetric time intervals. }
\end{figure*}

\begin{figure*}\begin{subfigure}{0.49\textwidth}
      \includegraphics[width=\linewidth]{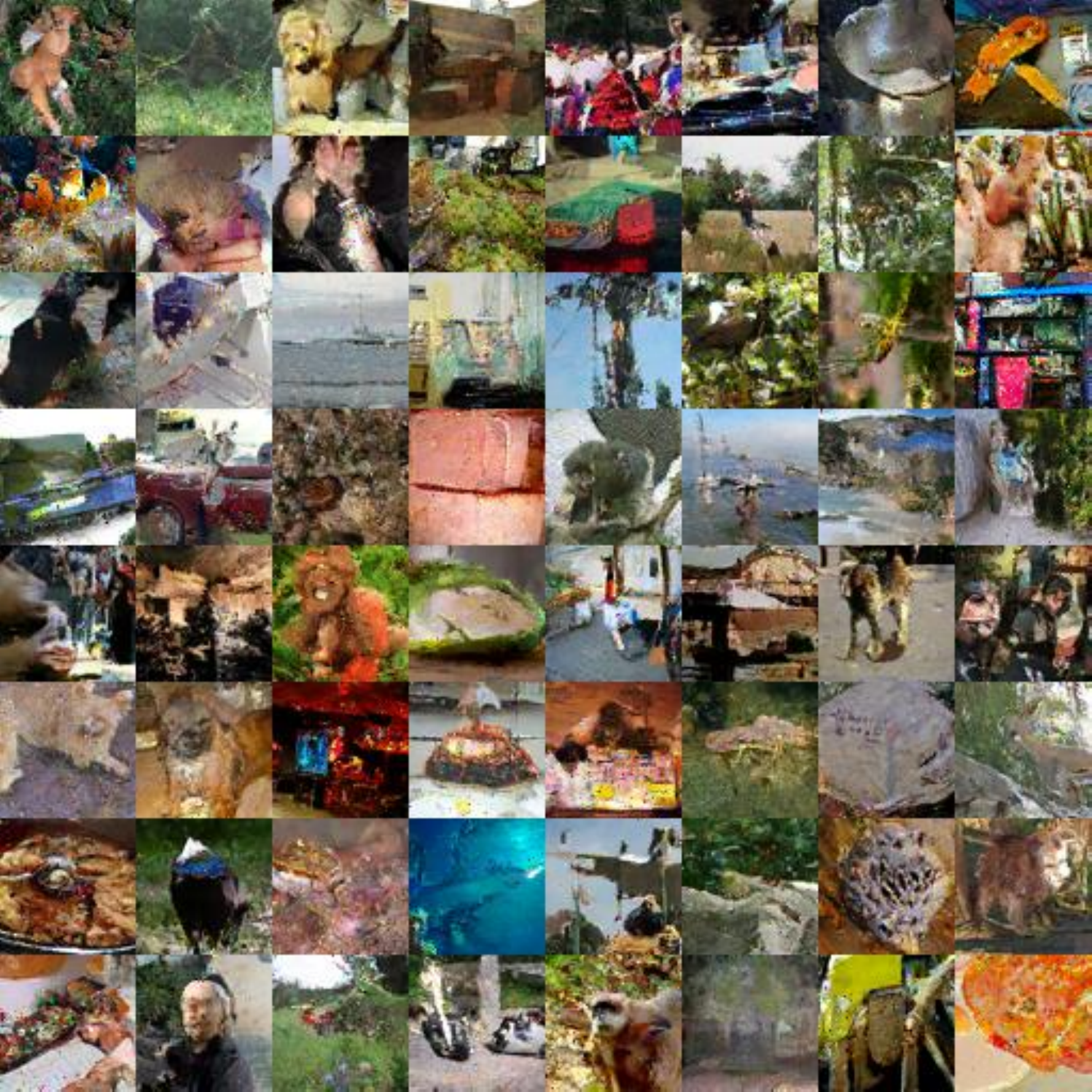}
      \caption{W/o Self-Conditioning, FID=$73.60$.}
    \end{subfigure}
    \begin{subfigure}{0.49\textwidth}
      \includegraphics[width=\linewidth]{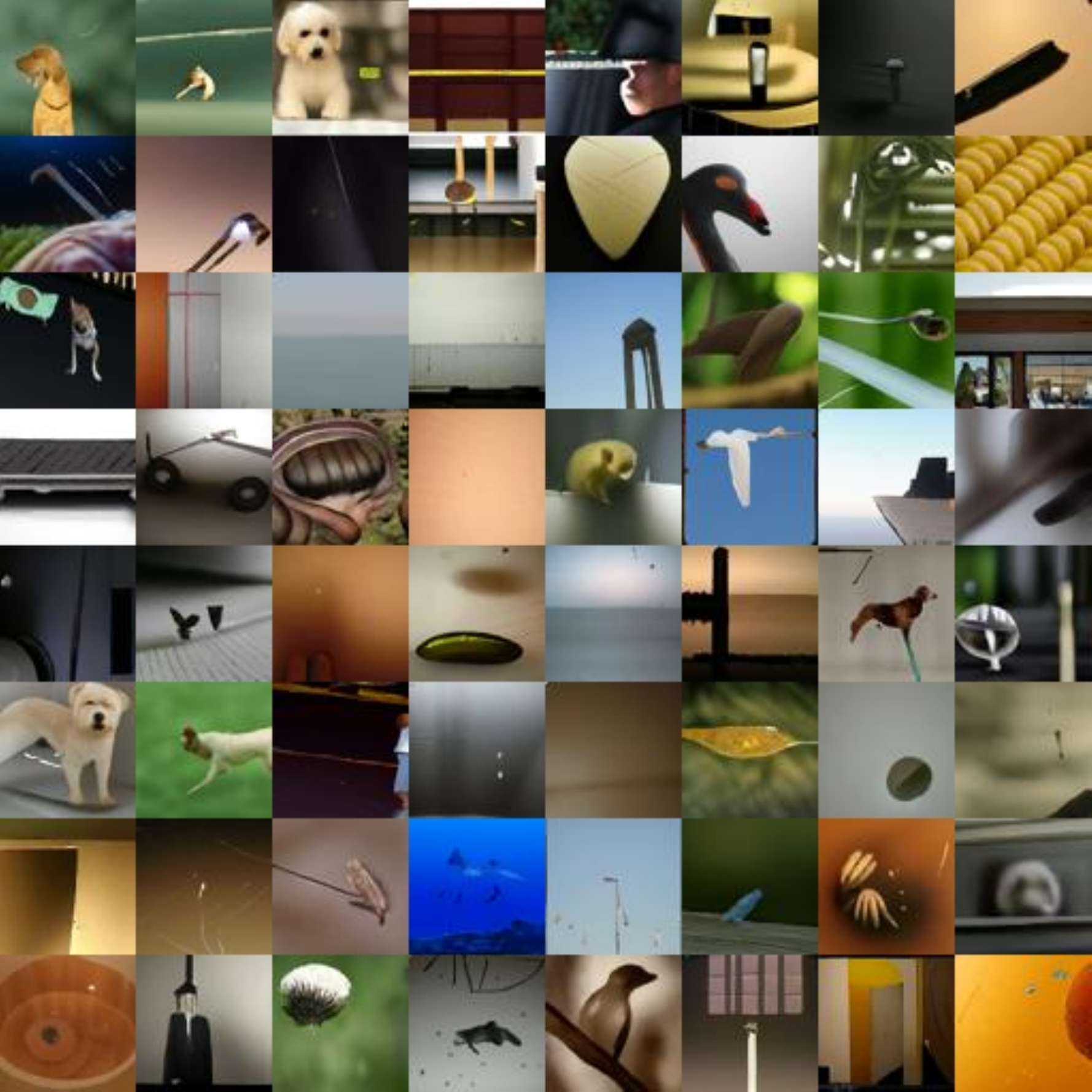}
      \caption{W/ previous estimate (momentum=0), FID=$58.6$.}
    \end{subfigure}
    \begin{subfigure}{0.49\textwidth}
      \includegraphics[width=\linewidth]{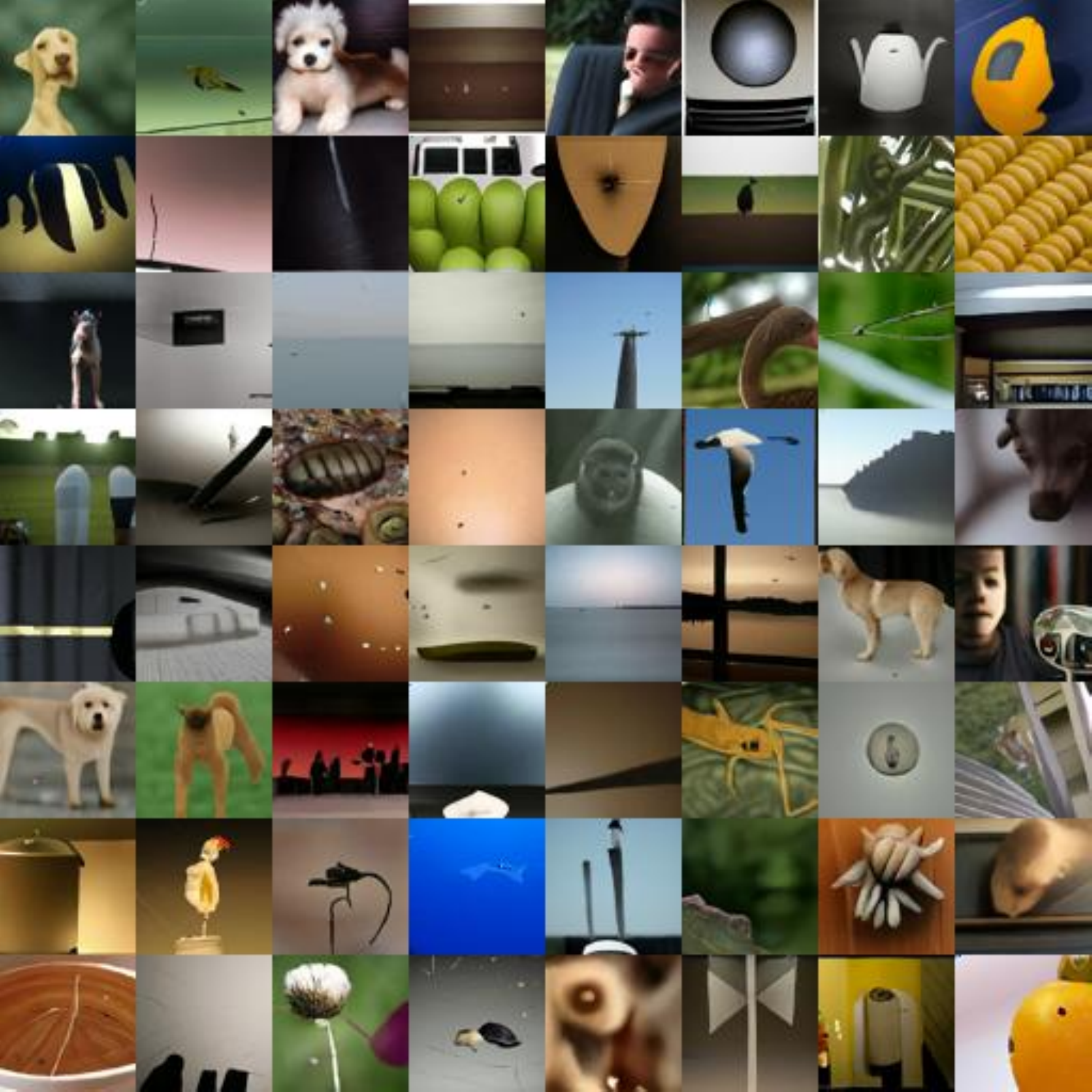}
      \caption{W/ previous estimate (momentum=0.5), FID=$46.0$.}
    \end{subfigure}
    \begin{subfigure}{0.49\textwidth}
      \includegraphics[width=\linewidth]{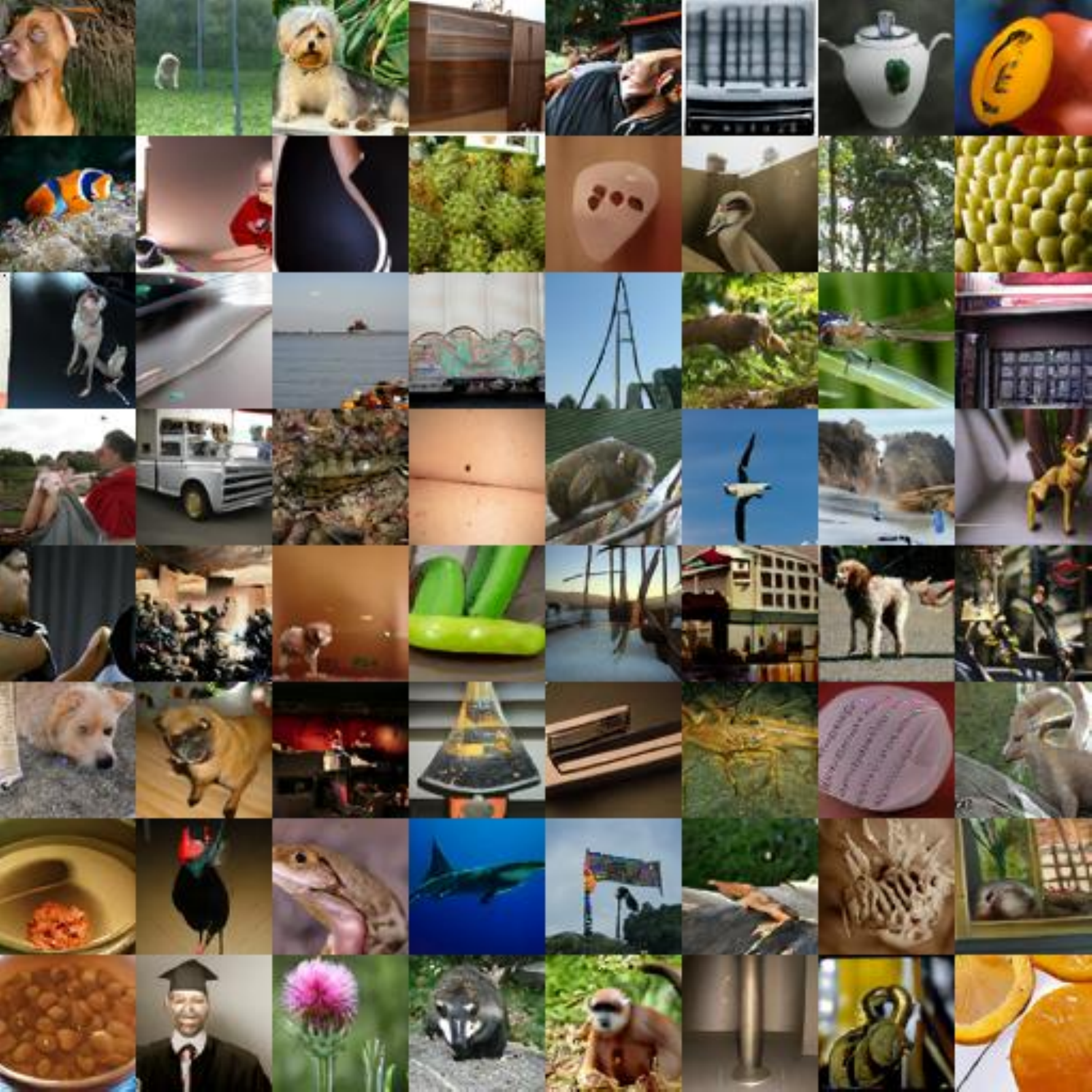}
      \caption{W/ previous estimate (momentum=0.9), FID=$11.7$.}
    \end{subfigure}
    \begin{subfigure}{0.49\textwidth}
      \includegraphics[width=\linewidth]{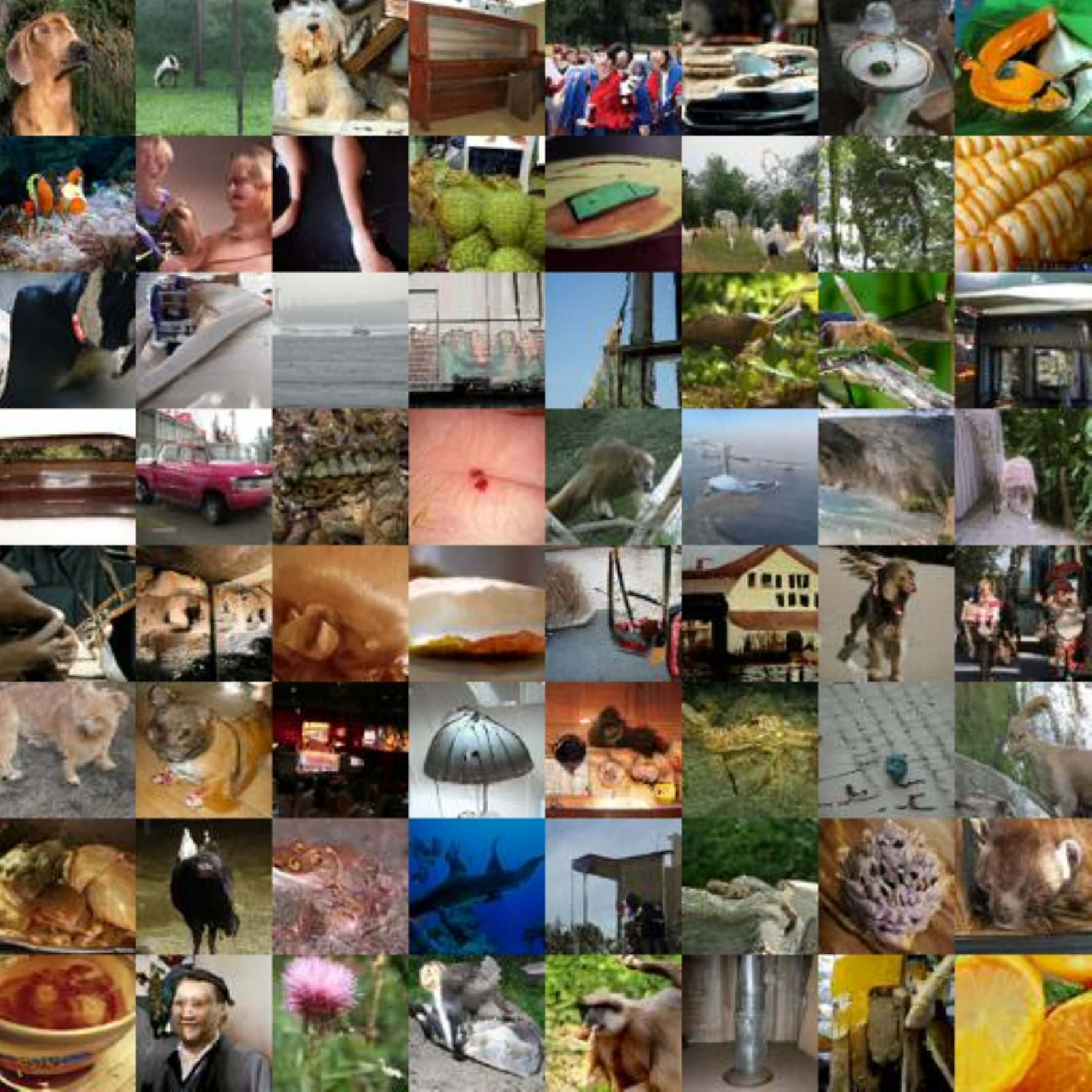}
      \caption{W/ same-step estimate ($w=3.0$), FID=$11.8$.}
    \end{subfigure}
    ~~
    \begin{subfigure}{0.49\textwidth}
      \includegraphics[width=\linewidth]{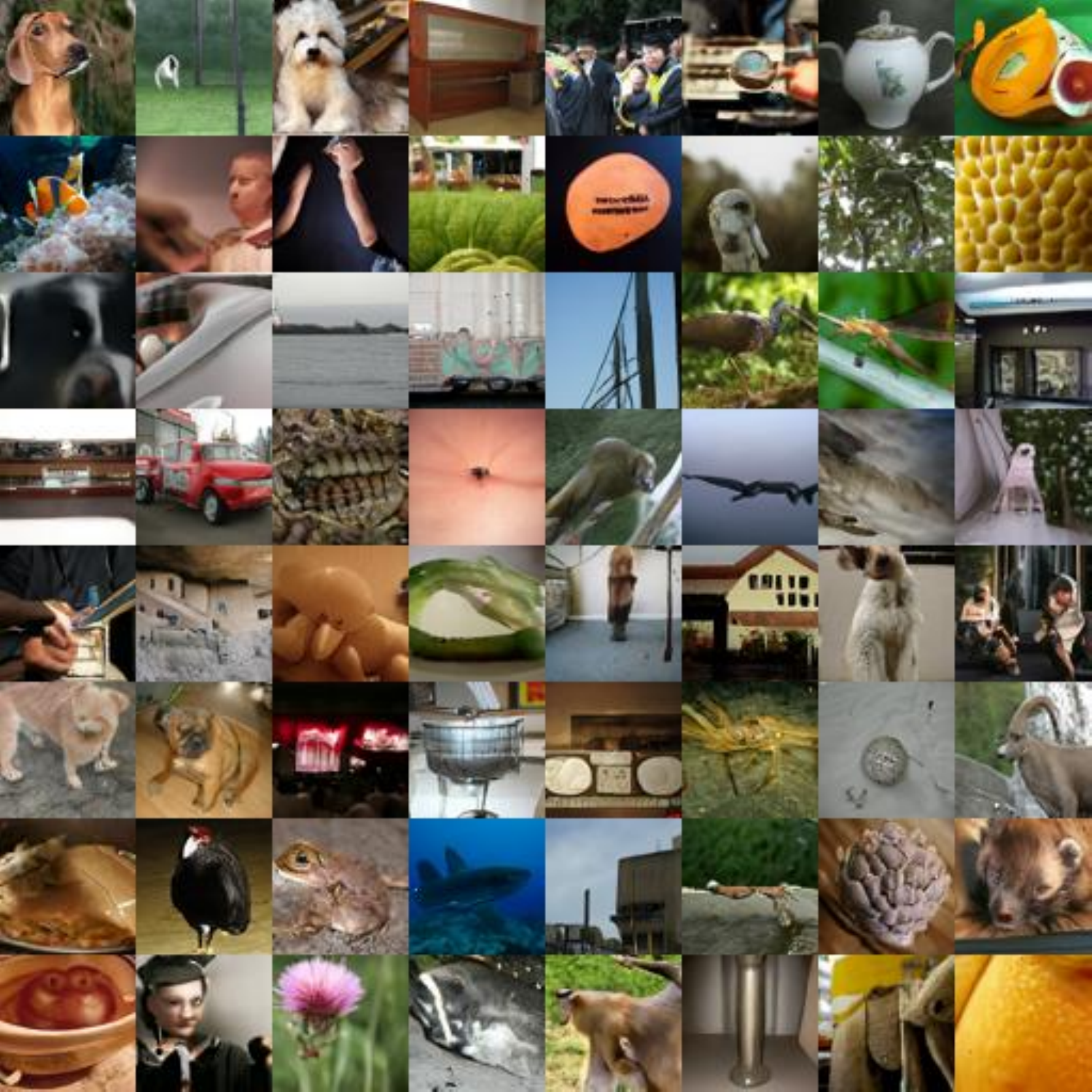}
      \caption{W/ same-step estimate ($w=5.0$), FID=$9.6$.}
    \end{subfigure}
    \caption{\label{fig:same_latent_samples_1000} Random samples of Bit Diffusion with \uints on categorical \imagenet using various Self-Conditioning sampling strategies. Different plots share the same set of $\bm{x}_T$. Sampling with {1000} steps of DDIM without asymmetric time intervals. }
\end{figure*}

Figure~\ref{fig:same_latent_samples_100} and~\ref{fig:same_latent_samples_1000} provide generated samples from different sampling strategies with 100 and 1000 DDIM sampling steps, respectively.

\end{document}